\documentclass[twoside,11pt]{article}

\usepackage{blindtext}

\usepackage[preprint]{jmlr2e}

\usepackage{amsfonts}
\usepackage{amsthm, amsmath,amssymb}
\usepackage{mathtools}
\usepackage{comment}
\usepackage{url}

\usepackage{bbm}

\usepackage{amsthm}
\usepackage{caption}
\usepackage{subcaption}

\newcommand{\B}{{\mathcal B}}

\newcommand{\Pc}{{\mathcal P}}

\newcommand{\Hc}{{\mathcal H}}

\usepackage{lastpage}
\jmlrheading{23}{2022}{1-\pageref{LastPage}}{1/21; Revised 5/22}{9/22}{21-0000}{Alan N. Amin and Eli N. Weinstein and Debora S. Marks}

\ShortHeadings{Kernels for Biological Sequences}{Amin, Weinstein and Marks}
\firstpageno{1}

\usepackage{xcolor}

\begin{document}

\title{Biological Sequence Kernels with Guaranteed Flexibility}

\author{\name Alan N. Amin\\
\AND
\name Eli N. Weinstein*\\
\AND
\name Debora S. Marks*}
\author{\name Alan N. Amin \email alanamin@g.harvard.edu \\
      \addr Harvard Medical School
      \AND
      \name Eli N. Weinstein\thanks{Equal contribution.} \email ew2760@columbia.edu \\
      \addr Columbia University
      \AND
      \name Debora S. Marks$^*$ \email debbie@hms.harvard.edu\\
      \addr Harvard Medical School\\
      Broad Institute of Harvard and MIT}

\editor{}

\maketitle

\begin{abstract}

Applying machine learning to biological sequences---DNA, RNA and protein---has enormous potential to advance human health, environmental sustainability, and fundamental biological understanding. However, many existing machine learning methods are ineffective or unreliable in this problem domain.
We study these challenges theoretically, through the lens of kernels.
Methods based on kernels are ubiquitous: they are used to predict molecular phenotypes, design novel proteins, compare sequence distributions, and more. 
Many methods that do not use kernels explicitly still rely on them implicitly, including a wide variety of both deep learning and physics-based techniques.
While kernels for other types of data are well-studied theoretically, the structure of biological sequence space (discrete, variable length sequences), as well as biological notions of sequence similarity, present unique mathematical challenges.
We formally analyze how well kernels for biological sequences can approximate arbitrary functions on sequence space and how well they can distinguish different sequence distributions. In particular, we establish conditions under which biological sequence kernels are universal, characteristic and metrize the space of distributions.
We show that a large number of existing kernel-based machine learning methods for biological sequences fail to meet our conditions and can as a consequence fail severely.
We develop straightforward and computationally tractable ways of modifying existing kernels to satisfy our conditions, imbuing them with strong guarantees on accuracy and reliability.
Our proof techniques build on and extend the theory of kernels with discrete masses.
We illustrate our theoretical results in simulation and on real biological data sets.
\end{abstract}


\section{Introduction} \label{sec:intro}

Consider the following machine learning problem for biological sequences.
A scientist wants to know the relationship between a protein's sequence and its fluorescent color.
They synthesize a large library of different sequences, and measure the color of each one. Their aim is to predict color from sequence.
One popular machine learning approach they might use is a Gaussian process.
Gaussian processes are flexible and can represent uncertainty, making them especially helpful if the scientist wants to design new proteins with a specific color.

To build a Gaussian process on biological sequences, one must specify a kernel $k(X, Y)$ over biological sequences $X, Y$.
Many possible options have been proposed, which depend on different notions of sequence similarity.
One approach is to compare the two sequences position by position, using a similarity matrix based on amino acids' biophysical properties~\citep{Schweikert_undated-qx,Toussaint2010-qn}.
Another is to compare the frequency of sub-string (kmer) occurrences in each sequence~\citep{Leslie2002-px}.
Still another is to score sequence similarity based on all possible alignments between them~\citep{Haussler1999ConvolutionKO}.
Yet another is to compare the sequences based on the embeddings produced by a deep neural network~\citep{Yang2018-ao,Alley2019-yy}.
All these approaches have seen some empirical success.

Unfortunately, as we will show, for all these different kernels the Gaussian process model can fail entirely: even if the scientist reduces their measurement error to zero, and scales up their experiment to collect infinite data on infinite sequences, with a whole rainbow of different colors, the trained model may predict that every single sequence (whether its in the training data or not) yields a fluorescent purple protein with high confidence.
The fundamental problem is that each of these biological sequence kernels can fail to be \textit{universal}: there are some functions relating sequences to their properties that they simply cannot describe.
As a consequence, methods based on these kernels can be unreliable, working well for some example problems but failing, sometimes spectacularly, on new problems.

This paper is about designing new kernels for biological sequences that possess strong theoretical guarantees against unreliability. 
In particular, we study kernel expressiveness, the question of what functions the kernel can and cannot describe (or, more technically, what functions are in the kernel's associated Hilbert space).
Our aim is to develop biological sequence kernels that are sufficiently expressive to ensure that the machine learning methods which use them are guaranteed to work reliably across many different problems.

We focus on three key measures of kernel expressiveness: universality, characteristicness, and metrizing the space of distributions. Here, we briefly introduce these concepts and explain their practical relevance. 

In supervised learning problems, the aim is to regress from biological sequences to another variable, such as a measurement of a phenotype. Constructing maps from genotype to phenotype is valuable for many areas of biology, biomedicine and bioengineering, whether one is inferring what ancient microbes ate, diagnosing genetic disease, designing sustainable catalysts, etc.
One approach to supervised learning is to use parametric models like linear regression, but they are typically unlikely to capture the true sequence-to-phenotype relationship.
A more flexible approach is to use a kernel-based regression method, such as a Gaussian process or support vector machine.
However, kernel methods are not necessarily as flexible as one would hope.
To ensure reliable inferences, we want kernels that are \textit{universal}, meaning they can capture any function on sequences.
We show that many existing biological sequence kernels are not universal; we also provide straightforward modifications that make them universal.
Supervised learning methods that use our proposed universal kernels can, provably, describe any genotype-to-phenotype map, regardless of its complexity.

In unsupervised learning problems, the aim is to learn a distribution over sequences.
Estimating sequence distributions is valuable for many areas of biology, biomedicine and bioengineering, whether reconstructing ancient immune systems, forecasting future viral mutations, designing humanized antibody therapeutics, etc.
When building unsupervised generative models, a key question is whether the distribution of the model actually matches that of the data.
One approach to comparing sequence distributions is to use summary statistics, and ask, for example, whether the two distributions match in terms of sequence length, hydrophobicity, predicted structure, etc.
However, even if the distributions match along some selected dimensions, they may be very different in other ways.
A more general comparison approach is to use a kernel-based two-sample or goodness-of-fit test, such as maximum mean discrepancy (MMD)~\citep{Gretton2012-wp}.
To ensure reliable inferences, we want kernels that are \textit{characteristic}, meaning that MMD is zero if and only if the two distributions we are comparing match exactly.
We show that existing kernels are, in general, not characteristic; we also provide straightforward modifications that make them characteristic.
A discrepancy that uses one of these modified kernels can, provably, detect any difference between two distributions over biological sequences, regardless of its subtlety.

Kernel-based methods are useful not only for evaluating generative models, but also for learning generative models in the first place. In particular, one can train an unsupervised model by minimizing a kernel-based discrepancy.
Such methods are especially useful for models with analytically intractable likelihoods (also known as implicit models), and have been applied to generative adversarial networks (GANs), approximate Bayesian computation (ABC), etc.~\citep{Li2017-kn,Park2016-yw}.
They are also useful for models that do not have support over the entire data space, which has proven useful for experimental design and quadrature~\citep{Chen2010-ry,Huszar2012-ab,Bach2012-ed,Pronzato2021-mr}. 
The key challenge here is that, as we optimize an MMD loss to learn our model, we want the model distribution to approach the target (data) distribution.
This might not happen, even if the kernel is characteristic.
Success depends on whether the kernel can \textit{metrize the space of distributions}.
We show that existing biological sequence kernels generally lack this property, and propose modified kernels that have it.
Minimizing the MMD with our proposed kernels will, provably, yield a model distribution that matches the target distribution exactly.

In many applications, practitioners use neural networks instead of or in addition to kernels.
For example, one common technique for predicting sequence properties is based on semi-supervised learning: first, learn a map from sequence to representation using large scale unlabeled data and a deep neural network; then, learn a map from representation to outcome using small scale labeled data and a Gaussian process~\citep{Yang2018-ao}. 
By composing the map from sequence to representation with the Gaussian process's kernel over representations, we can understand this approach as employing a kernel over sequences. 
We analyze such kernels in Section~\ref{sec: embedding}.
More broadly, there are many close connections between kernels and neural networks, and kernels are a powerful and commonly-used tool for studying deep learning methods theoretically~\citep{Neal1996-xn,Jacot2018-hn,De_G_Matthews2018-pi,Lee2017-gc,Simon2022-hu}. Our results thus also provide a starting point for theoretical analysis of deep neural networks applied to biological sequences.

Studying kernels on biological sequence space presents unique theoretical challenges.
Biological sequences are strings of characters (nucleotides or amino acids), and typically vary in length across a data set. Thus, following previous theoretical work on biological sequences, we take sequence space to be the set of all finite-length strings $\cup_{L=0}^\infty \B^L$ of an alphabet $\mathcal{B}$, where e.g. $\mathcal{B} = \{A, T, G, C\}$ for DNA~\citep{Amin2021-sk,Weinstein2022-yh}.
Biological sequence space is therefore discrete and infinite, while the vast majority of previous theoretical and empirical studies on kernels have been done on continuous or finite spaces.
To our knowledge, the only previous work on kernel flexibility guarantees in infinite discrete spaces is that of \citet{Jorgensen2015-na}, who only study a handful of kernels that are of limited use for biological sequences.

We study the flexibility of practical and popular kernels for biological sequences systematically.
Our approach centers on determining whether or not a kernel \textit{has discrete masses}, i.e. whether its associated Hilbert space contains delta functions~\citep{Jorgensen2015-na}.
We show that this property is sufficient to guarantee the kernel is universal, characteristic, and metrizes the space of probability distributions.
Since continuous kernels on Euclidean space cannot have discrete masses, our theoretical approach is unique to discrete spaces.
We explain how a wide range of combinations and manipulations of kernels preserve the discrete mass property,
and apply this theory to propose modifications to popular kernels that imbue them with discrete masses.
We thus provide a powerful new set of tools for the design and construction of biological sequence kernels with strong guarantees.

The layout of the paper is as follows.
Section \ref{sec: notation} sets up notation.
Section \ref{sec: toy example} gives a toy, motivating example that illustrates our results and how they can be used in practice.
Section \ref{sec: related work} reviews related work.
The next set of sections develop the key theoretical machinery we use to prove kernel flexibility.
In Section \ref{sec: delta func and univ} we introduce the various notions of kernel flexibility that we are interested in: universality, characteristicness, and metrizing the space of distributions. We also introduce the notion of kernels with discrete masses and prove that it implies all three.
Section \ref{sec: delta char and manip} develops tools for proving that kernels have discrete masses, in particular describing transformations that preserve the discrete mass property.
With these foundation in hand, we study a variety of kernels for biological sequences that are used in practice; for each, we find that common kernel choices do not have the properties we want, and develop alternatives that do. 
Section \ref{sec: hamming kernels} investigates kernels that compare sequences position-by-position, including kernels that use the Hamming distance.
Section \ref{sec: string kernels} investigates alignment kernels, which compare sequences by considering all possible alignments between them.
Section \ref{sec: spectra} investigates kmer spectrum kernels, which compare sequences based on their set of sub-strings (kmers).
Section \ref{sec: embedding} investigates kernels built by embedding sequences into Euclidean space.
We then illustrate our results, and the improved empirical performance of our proposed kernels, on both synthetic and real data in Section \ref{sec: experiments}. 
Section \ref{sec: discussion} concludes.

\section{Notation}\label{sec: notation}
In this section we establish notation for reference in the rest of the text.

We let $S$ be a countably infinite set with the discrete topology. 
In sections \ref{sec: delta func and univ} and \ref{sec: delta char and manip} where we study guarantees for kernels on arbitrary infinite discrete spaces, $S$ can be any countable infinite set;
in the rest of the paper we will specifically be interested in the case when $S$ is the space of sequences (defined below).
For any finite set $B$ we will define $|B|$ as its cardinality.
We let $\mathbb N$ be the set of natural numbers $\{0, 1, 2, \dots\}$.
We also define $\mathbbm{1}(P)$ to be the indicator function, which is $1$ if $P$ is true and $0$ if it is false.
If $y, z\in\mathbb R$, we define $y\vee z$ as their maximum and $y\wedge z$ as their minimum.

For $X\in S$, by $\delta_X$ we either mean the function or measure that is $1$ on $X$ and $0$ everywhere else.
We call $C_0(S)$ the space of functions on $S$ that vanish at infinity and $\Vert\cdot\Vert_\infty$ the infinity norm on $C_0(S)$.
We define $C_C(S)$ to be the set of functions on $S$ that are non-zero at only finitely many points.
We define $\Pc(S)$ to be the space of probability distributions on $S$.

A kernel is a function $k: S\times S\to \mathbb R$ such that for any finite set of $Y_1, \dots, Y_N\in S$ and $\alpha_1, \dots, \alpha_N\in \mathbb{R}$, $\sum_{n, m}\alpha_n\alpha_mk(Y_n, Y_m)\geq0$.
We define the inner product $(\cdot|\cdot)_k$ on linear combinations of functions $k_Y = k(Y, \cdot)$ so that $(\sum_{n=1}^N\alpha_nk_{X_n}|\sum_{m=1}^M\beta_mk_{Y_m})_k=\sum_{n, m}\alpha_n\beta_mk(X_n, Y_m)$ for
$X_n, Y_m\in S$ and $\alpha_n, \beta_m\in \mathbb{R}$.
We denote the associated norm as $\Vert\cdot\Vert_k$.
We say that a kernel is strictly positive definite if $ \sum_{n=1}^N\alpha_nk_{X_n}\neq 0$ when the $X_n$ are distinct and $\alpha_n$ are non-zero.
We define the reproducing kernel Hilbert space (RKHS) of the kernel $\Hc_k$ as the Hilbert space completion using the inner product $(\cdot|\cdot)_k$.
We can write every $f\in\Hc_k$ as a function on $S$ given by $f(X)=(f|k_X)_k$.
For a set of vectors $\{v_\lambda\}_{\lambda}\subset \Hc_k$, we define
$\text{span}\{v_\lambda\}_\lambda$ as the set of \textit{finite} linear combinations of elements of $\{v_\lambda\}_{\lambda}$.

Say $\mu$ is a signed measure on $S$ such that $\int d|\mu|(X) \sqrt{k(X, X)}<\infty$.
Then there is a $f\in\Hc_k$ such that for all $g\in \Hc_k$ we have $(f|g)_k=\int d\mu(X)g(X)$
and $\Vert f\Vert_k^2=\int d\mu(X)\int d\mu(Y) k(X, Y)$.
This element of $\Hc_k$ is called the ``kernel embedding'' of the measure $\mu$, denoted $\int d\mu(X) k_X$.

Let $\B$ be a finite set, the ``alphabet'' (for example, this would be the four nucleotides for DNA, or the twenty amino acids for proteins).
We define the set of sequences as $\cup_{L=0}^\infty \B^L$ where $\B^0$ is defined to be the set containing just the sequence of length zero, $\emptyset$.
For a sequence $X$, and numbers $l, l'$, we define $X_{(:l)}$ as the first $l$ letters in $X$, $X_{(l:l')}$ as the $l'-l$ letters after the first $l$ letters and $X_{(l:)}$ as all letters after the first $l$ letters, which is potentially the empty sequence $\emptyset$.
We call $X_{(l)}$ the $l$-th letter of $X$, starting counting at $0$.
The concatenation of sequences $X, Y$ is denoted $X+Y$.
For any number $l$, the sequence $X$ concatenated to itself $l$ times is denoted $l\times X$.
We call $d_H(X, Y)$ the Hamming distance between the sequences $X$ and $Y$, that is the total number of mismatches between the two sequences after they have been padded with an infinite tail of stop symbols $\$$.

\section{Illustrative example}\label{sec: toy example}

Before describing our theoretical results in full generality, we first illustrate their practical use via an example.
We consider the problem of sequence regression, where the aim is to predict a sequence's phenotype, e.g. fluorescent color, enzymatic activity, binding strength, etc.
We show that kernel regression using a classic biological sequence kernel, a \textit{Hamming kernel of lag $L$} (which is an instance of a \textit{weighted degree kernel}) can fail entirely to fit the true sequence-to-phenotype map. We then introduce a small but non-obvious modification to the kernel, which our theory guarantees will allow kernel regression to succeed.

Hamming kernels compare sequences based on a sliding window, counting the number of times the subsequence (kmer) in the window matches exactly~\citep{Schweikert_undated-qx}.
More precisely, we consider a Hamming kernel that uses features $\Phi_{l, V}(X)=\mathbbm{1}(X_{(l:l+L)}=V)$ which indicate if the $L$-mer at position $l$ of $X$ is $V$.
The kernel is then $k_{\mathrm{H}}(X, Y)=(\Phi(X)|\Phi(Y)) = \sum_{l=0}^\infty \sum_{V \in \mathcal{B}^L} \Phi_{l, V}(X) \Phi_{l, V}(Y) $ for sequences $X,Y$, i.e. it counts the number of $L$-mers found in the same position in $X$ and $Y$.
Note that in the case where $L=1$, the representation $\Phi(X)$ reduces to the extremely popular \textit{one-hot-encoding} representation of $X$, and the kernel takes the form $k_{\mathrm{H}}(X, Y) = |X|\vee|Y|-d_H(X, Y)$, where $d_H$ is the Hamming distance.

This kernel, however, might not be able to describe the true relationship between sequences and the phenotype that the scientist is interested in.
The reason is that the kernel is not \textit{universal}, and so has limited flexibility.
To see this, we will show that functions $f$ in $\Hc_k$ are restricted to only take certain values. In particular, the values that a function $f$ takes on set of sequences $A_1$ can be restrained by the values it takes on a subset $A_2 \subset A_1$. 
The problem arises as soon as the length of sequences in the data set exceed the length of the sliding window in the kernel, $L$.
Define $A_1=\{X\}_{|X|=L+1}$ to be the set of all sequences of length $L+1$.
Define $A_2=\{X+X_{(0)}\}_{|X|=L}$ to be the set of those sequences of length $L+1$ that end with the same letter they started with (namely, $X_{(0)}$).
Now, when performing kernel regression, we are fitting a function $f = \sum_{n=1}^N\alpha_nk_{\mathrm{H}}(Y_n, \cdot)$. We find that $f$ must satisfy the constraint that its average value on $A_1$ equals its average value on $A_2$,
\begin{equation}\label{eq: degen identity}
\begin{split}
  \frac 1 {|A_1|}\sum_{X\in A_1} f(X) = \sum_{n=1}^N &\alpha_n \left(\Phi(Y_n) \,\Big| \, \frac{1}{|A_1|}\sum_{X \in A_1} \Phi(X)\right)\\ & =  \sum_{n=1}^N \alpha_n \left(\Phi(Y_n) \,\Big| \, \frac{1}{|A_2|}\sum_{X \in A_2} \Phi(X)\right) = \frac 1 {|A_2|}\sum_{X\in A_2} f(X).  
\end{split}
\end{equation}
The reason is that for every $L$-mer $V \in \mathcal{B}^L$ and starting position $l \in \{0, 1\}$, the proportion of sequences that have $V$ at position $l$ is the same in $A_2$ as it is in $A_1$, namely $|\B|^{-L}$. Thus, $\frac 1 {|A_1|}\sum_{X\in A_1} \Phi(X)_{l,V} = |\B|^{-L} = \frac 1 {|A_2|}\sum_{X\in A_2} \Phi(X)_{l,V}$ for all $l,V$, yielding Eqn.~\ref{eq: degen identity}.
The implication of Eqn.~\ref{eq: degen identity} is that functions in the RKHS are constrained, and cannot describe any possible relationship between sequences and their phenotypes; in other words, it says that the kernel is not universal.
Note that this argument also generalizes to more complex versions of the Hamming kernel, such as those that consider all kmers up to length $L$, or those that score kmer similarity based on the biophysical properties of amino acids instead of exact matches~\citep{Schweikert_undated-qx,Toussaint2010-qn}. 

The limited flexibility of the Hamming kernel can lead to serious practical failures, where the model is not just slightly wrong on some data points but instead makes terrible predictions for all data points.
Say a scientist has taken measurements of all sequences of length $L+1$, obtaining a data set $\{g(X): X \in A_1\}$. They find that sequences $X \in A_2$ that start and end with the same letter have a score of $g(X) = |\B|-1$, and those that do not (that is, $X \in A_1 \setminus A_2$) have a score of $g(X) = -1$. We assume there is no measurement error, i.e. $g(X)$ is the ground truth value of the phenotype.
The least-squares fit for this big, clean data set, using kernel regression with the Hamming kernel, turns out to be awful: $f(X) = 0$ for all $X$. To see this, note that the least squares objective is minimized at $\alpha_1, \ldots, \alpha_N = 0$,
\begin{equation*}
\begin{aligned}
\frac{d}{d \alpha} &\sum_{X\in A_1} \frac{1}{2}(\alpha k_{\mathrm{H}}(Y, X)-g(X))^2 \Big|_{\alpha=0}=-\sum_{X\in A_1}k_{\mathrm{H}}(Y, X)g(X)\\
&=\sum_{X\in A_1} k_{\mathrm{H}}(Y, X) - |\B|\sum_{X\in A_2} k_{\mathrm{H}}(Y, X)\\
&= |A_1| \left(\frac 1 {|A_1|}\sum_{X\in A_1} k_{\mathrm{H}}(Y, X) - \frac 1 {|A_2|}\sum_{X\in A_2} k_{\mathrm{H}}(Y, X)\right)=0.
\end{aligned}
\end{equation*}
The lack of kernel flexibility is thus a serious practical concern, as it can lead to terrible model performance even when we have large amounts of clean data.

Our theoretical results in Section~\ref{sec: hamming kernels} will provide a straightforward solution, in the form of a modified kernel: the inverse-multiquadratic Hamming (IMQ-H) kernel,
$$k_{\mathrm{IMQ-H}}(X, Y) = \frac{1}{(1+|X|\vee|Y|-(\Phi(X)|\Phi(Y)))^2}.$$
This kernel is just as tractable to compute as the original Hamming kernel, $k_{\mathrm{H}}(X, Y) = (\Phi(X)|\Phi(Y))$. Like the original kernel, it treats sequences that have the same kmers in the same positions as similar to one another. 
Thus, if there is a biological reason to use a Hamming kernel for a specific problem, they same justification likely holds for a IMQ-H.\footnote{For instance, if we are examining promoters, whose biological activity depends on the arrangement of transcription factor binding sites, we have good biological reasons to believe that sequences with the same kmers in the same positions will have similar function, as transcription factors typically bind conserved sites of roughly $8-10$ nucleotides. In this context, it makes sense to use  $(\Phi(X)|\Phi(Y))$ as a similarity score, but the biology offers no particular reason to prefer $k_{\mathrm{H}}$ over $k_{\mathrm{IMQ-H}}$.}
However, we will prove that the IMQ-H has a fundamental advantage: it has discrete masses, which means that it is universal.
If we perform a regression using the IMQ-H instead of the original Hamming kernel, we are guaranteed to be able to describe any sequence-to-property relationship.

\begin{figure}
\caption{Least squares regression with Hamming and IMQ-H kernels. $N$ is the size of the data set. Performance is measured by root mean squared error, normalized by the standard deviation of the outcome variable.}\label{fig: toy example}
\centering
\includegraphics[width=0.45\textwidth]{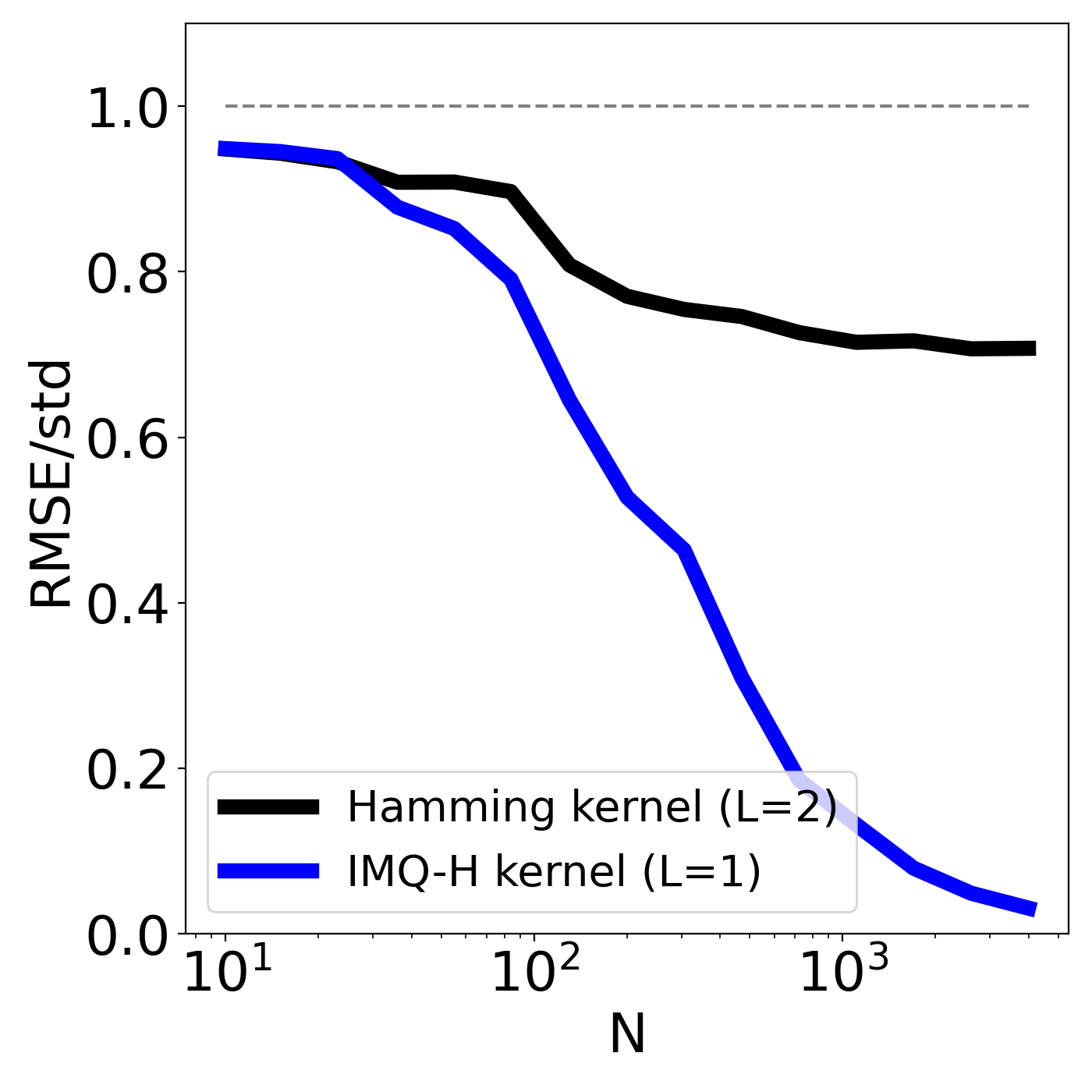}
\end{figure}

We illustrate the benefits of our IMQ-H kernel in simulation.
As the ground truth sequence-to-phenotype map, we consider
$$g(X) = \max_b \sum_{l=0}^\infty \mathbbm{1}(X_l=b).$$
which counts the number of occurrences of the most common letter in the sequence. 
We draw data points $X_1, X_2, \ldots$ from a uniform distribution on DNA sequences of length 4, and record noiseless observations of $g(X_n)$ for each.
We estimate $g$ using least-squares kernel regression, and compare two different kernels: the Hamming kernel with $L=2$ and the IMQ-H kernel with $L=1$.
Naively, one might expect the IMQ-H kernel to perform more poorly, as it uses a smaller sliding window.
In fact, however, we find that regression with the Hamming kernel asymptotes at a high error as we collect more and more data, while with the IMQ-H the error converges to zero (Figure~\ref{fig: toy example}).
Our theoretical guarantees can thus lead to dramatic performance gains at negligible computational cost.

In the following sections, we extend the ideas in this illustrative example systematically.
We describe theoretical guarantees that are practically relevant to a wide range of problems, well beyond regression. 
We describe serious flaws in commonly used kernels, besides the Hamming kernel. 
And, we describe alternative kernels that capture the same biological ideas as the original flawed kernels, and are just as computationally tractable, but possess strong theoretical guarantees.  

\section{Related work}\label{sec: related work}
\subsection{Theoretical guarantees for kernels}

There is a large body of theoretical work studying kernel flexibility, much of it focused on the properties of universality, characteristicness, and metrizing probability distributions~\citep{Sriperumbudur2009-wh, Sriperumbudur2010-ii, Sriperumbudur2011-ay, Fukumizu2008-iw, Christmann2010-ox}.
These results have been foundational in providing strong guarantees on a wide variety of kernel methods, including not only regression methods but also two-sample tests~\citep{Gretton2012-wp}, independence tests~\citep{Gretton2007-gu}, sample quality evaluation~\citep{Gorham2017-rt}, learning implicit models~\citep{Li2017-kn}, learning models with intractable normalizing constants~\citep{Dai2019-ei,Matsubara2022-qz}, causal inference~\citep{Singh2021-eq}, Bayesian model selection and data selection~\citep{Weinstein2023-rd} and much more.
However, virtually all existing flexibility guarantees are for kernels on continuous space.
Thus, when it comes to applying these powerful kernel methods to infinite discrete spaces in general, and biological sequences in particular, theoretical justification is lacking.

The situation is particularly challenging for kernels that are used in practice, which must capture real-world structure (the underlying biology) and be tractable computationally.
Indeed, it is not entirely obvious \textit{a priori} that strong flexibility guarantees can even be established for practical biological sequence kernels: in the case of graphs, for instance, kernels that are characteristic are at least as hard to compute as the graph isomorphism problem, so no polynomial time algorithm exists~\citep{Kriege2020-xc,Gartner2003-jy}. In other words, tractability and flexibility are not necessarily compatible.

\citet{Jorgensen2015-na} established pioneering results on kernel flexibility for infinite discrete spaces, exploring the idea of kernels with discrete masses.
We extend the scope and implication of their results dramatically, by (a) connecting the discrete mass property to the notions of flexibility important for machine learning applications, and (b) studying kernels relevant to biological sequences.
More broadly, we make the theory of \citet{Jorgensen2015-na} more ``user-friendly'', by establishing easier routes to proving the discrete mass property, and elaborating its practical implications.

\subsection{Kernels for biological sequences}

There is a rich and long-running empirical literature on kernels for biological sequences, despite the lack of theoretical guarantees.
We will describe major biological sequence kernel families in detail in later sections; here, we highlight important overall trends and challenges.
First, many kernels---such as Hamming and weighted degree kernels---compare sequences position-by-position, often relying (in the case of proteins) on amino acid similarity measures (Section~\ref{sec: hamming kernels})~\citep{Sonnenburg2007-yr,Toussaint2010-qn}. 
These position-wise comparison kernels also typically rely on data pre-processing, in the form of multiple sequence alignment, which forces the data set to consist of sequences of the same length and allows meaningful comparison across positions. This pre-processing can be unreliable, and limit model generalization to unobserved sequences~\citep{Weinstein2021-ce}.

One alternative that does not depend on alignment pre-processing is to use a kmer spectrum kernel, which featurizes sequences based on their kmer content~\citep{Leslie2002-px}.
Another alternative, which explicitly takes into account common biological mutations (substitutions, insertions and deletions), is to use an alignment kernel, which integrates over all pairwise alignments~\citep{Haussler1999ConvolutionKO}.
Practically, however, the alignment kernel exhibits strong ``diagonal dominance'', such that sequences are, depending on the kernel parameters, all ``very close'' or ``very far'' with nothing in between.
Attempts to fix this problem have been largely unsuccessful; for instance, some proposed methods are not non-negative definite, or violate other key conditions of kernel methods~\citep{Saigo2004-go,Weston2003-sg}.
We study alignment kernels and diagonal dominance in Section~\ref{sec: string kernels}, and kmer spectrum kernels in Section~\ref{sec: spectra}.

More recently, embedding kernels, which take advantage of advances in representation learning using deep neural networks, have gained in popularity~\citep{Yang2018-ao}.
Here, the kernel function of two sequences is defined as a standard Euclidean space kernel (such as a radial basis function kernel) applied to the representation of each sequence.
Embedding kernels can be used with variable length sequences, without alignment pre-processing, and can take advantage of large scale unlabeled data sets to learn meaningful representations.
However, one way these methods can fail is by embedding quite unrelated sequences to nearby points, especially if one of those sequences is outside the training data set. 
Thus in practice, researchers often restrict the sequences they consider to a neighborhood around the training data, limiting the method's scope~\citep{Yang2018-ao, Notin2021-gj, Stanton2022-ed,Detlefsen2022-jl}.
We study embedding kernels and their limitations in Section~\ref{sec: embedding}.

\section{The discrete mass property}\label{sec: delta func and univ}

In this section we introduce the discrete mass property and describe its implications for kernel flexibility.
In particular, we show that if a kernel has discrete masses, it is (1) universal (it can be used to describe arbitrary functions on sequence space), (2) characteristic (it can be used to discriminate between arbitrary sequence distributions) and (3) metrizes the space of distributions (it can be used to optimize one sequence distribution to match another).
One or more of these three properties are typically needed for strong guarantees on any kernel-based machine learning method.
Later, we will design biological sequence kernels that have discrete masses, and thus satisfy these three properties.

We say a kernel has discrete masses if its Hilbert space includes delta functions at all points in the data space $S$. Note that in this section we take $S$ to be an arbitrary discrete space; later (starting in Section~\ref{sec: hamming kernels}) we will specialize to the setting where $S$ is sequence space.
\begin{definition}[The discrete mass property]
    We say $k$ has discrete masses if $C_C(S)\subset \Hc_k$, where $C_C(S)$ is the set of all functions on $S$ that are non-zero at only finitely many points. Since $\Hc_k$ is a linear space, this is equivalent to $\delta_X\in\Hc_k$ for all $X\in S$.
\end{definition}
\noindent \sloppy 
In Section \ref{sec: func approx} we explain how the discrete mass property leads to guarantees on approximating functions and in Section \ref{sec: measure distinguish} we explain how it leads to guarantees on distinguishing distributions.

\subsection{Universal kernels and function approximation}\label{sec: func approx}

When performing regression with a kernel $k$, one fits the data with a member of its reproducing kernel Hilbert space (RKHS), $\Hc_k$. The RKHS is the closure of the set of functions on $S$ of the form $\sum_{n=1}^N \alpha_n k(\cdot, X_n)$, where $X_1, \dots, X_N\in S$ and $\alpha_1, \dots, \alpha_N\in\mathbb R$.
The accuracy of the regression depends crucially on the question of exactly what functions are in $\Hc_k$. Roughly speaking, the larger $\Hc_k$ is, the more accurate and reliable the regression method is in the big data setting.

If the RKHS is large enough to describe any function in some very general set, we say it is universal. In particular, we focus on $C_0$-universality, which says that the RKHS can approximate any function in $C_0$, the set of all functions on $S$ that vanish at infinity~\citep{Sriperumbudur2011-ay}.
\begin{definition}[$C_0$-universality]
    Say $k(X, X)=1$ and $k_X\in C_0(S)$ for all $X\in S$.
    We say $k$ is $C_0$-universal if for every $f\in C_0(S)$ and $\epsilon>0$ there is a $g\in \Hc_k$ such that $\Vert f-g\Vert_\infty<\epsilon$.
\end{definition}
\noindent We will also briefly touch on $L^p$-universality, which is defined in the same way, only replacing $C_0(S)$ with $L^p(S, \mu)$, and replacing the infinity norm with the $p$-norm under any measure $\mu$ on $S$. 

Are common biological sequence kernels universal? Can they describe any genotype-to-phenotype map?
Many are not, because they fail a simple criteria: they do not have infinite features~\citep{Leslie2002-px, Tsuda2002-qb, Jaakkola2000-dk}.
\begin{proposition}[Kernels with finite features, on infinite data spaces, are not universal] \label{prop:finite_kernels_not_universal}
    Consider a kernel $k(X, Y) = (\phi(X) \mid \phi(Y))$  defined using a feature vector $\phi(X)$ of finite length. If $S$ is infinite, the kernel is not $C_0$-universal (or $L^p$-universal).
\end{proposition}
\begin{proof}
In this case $\Hc_k$ is finite dimensional, as the RKHS $\Hc_k$ is isomorphic to $\mathbb{R}^d$, where $d$ is the length of the feature vector.
If the data space $S$ is infinite, then the function space $C_0(S)$ is infinite-dimensional, and likewise the function space $L^p(\mu)$ for any $p, \mu$. Thus, $\Hc_k$ cannot be dense in $C_0(S)$ or $L^p(\mu)$; every function in $\Hc_k$ is determined by its values on $d$ data points.
\end{proof}
To illustrate, we apply this proposition to show that a popular biological sequence kernel, the kmer spectrum kernel, is not universal.
\begin{example}[The kmer spectrum kernel is not universal]\label{ex: spectrum kernels}
    \sloppy The $L$-mer spectrum kernel uses as features the number of times each subsequence (kmer) up to length $L$ occurs in a sequence~\citep{Leslie2002-px}. It is defined as $k(X, Y) = (\Phi(X)|\Phi(Y))$ where $\Phi(X)$ is a vector with entries $\Phi(X)_V$ specifying the number of times the kmer $V$ appears in $X$, for all $V \in S$ such that $|V|\leq L$.
    Since the total number of features, $\sum_{l=1}^L |\mathcal{B}|^l$, is finite, it is not universal (Proposition~\ref{prop:finite_kernels_not_universal}).
\end{example}
\noindent Note the same result also holds for variants of the spectrum kernel that count the number of times each kmer $V$  occurs in $X$ while allowing for mismatches~\citep{Leslie2004-ty, Kuang2004-gh}.

Although infinite features are necessary for a kernel to be universal, they are not sufficient.
In Section~\ref{sec: toy example} we gave a proof that the Hamming kernel, which has infinite features, is not universal.
Instead, to prove our proposed new biological sequence kernels are universal, we will prove that they have discrete masses.
\begin{proposition}[Kernels with discrete masses are universal]
    Kernels with discrete masses are $C_0$-universal (and $L^p$-universal).
\end{proposition}
\begin{proof}
    $C_C(S)$ is dense in $C_0(S)$, and in $L^p(\mu)$ for any (possibly infinite) measure $\mu$ on $S$ and $\infty>p\geq 1$
\end{proof}
Intuitively, the idea is that the RKHS of a kernel with discrete masses includes delta functions at each sequence, and (by linearity) we can add up those delta functions to approximate any other function arbitrarily well.

\subsection{Characteristic kernels for distribution comparison}\label{sec: measure distinguish}

Besides regression, kernels can also be used to compare distributions.
We can measure the difference between two distributions $\mu$ and $\nu$ using the maximum mean discrepancy (MMD), which is the maximum difference in expected value of a function in the kernel's RKHS~\citep{Gretton2012-wp}.
For ease of exposition, in this section we will also assume that $k(X, X) = 1$ for all $X \in S$.\footnote{This does not result in a loss of generality as we can replace, in the arguments below, $\Pc(S)$ with $\{\mu\in\Pc(S)\ |\ \int d\mu(X)\sqrt{k(X, X)}<\infty\}$.}
\begin{definition}[Maximum mean discrepancy]
Recall the embedding of a measure $\mu$, denoted $\int d\mu(X)k_X$, is the element of $\Hc_k$ that satisfies $(\int d\mu(X)k_X| g)_k = \int d\mu(X) g(X)$ for all $g \in \Hc_k$. The MMD is the norm of the difference in embeddings of two distributions $\mu, \nu$,
\begin{equation}\label{eq: mmd def}
\textsc{MMD}_k(\mu, \nu)=\max_{\{f \in \Hc_k: \Vert f\Vert_k\leq 1\}} E_\mu f - E_\nu f = \left\Vert \int d\mu(X)k_X-\int d\nu(Y) k_Y\right\Vert_k,
\end{equation}
where $\Vert f\Vert_k^2 = (f | f)_k$.
\end{definition}
\noindent Intuitively, the larger the RKHS $\Hc_k$, the more subtle the difference between $\mu$ and $\nu$ that can be detected.
In order for the MMD to be able to tell the difference between \textit{any} two distributions, it must be characteristic.
\begin{definition}[Characteristic kernel]
    We say a kernel $k$ is characteristic if $\mu\mapsto\int d\mu(X)k_X$ is injective on $\Pc(S)$, the space of probability distributions on $S$.
\end{definition}
\noindent Characteristicness implies that $\mathrm{MMD}_k(\mu, \nu)=0$ if and only if $\mu=\nu$.
Practically, as an example, characteristicness ensures that kernel-based two-sample tests and conditional independence tests do not have zero power~\citep{Gretton2012-wp,Fukumizu2007-pl}.

Many popular biological sequence kernels are not characteristic. 
For example, kernels with finitely many features, in addition to not being universal, are also not characteristic.
\begin{proposition}[Kernels with finite features, on infinite data spaces, are not characteristic] \label{prop: in text char finite features}
    Consider a kernel $k(X, Y) = (\phi(X) \mid \phi(Y))$  defined using a feature vector $\phi(X)$ of finite length. If $S$ is infinite, the kernel is not characteristic.
\end{proposition}
\begin{proof}
    See Appendix~\ref{sec: not char with finite features} for the proof.
\end{proof}
As with universality, having an infinite number of features is not enough to make a kernel characteristic. Here is an example.
\begin{example}[The weighted degree kernel is not characteristic]
    Consider the weighted degree kernel from Section~\ref{sec: toy example}, along with the sets $A_1$ and $A_2$. Note this kernel has an infinite number of features.
    Take $\mu = \frac{1}{A_1} \sum_{X \in A_1} \delta_X$ to be the uniform distribution over $A_1$, and let $\nu = \frac{1}{A_2} \sum_{X \in A_2} \delta_X$ be the uniform distribution over $A_2$.
    Then $\mu \neq \nu$ but $\textsc{MMD}_k(\mu, \nu)=0$, since the distribution embeddings satisfy $\int d\mu(X) \Phi(X)_{l,V} = \frac 1 {|A_1|}\sum_{X\in A_1} \Phi(X)_{l,V} = |\B|^{-L} = \frac 1 {|A_2|}\sum_{X\in A_2} \Phi(X)_{l,V} = \int d\nu(X) \Phi(X)_{l,V}$ for all $l,V$.
    As a result, $\textsc{MMD}_k$ is not a reliable method for distinguishing distributions. We illustrate this with simulations in Appendix~\ref{sec:mmd_test_sims}. 
\end{example}

MMD is useful not only for comparing two given distributions $\mu$ and $\nu$, but also as an optimization objective that we can minimize to find a distribution $\nu$ that matches $\mu$.
For example, we could train a model $q_\theta$ by looking for $\mathrm{argmin}_\theta \textsc{MMD}_k(q_\theta, \hat{q})$, where $\hat{q}$ is the empirical distribution of the training data. This is the idea behind, for instance, MMD GANs~\citep{Li2017-kn}. MMD is also used as a training objective in a variety of other contexts, including methods for experimental design, high-dimensional integration, and approximate Bayesian inference~\citep{Pronzato2020-rb,Huszar2012-ab,Futami2019-jc}.

Ideally, a good optimization objective should not only tell us if we have reached the right answer, but also if we are headed in the right direction.
So it is not enough for MMD to tell us if our approximation $\nu$ matches the target distribution $\mu$; it must also be that as we reduce the MMD, by finding a sequence of distributions $\nu_1, \nu_2, \ldots$ with smaller and smaller $\textsc{MMD}_k(\mu, \nu_n)$, we get better and better approximations to $\mu$.
For this to hold, the kernel must metrize the space of distributions $\mathcal{P}(S)$.
\begin{definition}[Metrizing $\mathcal{P}(S)$]
	We say a kernel $k$ metrizes $\Pc(S)$ if for every $\mu\in\Pc(S)$ and sequence $\nu_1, \nu_2, \dots\in\Pc(S)$, it holds that $\mathrm{MMD}_k(\mu,\nu_n)\to 0$ implies $\nu_n\to \mu$.
 \end{definition}
 \noindent Note, for kernels on coninuous space, the analogous property is sometimes called ``metrizing weak convergence''~\citep{Sriperumbudur2010-ii}. On $\Pc(S)$, however, the weak topology is equivalent to the total variation topology (convergence in total variation implies weak convergence, and if $\nu_n(X)\to\mu(X)$ for all $X\in S$, then by Scheffé's lemma, $\sum_X|\nu_n(X)-\mu(X)|\to 0$).
In short, kernels that metrize $\mathcal{P}(S)$ allow us to reliably use MMD as an optimization objective, while kernels that do not metrize $\mathcal{P}(S)$ can lead to arbitrarily bad solutions.

For optimizing biological sequences, representation learning methods are a popular choice in practice, since low dimensional continuous optimization can be more tractable than high dimensional discrete optimization~\citep{Yang2018-ao,Stanton2022-ed,Notin2021-gj}.
However, even very faithful representations, that preserve large amounts of information about sequence space, may not be able to metrize $\mathcal{P}(S)$. Embedding kernels based on representations can thus be unreliable for optimizing sequence distributions.
\begin{example}[Embedding kernels are not guaranteed to metrize $\mathcal{P}(S)$] \label{ex:embedding_metrize}
    A scientist is interested in a distribution $p \in \mathcal{P}(S)$ over biological sequences, for instance, a distribution over proteins that are predicted by a generative model to fluoresce yellow. 
    They want to choose a sequence that is representative of $p$ to synthesize and test in the laboratory. 
    One approach is to choose $\mathrm{argmin}_{X \in S} \textsc{MMD}_k(p, \delta_X)$, the sequence that is closest to $p$ as measured by MMD.
    To define a kernel, the scientist uses an embedding $F: S \to \mathbb{R}$ that is injective, such that every sequence receives a distinct representation.
    
   To illustrate what can go wrong, consider the simple case where we have an alphabet $\mathcal{B}$ of just one letter, $\mathcal{B} = \{A\}$, and sequence space is $S = \{\emptyset, A, AA, \ldots\}$.
   We take the target distribution to be $p = \delta_A$, a point mass at the length one sequence $A$. 
   To define a kernel, we consider the injective embedding $F(A) = 0$ and $F(n \times A) = 1/n$ for $n > 1$, and use a radial basis function kernel in the embedding space, such that the complete embedding kernel is $k(X, Y) = \exp(-(F(X)-F(Y))^2)$.
   Intuitively, since the radial basis function kernel is characteristic over $\mathbb{R}$, and since the embedding $F$ is one-to-one, the embedding kernel is characteristic (we prove this rigorously in Section \ref{sec: embedding}).

   However, this embedding kernel does not metrize $\mathcal{P}(S)$. If we try to minimize $\textsc{MMD}_k(\delta_A, \delta_X)$ with respect to $X$, we find that choosing longer and longer sequences brings the objective closer and closer to zero, $\textsc{MMD}(\delta_A, \delta_{n\times A})\to 0$ as $n\to\infty$.
   Thus, even though there is a choice of sequence, $X = A$, such that the approximation $\delta_X$ exactly matches the target $p=\delta_A$, optimizing the MMD will not not lead us to the correct answer.
\end{example}

To design new biological sequence kernels that are guaranteed to be characteristic and to metrize $\mathcal{P}(S)$, we again turn to discrete masses.
\begin{proposition}[Kernels with discrete masses are characteristic and metrize $\mathcal{P}(S)$]
    Say $k$ is a kernel such that $k(X, X)=1$ and $k_X\in C_0(S)$ for all $X\in S$.
	$$k \text{ has discrete masses }\implies \mathrm{MMD}_k\text{ metrizes }\Pc(S)\implies k\text{ is characteristic}.$$
\end{proposition}
\begin{proof}
    The second implication is clear.
	Now say $k$ has discrete masses and $\mu, \nu_1, \nu_2, \dots\in\Pc(S)$ such that $\mathrm{MMD}_k(\mu, \nu_n)\to 0$.
    For each $X\in S$, $\delta_X, -\delta_X\in\Hc_k$, so by Equation \ref{eq: mmd def}
	we have $\Vert \delta_X\Vert_k^{-1}|\mu(X)-\nu_n(X)|\leq \mathrm{MMD}_k(\mu, \nu_n)\to 0$. Thus, $\nu_n\to \mu$.
\end{proof}
	
\noindent The discrete mass property thus guarantees that a kernel is universal, characteristic and metrizes $\mathcal{P}(S)$, making the kernel a good choice for a wide range of applications.\footnote{ 
Note that the discrete mass property is stronger than all three of these properties, as there exist kernels that are universal and metrize $\Pc(S)$ but do not have discrete masses (Appendix~\ref{sec: mmd but not delta}).}
We will show how to modify existing biological sequence kernels to have discrete masses, and thereby ensure they are universal, characteristic and metrizes $\mathcal{P}(S)$.

\subsection{Degenerate examples}

The remainder of the paper will be concerned with designing biological sequence kernels with discrete masses. 
Before studying complicated kernels, however, we first consider two simple but degenerate kernels with discrete masses, and explain why they are unsatisfactory.

One kernel with discrete masses is the identity kernel.
\begin{example}[The identity kernel has discrete masses] \label{ex:identity_kernel}
Consider a kernel of the form $k(X, Y)=B(X)\delta_X(Y)$ for all $X, Y\in S$, where $B:S\to(0, \infty)$ is a function from sequence space to the positive real numbers.
This kernel has discrete masses, since $B(X)^{-1}k_X=\delta_X$ for all $X$.
The problem is that it leads to poor generalization.
For instance, if we use this kernel in a Gaussian process, its predictions on points outside the training set depend only on the prior.
\end{example}
\noindent As this example demonstrates, the mere fact that a kernel has discrete masses does not mean it is a good modeling choice.
Instead, we want kernels that not only have discrete masses but also capture biological notions of sequence similarity. 
Indeed, we will encounter sequence kernels that have discrete masses but are \textit{diagonal dominant}, i.e. they behave very much like the identity kernel.
In these cases we will be interested in modifying the kernel to be less diagonally dominant while preserving their discrete masses.

Another way to construct kernels with discrete masses is to assume that sequence space is finite rather than infinite.
\begin{proposition}[Strictly positive definite kernels on finite spaces have discrete masses] \label{prop:spd_finite}
Consider a data space $S$ that is finite, $|S| < \infty$. If the kernel $k$ is strictly positive definite then it has discrete masses.
\end{proposition}
\begin{proof}
    Recall that a kernel has discrete masses if for every $X \in S$ there exists some $\alpha$ such that $\delta_X(Y) = \sum_{X' \in S} \alpha_{X'} k(X', Y)$. 
    We can think of $\delta_X$ and $\alpha$ as vectors of length $S$, such that we have the equation $\delta_X = K \alpha$, where $K$ is the Gram matrix with entries $K_{XY} = k(X, Y)$ for all $X, Y \in S$.
    The fact that the kernel is strictly positive definite is equivalent to the fact that its Gram matrix is strictly positive definite, and so $K$ is invertible. 
    Thus a solution $\alpha = K^{-1} \delta_X$ always exists.
\end{proof}
\noindent Focusing on a finite $S$, while tempting from the perspective of theoretical convenience, can lead to unreliable methods in practice.
\begin{example}[Sequence space with bounded length]
If sequence space $S$ includes only sequences $X$ with length less than some maximum $L_{\max}$, that is $S = \cup_{L=0}^{L_{\max}} \mathcal{B}^L$, and the kernel is strictly positive definite, then the kernel has discrete masses.

The problem is that we want our kernel methods to be reliable even as we observe or optimize longer and longer sequences.
For instance, we do not want our methods to fail if new data appears past a pre-specified length scale. 
Even if we choose $L_{\max}$ to be astronomical, so that Proposition~\ref{prop:spd_finite} is technically met, studying infinite sequence space $S = \cup_{L=0}^{\infty} \mathcal{B}^L$ is useful in that it forces us to consider the behavior of our methods as sequences grow in length.
In other words, if the asymptotic behavior of a kernel method in the limit of infinite sequence length is bad, the method is likely to have poor performance on finite sequence lengths as well.
\end{example}

Motivated by these concerns, we aim to construct kernels that have discrete masses on infinite sequence space and that capture biological notions of sequence similarity.

\section{Characterizing and manipulating kernels with discrete masses}\label{sec: delta char and manip}

We have seen that kernels with discrete masses have strong guarantees on their ability to approximate functions and distinguish distributions.
However only a small number of kernels, with limited relevance for biological sequences, have been shown previously to have discrete masses on infinite data spaces \citep{Jorgensen2015-na}.
In this section we develop theoretical tools that can be used to prove that kernels have discrete masses.
Section~\ref{sec: delta char} describes conditions that are equivalent to having discrete masses; Section \ref{sec: operations} describes transformations of kernels that preserve the discrete mass property.
We will later apply these techniques to design new biological sequence kernels with discrete masses.
Note in this section $S$ is still an arbitrary infinite discrete space, not necessarily the space of sequences.

\subsection{Equivalent formulations of the discrete mass property}\label{sec: delta char}
In this section we describe two equivalent formulations of the discrete mass property.

\subsubsection{Conditions on the span of kernel embeddings}

The first formulation puts conditions on $\text{span}\{k_Y\}_{Y\neq X}$ that guarantee $\delta_X \in \Hc_k$.
The intuition is as follows.
Consider some element of the span, $\sum_{Y' \in S}\alpha_{Y'} k_{Y'}\in\text{span}\{k_Y\}_{Y\in S}$.
If $\delta_X\in \Hc_k$ then
$(\delta_X|\sum_{Y' \in S}\alpha_{Y'} k_{Y'})_k=\sum_{Y' \in S}\alpha_{Y'} \delta_X(Y')=\alpha_X$,
that is, we can think of $\delta_X$ as a function that takes every element of $\text{span}\{k_Y\}_{Y\in S}$ to the coefficient in front of $k_X$.
In order for this function to exist, it must be (1) well defined and (2) bounded.
For the function to be well defined, $k_X$ must be linearly independent from $\{k_Y\}_{Y\neq X}$.
For the function to be bounded, it must also be difficult to approximate $k_X$ using elements in $\text{span}\{k_Y\}_{Y\neq X}$.
This intuition can be formalized as follows.
\begin{proposition}\label{prop: hahn banach delta}
    Let $X\in S$, and call $\mathcal K_X$ the closure of $\mathrm{span}\{k_{Y}\}_{Y\neq X}$.
	$\delta_X\in\Hc_k$ if and only if $\Hc_k\neq \mathcal K_X$, or in other words, $k_X\notin \mathcal K_X$.
\end{proposition}
\begin{proof}
	Say $\delta_X\in \Hc_k$.
	If $Y\neq X$, $(\delta_X|k_Y)_k=\delta_X(Y)=0$.
	On the other hand, $(\delta_X|k_X)_k=1$.
	Thus, $k_X\notin \mathcal K_X$.
	
	On the other hand, say $k_X\notin \mathcal K_X$.
    Then the orthogonal compliment of $\mathcal K_X$ is exactly one dimensional.
    Let $\phi:\Hc_k\to\mathbb R$ be the linear function projecting $\Hc_k$ to the orthogonal compliment of $\mathcal K_X$, scaled so that $\phi(k_X)=1$.
    Then $\phi(k_Y)=0$ if $Y\neq X$, so,
    $\phi(k_Y)=\delta_X(Y)$ for all $Y\in S$.
	By the Riesz representation theorem, since $\phi$ is continuous, there is a $f\in\Hc_k$ such that for all $Y\in S$, $f(Y)=(f|k_Y)_k=\phi(k_Y)=\delta_X(Y)$, so $f=\delta_X$.	
\end{proof}
One implication of this result is that continuous kernels on Euclidean space cannot have discrete masses.
\begin{proposition}[Continuous kernels on Euclidean space do not have discrete masses]
    If the kernel $k:\mathbb R^D\times \mathbb R^D\to \mathbb R$ is a continuous function, it does not have discrete masses.
\end{proposition}
\begin{proof}
    Consider a sequence $Y_1, Y_2, \ldots \in \mathbb R^D$ that converges to $X\in \mathbb R^D$. Then
    $$\Vert k_{Y_n}-k_{X}\Vert_k^2=k(X, X)+k(Y_n, Y_n)-2k(X, Y_n)\to 2k(X, X)-2k(X, X)= 0.$$
    In other words, $k_X$ is in the closure of $\mathrm{span}\{k_{Y}\}_{Y\neq X}$. 
\end{proof}
The theory of kernels with discrete masses is thus unique to infinite discrete spaces such as biological sequence space.

\subsubsection{Conditions on the kernel's Gram matrix}
A second formulation of discrete masses comes from pioneering work by \citet{Jorgensen2015-na}.
We give a concise restatement of the proof of their result here.
The basic idea builds off of Proposition~\ref{prop:spd_finite}, which says that on finite discrete spaces a strictly positive definite kernel has an invertible Gram matrix and thus discrete masses. On an infinite space, we consider the "invertibility" of a sequence of Gram matrices, defined on larger and larger finite subsets of $S$, to ensure the kernel has discrete masses.
\begin{theorem}\label{prop: matrix det delta}
    \textbf{\citep{Jorgensen2015-na}}
	Let $X\in S$ and let $k$ be a strictly positive definite kernel on $S$.
	For a finite subset $B\subset S$ with $X\in B$, define the Gram matrix $K_B$ indexed by $B$ such that $K_{B, Y, Y'}=k(Y, Y')$ for $Y, Y'\in B$.
 Let $C_{B,X}=\sqrt{\left(K_B^{-1}\right)_{X, X}}$.
	Then,
	$\delta_X\in \Hc_k$ if and only if $\sup_B C_{B,X}<\infty$ where the supremum is over all finite $B\subset S$.
	In this case, $\Vert\delta_X\Vert_k = \sup_B C_{B,X}$.
\end{theorem}

\begin{proof}
	Define a functional $\phi_X$ on $\mathrm{span}\{k_Y\}_{Y\in S}$ that is $\alpha_X k_X + \sum_{Y \neq X} \alpha_{Y}k_{Y}\mapsto \alpha_X$.
	Note $\phi_X$ is well defined since $k$ is strictly positive definite.
	Now, if $\delta_X\in\Hc_k$, then for all $f\in \mathrm{span}\{k_Y\}_{Y\in S}$,  we have $(\delta_X|f)_k=\phi_X(f)$ so that $\phi_X$ is bounded.
	On the other hand, if $\phi_X$ is bounded, it can be continuously extended to all of $\Hc_k$ and must be equal to $(\cdot|f)_k$ for some $f\in \Hc_k$ by the Riesz representation theorem. Then,
	$f(Y)=(k_Y|f)_k=\phi_X(k_Y)=\delta_X(Y)$, so, $f=\delta_X$.
	We will show that $C_B$ is the norm of $\phi_X$ restricted to $\mathrm{span}\{k_Y\}_{Y\in B}$ and the result will follow.
	
	$\mathrm{span}\{k_Y\}_{Y\in B}$ is a finite dimensional space with $\{k_Y\}_{Y\in B}$ as a basis, and inner product $(v| w)= v^T K_B w$ for $v, w\in \mathbb{R}^B$ in this basis.
	Calling $e_X$ the indicator vector for $X$, $\phi_X(v) = (e_X^TK_B^{-1}|v)$.
	Finally, we see that the square norm of $\phi_X$ is $(e_X^TK_B^{-1}|e_X^TK_B^{-1})=\left(K_B^{-1}\right)_{X, X} = C_B^2$.
\end{proof}

\subsection{Transformations that preserve the discrete mass property}\label{sec: operations}

In this section we describe how to construct new kernels with discrete masses out of existing kernels with discrete masses. This allows us to construct large families of related kernels that all have flexibility guarantees.

\subsubsection{Summing}

First we consider linear combinations of kernels with discrete masses.
Kernels are often summed or integrated~\citep{Duvenaud2013-wr}. For example, to build a regression model that can easily learn about phenomena at multiple length scales, one can sum together kernels with different values of a bandwidth parameter, or integrate over all bandwidths.
The following result says that as long as one kernel in this sum has discrete masses, then so does the entire summed kernel.
\begin{proposition}\label{prop: delta integration}
	Say $\Lambda$ is a measurable space and $(k_\lambda)_{\lambda\in \Lambda}$ is a family of kernels. Assume for any $X, Y\in S$, $\lambda\mapsto k_\lambda(X, Y)$ is measurable.
	Say $\nu$ is a positive, nonzero measure on $\Lambda$ with $\int d\nu(\lambda) k_\lambda(X, X)<\infty$ for all $X\in S$ and $\nu(\{\lambda\ |\ k_\lambda$ has discrete masses$\})>0$, i.e. $\nu$ has positive mass on kernels with discrete masses.\footnote{If $\{\lambda\ |\ k_\lambda$ has discrete masses$\}$ is not measurable, then we instead require that $\nu(C)>0$ for all $C\supset \{\lambda\ |\ k_\lambda$ has discrete masses$\}$.}
	Then, $k_\nu=\int k_\lambda d\nu(\lambda)$ is a kernel on $S$ that has discrete masses.
\end{proposition}

\begin{proof}
We consider the possibility that $k_\nu$ does not have discrete masses and show that this leads to a contradiction.
By Proposition~\ref{prop: hahn banach delta}, there is some $X\in S$ and sequence $\mu_l = \delta_X - \sum_{Y \neq X} \alpha_{l,Y}k_{Y}$, where $\sum_{Y \neq X} \alpha_{l,Y}k_{Y} \in \text{span}\{k_Y\}_{Y \neq X}$, such that $\Vert\int k_{\nu, Z}d\mu_l(Z)\Vert_{k_\nu}\to 0$ as $l \to \infty$. Then,
\begin{equation}
\begin{aligned}
\left\Vert\int k_{\nu, Z}d\mu_l(Z)\right\Vert_{k_\nu}^2=&\int d\mu_l(Z_1)\int d\mu_l(Z_2) k_{\nu}(Z_1, Z_2)\\
=&\int d\nu(\lambda)\int d\mu_l(Z_1)\int d\mu_l(Z_2) k_{\lambda}(Z_1, Z_2)\\
=&\int d\nu(\lambda) \left\Vert\int k_{\lambda, Z}d\mu_l(Z)\right\Vert_{k_\lambda}^2.
\end{aligned}
\end{equation}
Thus, by Fatou's lemma,
$$\int d\nu(\lambda) \liminf_{l\to\infty}\left\Vert\int k_{\lambda, Z}d\mu_{l}(Z)\right\Vert_{k_\lambda}^2=0.$$
In particular, there is a $\lambda$ such that $k_\lambda$ has discrete masses and $\liminf_{l\to\infty}\left\Vert\int k_{\lambda, Z}d\mu_{l}(Z)\right\Vert_{k_\lambda}^2=0$, which is a contradiction. 
\end{proof}
A special case is where we are summing over a set of kernels.
\begin{corollary}
	Let $k_0, k_1, \dots k_N$ be kernels such that $k_0$ has discrete masses
	and $\alpha_0, \dots, \alpha_N >0$.
	Then $\sum_{n=0}^N \alpha_n k_n$ is a kernel with discrete masses.
\end{corollary}

\begin{proof}
Let $\nu = \sum_{n=0}^N \alpha_n\delta_{n}$. Now, $\sum_{n=0}^N \alpha_n k_n=\int d\nu(n)k_n$ has discrete masses by the above proposition.
\end{proof}

\subsubsection{Changing domains}

Next we consider using different kernels over different regions of sequence space.
The following result says that if a kernel has discrete masses over all of $S$, it also has discrete masses when restricted to just one region of $S$. As well, if we have separate kernels with discrete masses over separate orthogonal regions of $S$, we can then combine them to construct a new kernel with discrete masses over all of $S$.
\begin{proposition}\label{cor: ortho decomp delta}
    Say $\{W_V\}_V$ is a collection of disjoint subsets of $S$ such that $S=\cup_{V}W_V$.
    If $k$ has discrete masses, then it also has discrete masses when restricted to any $W_V$.
    On the other hand, if $k$ has discrete masses when restricted to each $W_V$, and $k(X, X') = 0$ for any $X\in W_V, X'\in W_{V'\neq V}$, then $k$ has discrete masses over $S$.
\end{proposition}
\begin{proof}
First assume $k$ has discrete masses over $S$. 
A standard property of kernels, detailed in Proposition~\ref{prop: restrict kern} in Appendix~\ref{sec: trans proofs}, is that the Hilbert space of a kernel restricted to a domain $W_V$, that is $\Hc_{k|_{W_V}}$, is the closure of $\text{span}\{k_Y\}_{Y\in W_V}$ in the original Hilbert space $\Hc_k$.
Now consider any $X \in W_V$.
By Proposition~\ref{prop: hahn banach delta}, $k_X$ is not in the closure of $\text{span}\{k_Y\}_{Y\in S\ |\ Y\neq X}$, so it's also not in the closure of $\text{span}\{k_Y\}_{Y\in W_V\ |\ Y\neq X}$. Thus, applying Proposition~\ref{prop: hahn banach delta} again, $k$ has discrete masses over $W_V$.

   Now consider the case where $k$ has discrete masses when restricted to each $W_V$.
   Assume $\delta_X\not\in\Hc_k$ for some $X\in W_V$; we will show this leads to a contradiction.
   By Proposition~\ref{prop: hahn banach delta}, there exists a sequence of functions $f_1 = \sum_{Y \neq X} \alpha_{1,Y} k_{Y}, f_2 = \sum_{Y \neq X}\alpha_{2,Y} k_{Y}, \dots\in \mathrm{span}\{k_Y\}_{Y\neq X}$ such that $f_m\to k_X$ as $m \to \infty$. Let $f_{V, m}$ be the orthogonal projection of $f_m$ onto $\mathrm{span}\{k_Y\}_{Y\in W_V}$, that is $f_{V, m}= \sum_{Y \neq X}\mathbbm{1}(Y\in W_V)\alpha_{m,Y} k_{Y}$. Note that for every $m$, we have $f_{V, m}\in \mathrm{span}\{k_Y\}_{Y\in W_V\ |\ Y\neq X}$, so $f_{V, m}$ is in the Hilbert space $\Hc_{k|_{W_V}}$ of the kernel restricted to $W_V$. Define $f_{V, m}^\perp = f_m-f_{V, m} = \sum_{Y \neq X}\mathbbm{1}(Y\not \in W_V)\alpha_{m,Y} k_{Y}$.
   Now $(f_{V, m}^\perp|f_{V, m})_k=0$ and $(f_{V, m}^\perp|k_X)_k=0$, so
   $$\Vert f_{V, m} - k_X\Vert_k^2\leq\Vert f_{V, m}^\perp\Vert_k^2 + \Vert f_{V, m} - k_X\Vert_k^2=\Vert f_m - k_X\Vert_k^2 \to 0.$$
    In other words, $k_X$ is in the closure of $\mathrm{span}\{k_Y\}_{Y\in W_V\ |\ Y\neq X}$. This implies that the kernel $k$ restricted to $W_V$ does not have discrete masses, a contradiction.
\end{proof}

\subsubsection{Tensorizing}
We next consider tensorizing kernels, so that they can be applied to pairs of sequences. If we have two kernels $k_1$ and $k_2$ on $S$, the tensorized kernel $k_1 \otimes k_2$ is $k_1 \otimes k_2((X_1, X_2), (Y_1, Y_2)) = k_1(X_1, Y_1)k_2(X_2, Y_2)$ for $X_1, X_2, Y_1, Y_2 \in S$.
Tensorized kernels can be useful, for instance, in determining whether two random variables are independent~\citep{Fukumizu2007-pl}.
Tensorization preserves discrete masses.

\begin{corollary}\label{prop: tensor delta}
	Let $k_1$ and $k_2$ be kernels on $S$ with discrete masses.
	Then $k_1\otimes k_2$ is a kernel on $S^2$ with discrete masses.
\end{corollary}

\begin{proof}
A standard property of kernels, detailed in Proposition~\ref{prop: tensor kern} in Appendix~\ref{sec: trans proofs}, says that if $f_1\in \Hc_{k_1}$ and $f_2\in \Hc_{k_2}$, there is a $f_1\otimes f_2\in \Hc_{k_1\otimes k_2}$ such that
	for $(X_1, X_2)\in S^2$, we have $f_1\otimes f_2(X_1, X_2)=f_1(X_1)f_2(X_2)$.
Say $X, Y\in S$. 
Then $\delta_{(X, Y)}=\delta_X\otimes\delta_Y\in \Hc_{k_1\otimes k_2}$.
\end{proof}

\subsubsection{Tilting}

Next we consider re-weighting kernels to emphasize or de-emphasize certain areas of sequence space. More precisely, we consider \textit{tilting} a kernel $k$ by some function $A: S \to (0, \infty)$ to obtain a new kernel $k^A(X, Y)=A(X)k(X, Y)A(Y)$ for $X, Y\in S$.
One reason to tilt a kernel $k$ is to ``normalize'' it, such that $k^A(X, X)=1$ for all $X\in S$; this corresponds to the tilting function $A(X)=\sqrt{k(X, X)}^{-1}$.
The discrete mass property is preserved after tilting.
\begin{proposition}\label{cor: tilting delta}
	If $k$ is a kernel on $S$ that has discrete masses,
	then $k^A$ has discrete masses.
 \end{proposition}
 \begin{proof}
 A standard property of kernels, detailed in Proposition~\ref{prop: tilt delta} in Appendix~\ref{sec: trans proofs}, is that if we have a function $f(\cdot)$ in the original RKHS, $\Hc_k$, then the function $A(\cdot) f(\cdot) $ is in the RKHS of the tilted kernel, $\Hc_{k^A}$, and has the same norm, $\Vert f \Vert_k = \Vert A f \Vert_{k^A}$. In other words, $f\mapsto Af$ is an isometric isomorphism of $\Hc_k$ to $\Hc_{k^A}$.
 So for any $X\in S$, we have $A \delta_X \in \Hc_{k^A}$. We can always multiply a function in an RKHS by a finite scalar. Thus, $\delta_X=\frac 1 {A(X)}A(\cdot)\delta_X(\cdot)\in\Hc_{k^A}$.
 \end{proof}

\subsubsection{Reparameterizing alphabets} \label{sec:reparameterize}

Finally, we look at a novel transformation of kernels, specific to biological sequences, that involves ``reparameterizing'' the alphabet $\B$. 
The basic idea starts from representing letters in the alphabet $\B$ as one-hot encoded vectors, i.e. we treat each $b \in \B$ as a vector of length $|\B|$ with zeros everywhere except at one position. 
The alphabet $\B$ thus forms a basis of $\mathbb{R}^{\B}$.
However, we may also consider an alternative basis $\tilde \B$ of $\mathbb{R}^{\B}$.
This alternative basis gives rise to an alternative set of sequences $\tilde S = \cup_{L=0}^\infty \tilde{\B}^L$.
By treating $S$ and $\tilde S$ as sets of vectors, there will be a natural way to extend a kernel on $S$ to $\tilde S$.
We will see that the property of discrete masses is invariant to this change in basis: if the kernel has discrete masses over $S$, it also has discrete masses over $\tilde S$, and vice versa. 
(Note in this section and all following sections, $S$ will be the set of sequences, rather than an arbitrary infinite discrete space.)

To be more precise, we must define what it means to apply a sequence kernel to vector encoding of a sequence.
Any sequence $Y$ in $S=\cup_{L=0}^\infty \B^L$ can be represented as a one-hot encoding, which consists of a vector $V \in \mathbb{R}^{|Y| \times \B}$ such that $V_{(l)} \in \mathbb{R}^{\B}$ is the one-hot encoding of the letter $Y_{(l)}$.
If the vector $V\in \mathbb{R}^{L \times \B}$ is a one-hot encoding of a sequence then we define its embedding into $\Hc_k$ as the embedding of the sequence it encodes.
We can write this using the formula
\begin{equation} \label{eqn:vector_alphabet_kernel}
    k_V = \sum_{X \in S: |X|=L}\left( \prod_{l=0}^{L-1}V_{(l), X_{(l)}}\right) k_X.
\end{equation}
Thus if $V$ is a one-hot encoding of $Y$, we recover $k_V = k_Y$.
The kernel has not changed; all we have done is rewrite it to embed vector encoded sequences.

We now apply the same kernel to different sequences that use different encodings, which are not one-hot. 
In particular, let $\tilde \B$ be an alternative basis of $\mathbb{R}^\B$.
Each sequence $\tilde{Y}$ in $\tilde S = \cup_{L=0}^\infty \tilde{\B}^L$ can be represented by a vector encoding $\tilde V \in \mathbb{R}^{|\tilde Y| \times \B}$, where $\tilde {V}_{(l)}$ is the encoding of the letter $\tilde Y_{(l)}$.
We define the kernel embedding of $\tilde Y$ by plugging $\tilde V$ into Equation~\ref{eqn:vector_alphabet_kernel}; it consists of a linear combination of kernel embeddings $k_X$ from each sequence $X \in S$ with $|X| = |\tilde V|$.
We will show that if the kernel $k$ has discrete masses over $S$, and we apply $k$ to $\tilde S$, it will still have discrete masses.
We call this shift from the alphabet $\B$ to $\tilde \B$ a ``reparameterization'' of the kernel's alphabet.

Intuitively, in the case of proteins, we can think of the reparameterized alphabet as a set of amino acid properties, such as mass, charge, etc. 
Each amino acid is a letter in the original alphabet $\B$, while each property is a letter in the reparameterized alphabet $\tilde \B$; since both alphabets form bases of $\mathbb{R}^\B$, each amino acid can be described as a linear combination of properties.
We can analyze the flexibility of a kernel over sequences of amino acids by analyzing the flexibility of the same kernel applied to sequences of amino acid properties. 
This is useful theoretically for studying complex kernels.

\begin{proposition}\label{prop: reparam delta}
Say $k$ is a strictly positive definite kernel on $S$, and $\tilde{\B}$ is a basis of $\mathbb{R}^\B$.
    Then $k$ has discrete masses as a kernel on $\tilde{S}$ if and only if it has discrete masses as a kernel on $S$.
\end{proposition}
\begin{proof}
Both $\B$ and $\tilde \B$ are bases of $\mathbb{R}^{\B}$, so the kernel over $S$ is a reparameterization of the kernel over $\tilde S$ and vice versa. Thus, we only need to show that $k$ has discrete masses as a kernel on $S$ if it has discrete masses as a kernel on $\tilde S$.
    First, note that reparameterizing the alphabet does not change the span of the kernel embedding vectors, $\text{span}(k_V)_{V\in (\tilde{\mathcal{B}})^L}=\text{span}(k_{V'})_{V'\in (\mathcal{B})^L}$, since both $\B$ and $\tilde \B$ are bases of $\mathbb{R}^{\B}$.
    Now, consider some length $L$.
    If $X\in S$ with $|X|=L$, then 
    $k_X = \sum_{Y\in \tilde S\ |\ |Y|=L}\alpha_{X, Y}k_Y$ for some $\alpha_{X, Y}$.
    As $k$ is strictly positive definite, $\{k_X\}_{|X|=L}$ is a linearly independent set, so $\alpha$ is an invertible square matrix with dimension $|\B|^L \times |\B|^L$.
    Let $f_X = \sum_{Y\in \tilde S \ |\ |Y|=L}\left(\alpha^{-1}\right)_{X, Y}\delta_Y$.
    Then $f_X(Z)=0$ for $Z\in S$ with $|Z|\neq L$, and if $|Z| = L$,
    $f_X(Z) = \sum_{Y\in \tilde S\ |\ |Y|=L}\left(\alpha^{-1}\right)_{X, Y}\alpha_{Z, Y}=\delta_X(Z)$.
    Thus $\delta_X\in\Hc_k$.
\end{proof}

Throughout the rest of the paper, for any kernel on $S$ and any $V, W\in\cup_{L=0}^\infty \mathbb{R}^{L \times \mathcal{B}}$ we will write $k(V, W)=(k_V|k_W)_k$ where $k_V$ and $k_W$ are defined by Eqn~\ref{eqn:vector_alphabet_kernel}.

\section{Position-wise comparison kernels}\label{sec: hamming kernels}
In this section, we design kernels with discrete masses that compare sequences position-by-position.
We saw in Sections~\ref{sec: toy example} and \ref{sec: measure distinguish} that the Hamming kernel, the weighted degree kernel, and other related kernels that compare sequences position-by-position are neither universal nor characteristic. 
Here, we develop alternative kernels that capture the same biological ideas but are also highly flexible.

Position-wise sequence comparison is ubiquitous in biology, and has strong biological justification for many problems. 
For instance, a common observation is that nucleotides or amino acids at a specific position have a specific biological function; for instance, the amino acids at a few particular positions may chemically react to form a fluorophore, making the protein fluorescent.
So, when predicting phenotype from sequence, a reasonable measure of sequence similarity is one that compares sequences position-by-position, as is done in the Hamming kernel.
Moreover, a very common form of mutation during evolution is a substitution, which switches one letter for another.
Thus, a position-wise measure of sequence similarity can capture evolutionary distance as well as phenotypic distance, which may be desirable, for instance, when comparing sequence distributions.

Our new kernels preserve existing notions of position-wise sequence similarity. For example, as two sequences differ more and more according to the Hamming kernel, they will also differ more and more according to our proposed replacement for the Hamming kernel. 
What we modify is not the measure of sequence similarity but instead the functional form of the dependence of the kernel on sequence similarity, i.e. how exactly a change in the similarity of $X$ and $Y$ translates into a change in $k(X, Y)$.
For existing kernels, the functional form rarely has strong biological justification, and instead is often motivated solely by convenience.
Our results demonstrate that the functional form in fact matters a great deal for the reliability of kernel methods for biological sequences.

We start by studying a simple kernel that compares sequences position-by-position. We then use this ``base'' kernel to derive many other varieties of position-wise comparison kernels, using the transformations developed in Section~\ref{sec: operations}.
The base position-wise comparison kernel is defined as the product of individual kernels applied to the letters at each position.
\begin{definition}[Base position-wise comparison kernel] \label{def:position-wise}
    We represent each sequence $X \in S$ as terminated by an infinite tail of stop symbols $\$$. 
    Let $k_s$ be a strictly positive definite kernel on letters $\mathcal{B} \cup \{\$\}$ with $k_s(\$,\$) = 1$.
    Now, the base position-wise comparison kernel is
    \begin{equation}
    k(X, Y) = \prod_{l=0}^\infty k_s(X_{(l)}, Y_{(l)}).
    \end{equation}
\end{definition}
\noindent Note that because $k_s(\$, \$)=1$, the infinite product is always finite. This kernel compares the sequences $X$ and $Y$ at each position $l$, according to $k_s$.
For example,
one natural choice of $k_s$ is to set $k_s(b, b')=1$ if $b=b'$ and $k_s(b, b')=e^{-\lambda}$ if $b\neq b'$, for $\lambda > 0$.
This gives the ``exponential Hamming kernel'',
$$k(X, Y)=\exp\left(-\lambda\sum_{l}\mathbbm{1}(X_{(l)}\neq Y_{(l)})\right)=\exp(-\lambda d_H(X, Y)),$$
where $d_H$ is the Hamming distance. 
Recall that the Hamming kernel of lag $L = 1$ takes the form $k(X, Y) = |X|\vee|Y|-d_H(X, Y)$. Thus, the two kernels both measure sequence similarity using the Hamming distance, but differ in the functional form of their dependence.

The base position-wise comparison kernel, unlike the weighted degree kernel, has discrete masses.
\begin{theorem}\label{thm: hamming delta}
The base position-wise comparison kernel has discrete masses.
\end{theorem}
\begin{proof} \label{ex: hamming letter comparison reparam}
    Our proof strategy will be to reparameterize the alphabet $\B$ (using Proposition~\ref{prop: reparam delta}) such that the RKHS $\Hc_k$ decomposes into a product of orthogonal hyperplanes, each spanned by a subset of sequences.
    We prove that the kernel restricted to each of these hyperplanes takes a simple form, and has discrete masses. 
    We then apply Proposition~\ref{cor: ortho decomp delta} to merge the separate hyperplanes and prove the result.

    We start by reparameterizing the base position-wise comparison kernel, $k$.
    First note that $k$ is a strictly positive definite kernel by the Schur product theorem.
    Define $K$ as the matrix with $K_{b', b}=k_s(b, b')$ for $b, b'\in\B$.
    Call $t\in\mathbb R^{\B}$ the vector with $t_b=k_s(b, \$)$ for $b \in \B$.
    Let $U = K^{-1}t$ and call $\sigma=U^TKU$.
    If $V, W\in\mathbb R^{\B}$ define $k_{s, V}=\sum_{b\in\B}V_bk_{s, b}$, $k_{s, W}$ analogously, $k_s(V, W)=(k_{s, V}|k_{s, W})_{k_s}$ and $k_s(V, \$)=(k_{s, V}|k_{s,\$})_{k_s}$.
    Then if $V\in\mathbb R^{\B}$ $k_s(V, U)=V^TKU=V^Tt=k_s(V, \$)$.
    Note in particular that since $k_s$ is strictly positive definite, we have the strict Cauchy-Schwartz inequality,
    $$\sigma=k_s(U, U)=k_s(U, \$)<\sqrt{k_s(U, U)k_s(\$, \$)}=\sqrt{\sigma},$$
    which can only be the case if $\sigma<1$.
    Let $B_1, \dots, B_{|\B|-1}\in\mathbb R^{\B}$ be chosen so that $\{U, B_1, \dots, B_{|\B|-1}\}$ is an orthogonal basis of the vector space $\mathbb R^{\B}$ when using the dot product $(v|x)=v^TKw$, with $(B_i|B_i)=1$ for all $i$.
    Thus, $k_s(B_i, B_j)=\delta_i(j)$ for all $i, j$ and $k_s(U, B_i)=0$ for all $i$.
    Call $\tilde \B = \{U, B_1, \dots, B_{|\B|-1}\}$ and $\tilde S$ the set of sequences made up of letters in $\tilde \B$.
    By Proposition~\ref{prop: reparam delta}, if we can show that the kernel $k$ has discrete masses on $\tilde S$, we know that it has discrete masses on $S$.

    We now break up $\tilde S$ into separate domains $W_V$.
    Recall the base position-wise comparison kernel takes the form $k(V, V')=\prod_{l=0}^\infty k_s(V_l, V_l')$. When applied to sequences in the reparameterized alphabet, i.e. $V, V'\in\tilde S$, we have for all $b, b'\in \tilde \B\cup\{\$\}$ that $k_s(b, b')=0$ if $b\neq b'$ except $k_s(U, \$)=\sigma$.
    Thus, if $|X|\geq |Y|=L$ for $X, Y\in \tilde S$ then $k(X, Y)\neq 0$ if and only if the first $L$ letters of $X$ and $Y$ are identical and all the letters of $X$ after position $L$ are $U$.
    If $V\in\tilde S$ and does not end in $U$, define $W_V=\{V+m\times U\}_{m=0}^\infty$ to be all the sequences in $\tilde S$ that start with $V$ and have a tail of $U$s.
    Then if $V' \in \tilde S$ is a different sequence that does not end in $U$, $W_{V'}$ is orthogonal to $W_V$ in $\Hc_k$.
    Thus $\Hc_k$ is made up of orthogonal hyperplanes spanned by the sets $\{W_V\}_V$.
    By Proposition \ref{cor: ortho decomp delta}, $k$ has discrete masses if and only if $k$ has discrete masses when restricted to each $W_V$.
    
    We now show that the kernel applied to each $W_V$ has discrete masses.
    Note, since $k_s(U, \$)=k_s(U, U)=\sigma$, we have $k(V+m\times U, V+m'\times U)=k(V, V)k(m\times U, m'\times U)=k(V, V)\sigma^{m\vee m'}$.
    Thus $k$ restricted to $W_V$ is equivalent to the kernel $k^{(n)}(m, m') = \sigma^{m\vee m'}$ applied to $m, m' \in \mathbb N$, times a constant, $k(V, V)$. 
    We therefore just need to show that $k^{(n)}(m, m')$, a one-dimensional kernel defined over the natural numbers, has discrete masses.
    We prove this by induction.
    First, noting $\sigma<1$, let $f=(\sigma-1)^{-1}(k_1^{(n)}-k_0^{(n)})$.
    We see that $\delta_0=f$, so $\delta_0\in\Hc_k$.
    Now assume $\delta_0, \dots, \delta_{x-1}\in\Hc_k$ for some $x\in\mathbb N$.
    Let $f_x = (\sigma-1)^{-1}(k_{x+1}^{(n)}-k_x^{(n)})$.
    We see that $f_x(z)=0$ if $z> x$ and $f_x(x)=1$.
    Thus $\delta_x = f_x - \sum_{z=0}^{x-1}f(z)\delta_z$, so $\delta_x\in\Hc_k$.
    Invoking Proposition~\ref{cor: ortho decomp delta}, the proof is complete.
\end{proof}

We can leverage the base position-wise comparison kernel to construct more complex position-wise comparison kernels that address specific modeling challenges, and also have discrete masses.
In each of these examples, we use for simplicity $k_s(b, b')=\exp(-\lambda\mathbbm{1}(b\neq b'))$, but one can just as well use any strictly positive definite kernel $k_s$, for example one which compares $b$ and $b'$ based on amino acid similarity.
\begin{example}[Heavy tailed Hamming kernels]\label{ex: hamming thick tail}
\textbf{Practical challenge:} The exponential Hamming kernel is ``thin-tailed'' in the sense that it decreases quickly (exponentially) as the Hamming distance increases. 
Consider a data set $\{X_1, X_2,\ldots\}$ consisting of distantly related sequences, and the Gram matrix $K$ with entries $K_{ij} = k(X_i, X_j)$. Thin tails can lead to ``diagonal dominance'', where the diagonal entries $K_{ii}$ are much larger than the off-diagonal entries, $K_{i j}$ for $i \neq j$.
The kernel may therefore behave much like the identity kernel (Example~\ref{ex:identity_kernel}), and exhibit similarly poor generalization when used in a Gaussian process or other regression method.
Fixing diagonal dominance by tuning $\lambda$ is difficult; typically the only other regime available is one in which the matrix $K$ is nearly constant, i.e. $K_{i,j} \approx constant$ for all $i, j$. In this case, the regression will make approximately the same prediction everywhere, and still exhibit poor generalization.

\textbf{Proposed kernel:}
    To address diagonal dominance, we propose a heavy-tailed kernel, whose value decays slowly as sequence distance increases.
	Take $\beta, C>0$.
    Then the ``inverse multiquadric Hamming kernel'' is defined by integrating over the bandwidth parameter $\lambda$,
    \begin{equation} \label{eqn:iqm_h}
    k(X, Y)=(C+d_H(X, Y))^{-\beta}=\Gamma(\beta)^{-1}\int_0^\infty d\lambda\left( \lambda^{\beta}e^{-C\lambda} \right)e^{-\lambda d_H(X, Y)},
    \end{equation}
	where $\Gamma$ is the gamma function. 
 As Hamming distance increases, this kernel decreases according to a power-law instead of exponentially, i.e. its tails are very heavy.
	By Proposition \ref{prop: delta integration}, this kernel has discrete masses.
\end{example}
\begin{example}[k-mer Hamming kernels]
    \textbf{Practical challenge:} The exponential Hamming kernel considers two sequences similar if they have the same letter in the same position. However, the biological function of a sequence can depend not just on whether it has a certain letter in a certain position, but, further, on whether it has a certain string of letters (such as a motif) at a certain position. It may therefore be more appropriate to judge the similarity of two sequences based on whether they have the same \textit{kmers} at the same position, rather than single letters.
    This is the idea behind the Hamming kernel of lag $L>1$. The problem is that this kernel is neither characteristic nor universal (Section~\ref{sec: toy example}).

    \textbf{Proposed kernel} We propose a kernel that uses the same measure of sequence similarity as the Hamming kernel of lag $L$, but takes a different functional form, which ensures it has discrete masses.
	Let $L$ be the context size, and define $\tilde \B=\B^L$ to be the set of sequences of length $L$.
We will treat $\tilde{\B}$ as an alphabet, and consider the set of sequences made from this alphabet, $\tilde S$.
	Now consider the kernel $\tilde k(X, Y)=(C+d_H(X, Y))^{-\beta}$ on $\tilde S$ for some $C, \beta>0$; recall that this kernel has discrete masses.
	Each sequence in $S$ (the original sequence space) can be uniquely embedded in $\tilde S$ by the mapping $i:S\to\tilde S$, defined as
	$$h(X) = (X_{(:L)}, X_{(1:L+1)}, X_{(2:L+2)}, \dots),$$
	that is, $h$ maps each sequence to its sequence of $L$-mers in order.
	Note $h$ is injective, but it is not surjective.
	By Proposition \ref{cor: ortho decomp delta}, 
	 $\tilde k$ restricted to the image of $h$ has discrete masses.
	Thus the ``inverse multiquadric Hamming kernel of lag $L$'',
	$$k(X, Y)=\tilde k(h(X), h(Y))=\left(C+\sum_{l=0}^{|X|\vee|Y|-1}\mathbbm{1}(X_{(l:l+L)}\neq Y_{(l:l+L)})\right)^{-\beta},$$
 which is defined for $X, Y\in S$, has discrete masses.
 Note that the sum in the above expression is precisely the Hamming kernel of lag $L$.
 In other words, the notion of sequence similarity has not changed, but the functional form of the kernel has, and this gives the new kernel discrete masses. 
\end{example}	
\begin{example}[Centre-justified Hamming kernels]
\textbf{Practical challenge} The exponential Hamming kernel compares sequences position-by-position starting from position one. This is sensible when position one is a good reference point, such as the start of a gene. In some cases, however, we are interested in a region of sequence to either side of a reference point, for instance the DNA sequence upstream as well as downstream of the start codon (i.e. both promoter and coding regions). 
For example, rather than comparing sequences as,
	\begin{gather*}
		|\texttt{AACTTCT\$\$\$\$\dots}\\
		|\texttt{GGACTTCTCT\$\dots}
	\end{gather*}
 where $|$ marks the reference point, we want to compare them as,
	\begin{gather*}
		\texttt{\dots\$\$AACT}|\texttt{TCT\$\$\$\dots}\\
		\texttt{\dots\$GGACT}|\texttt{TCTCT\$\dots}
	\end{gather*}
 Note that different sequences may now have variable lengths to either side of the reference point $|$.
 
\textbf{Proposed kernel} We can interpret each data point as consisting not of one sequence but rather a pair of two sequences, $(X^{(1)}, X^{(2)})$, where $X^{(1)}$ is the sequence to the left of the reference point and $X^{(2)}$ to the right. Thus, each data point is in $S^2$ rather than $S$.
Starting with the exponential Hamming kernel (or any other kernel with discrete masses), we can extend it to $S^2$ using tensorization, and Proposition \ref{prop: tensor delta} guarantees it has discrete masses.
\end{example}	
\begin{example}[Hamming kernels with shifts]
\textbf{Practical challenge} It is not always clear which positions exactly should be compared between two sequences. For instance, while protein sequences often have a clear starting position (the start codon), for other genetic elements (such as enhancers) the ``beginning'' of the sequence is less well-defined. 
One approach to this problem is to compare sequences under various offsets relative to one another, as in the Hamming kernel with shifts~\citep{Sonnenburg2007-yr}.
The Hamming kernel with shifts, however, does not have discrete masses. 

\textbf{Proposed kernel} We can define a shifted position-wise comparison kernel as
$$k_L(X, Y)=\sum_{l=0}^L k(X_{(l:)}, Y)+k(X, Y_{(l:)}),$$ 
where $k$ is, for instance, the exponential Hamming kernel. This kernel compares $X$ to $Y$ under various offsets, i.e. starting from position $l = 0$, then position $l = 1$, etc., up to some maximum $L$; it will be large if at least one of these offsets produces a good match.
The kernel is a sum over individual kernels with discrete masses, so by Proposition~\ref{prop: delta integration} it has discrete masses.
\end{example}
Note each of these proposed kernels, in addition to having discrete masses, is computationally tractable, and in particular can be efficiently computed using the techniques developed for weighted degree kernels~\citep{Sonnenburg2007-yr}.

\section{Alignment kernels}\label{sec: string kernels}
In this section, we design kernels with discrete masses that compare sequences based on pairwise alignments.
Whereas position-wise comparison kernels judge two sequences $X$ and $Y$ to be similar only if there are a small number of substitutions that can transform $X$ into $Y$, alignment kernels judge two sequences to be similar if there are a small number of substitutions, insertions and/or deletions that can transform $X$ into $Y$.
Biologically, sequences that are similar in this way often share similar phenotypes, and are closely related evolutionarily. 
However, this notion of similarity is poorly captured by position-wise comparison kernels.
For example, the sequence $X = ATGC$ differs from the sequence $Y = TGC$ only by the insertion of a single $A$ at the start of the sequence, but kernels based on the Hamming distance will judge the two sequences to be completely different, with distance $d_H(X, Y) = 4 = |X|$.

One popular, but problematic, technique for dealing with insertions and deletions is to pre-process the data set using a multiple sequence alignment algorithm, which adds gap symbols to each sequence to make it the same length.
In essence, the idea behind multiple sequence alignment algorithms is to construct a point estimate of which letters in each sequence are evolutionarily related to one another via substitutions, and place each related letter in the same position, i.e. in the same column of the pre-processed data matrix. 
Once sequences have been aligned, it is more reasonable to apply a position-wise comparison kernel; insertions and deletions are compared by including the gap symbol in the alphabet $\B$. 
The problem with multiple sequence alignment pre-processing is that (a) it does not take into account uncertainty in which positions are related to one another, instead relying on a point estimate, and (b) it prevents downstream machine learning methods from generalizing to unseen sequences, since adding new sequences to the data set can change the multiple sequence alignment, for instance, if the new sequence is longer than those previously observed~\citep{Weinstein2021-ce}.

Alignment kernels offer an alternative approach to accounting for insertions and deletions~\citep{Haussler1999ConvolutionKO}.
Rather than transform the entire data set, they consider two sequences $X$ and $Y$ to be similar if they differ by a small number of insertions and deletions, as well as substitutions.
Alignment kernels are of special relevance to problems involving sequences with high length variation, such as in the analysis of human antibodies or of distantly related evolutionary homologs.

\subsection{The alignment kernel}\label{sec: the string kernel}
In this section we define the alignment kernel and show that it has discrete masses if and only if its hyperparameters are set to values in a certain range.

In biological sequence analysis, a pairwise alignment between two sequences is a matching between a subset of positions in each sequence, with the restrictions that (a) each position in each sequence can be matched to at most one position in the other sequence, and (b) matchings must be ordered, such that if site $l_X$ in sequence $X$ is matched to $l_Y$ in sequence $Y$, site $l'_X > l_X$ in $X$ cannot match to a site $l'_Y < l_Y$ in $Y$~\citep[chap.~2]{Durbin1998-kn}. For example, one alignment between the sequences $X = \verb|ATGC|$ and $Y=\verb|ACC|$ is,
\begin{gather*}
    \texttt{ATGC}\\
    \texttt{|| |}\\
    \texttt{AC-C}
\end{gather*}
where vertical lines $|$ denote a matching. Here, \verb|-| is a gap symbol, denoting the fact that the nucleotide \verb|G| in $X$ is unmatched; we can interpret the \verb|G| as an insertion in $X$ relative to $Y$, or, conversely, interpret the gap \verb|-| as a deletion in $Y$ relative to $X$.
The alignment kernel $k(X, Y)$ considers all possible pairwise alignments between $X$ and $Y$, scores each alignment according to a position-wise comparison kernel applied to the aligned sequences, and sums up those scores to produce its value.

Mathematically, to define the alignment kernel, we start with two simpler kernels and then convolve them, following the construction of~\citet{Haussler1999ConvolutionKO}.
The first kernel, $k_s$, evaluates matches between aligned letters. 
The second kernel, $k_I$, evaluates insertions and deletions; it applies an ``affine'' gap penalty, which separately penalizes the presence of a gap and the gap's total length.
For any two kernels $k_1$ and $k_2$ on $S$, the convolution kernel $k_1\star k_2$ is,
$$(k_1\star k_2)(X, Y)=\sum_{X^{(1)}+X^{(2)}=X}\, \sum_{ Y^{(1)}+Y^{(2)}=Y} k_1(X^{(1)}, Y^{(1)})k_2(X^{(2)}, Y^{(2)})$$
where the sum is over all sequences $X^{(1)}, X^{(2)}, Y^{(1)}, Y^{(2)} \in S$ such that $X^{(1)}+X^{(2)}=X$ and $Y^{(1)}+Y^{(2)}=Y$. Recall that $S$ includes the sequence of length zero, so the sum includes such terms as $X^{(1)} = X, X^{(2)} = \emptyset$.
\begin{definition}[Alignment kernel] \label{def:alignment_kernel}
Let $k'_s$ be a strictly positive definite kernel on $\B$. Define $k_s$ by extending $k'_s$ to $S$ with $k_s(X, Y)=0$  if $|X|\neq 1$ or $|Y|\neq 1$.

Let $\Delta\mu > 0$ be the penalty for starting an insertion and let $\mu > 0$ be the penalty for the insertion's length.
The insertion penalty function $g$ is then defined as $g(X)=1$ if $|X|=0$ and $g(X)=\exp\left(-\Delta\mu-|X|\mu\right)$
otherwise.
Define the non-negative-definite kernel $k_I(X,Y)=g(X)g(Y)$ for $X, Y\in S$.

The alignment kernel $k$ is,
$$k = \sum_{l=0}^\infty k_I\star(k_s\star k_I)^{\star l},$$
where the exponent $\star l$ denotes the convolution of $l$ kernels $k_s\star k_I$.
\end{definition}
\noindent To see that this kernel indeed considers all possible pairwise alignments of $X,Y\in S$, note that we can rewrite $k(X, Y)$ as,
\begin{equation} \label{eqn:pairwise_alignment_terms} \sum_{l=0}^\infty\, \sum_{X^{(1)}+ \dots+ X^{(2l+1)}=X}\, \sum_{Y^{(1)}+ \dots+ Y^{(2l+1)}=Y} \!\!\!\!\!\!\!\!\!\!\!k_I(X^{(1)}, Y^{(1)})\prod_{i=1}^l k_s(X^{(2i)}, Y^{(2i)})k_I(X^{(2i+1)}, Y^{(2i+1)}).
\end{equation}
Each term in the sum represents an alignment. Each alignment has $l$ matches, in which the even subsequences $X^{(2)}, X^{(4)},\ldots, X^{(2l)}$ of $X$ are matched to the even subsequences $Y^{(2)}, Y^{(4)},\ldots, Y^{(2l)}$ of $Y$. For the term to be non-zero, each of these subsequences must consist of a single letter, since otherwise $k_s(X^{(2i)}, Y^{(2i)}) = 0$. 
Between each of the match positions in $X$ and $Y$ there may be insertions, corresponding to the odd subsequences $X^{(1)}, X^{(3)},\ldots, X^{(2l+1)}$ and $Y^{(1)}, Y^{(3)},\ldots, Y^{(2l+1)}$. These subsequences can be of any length, including zero. 
The example alignment between $X = \verb|ATGC|$ and $Y=\verb|ACC|$ we saw earlier in this section corresponds to $X^{(1)} = \emptyset$, $X^{(2)} = \verb|A|$, $X^{(3)} = \emptyset$, $X^{(4)} = \verb|T|$, $X^{(5)} = \verb|G|$, $X^{(6)} = \verb|C|$, $X^{(7)}=\emptyset$, $Y^{(1)} = \emptyset$, $Y^{(2)} = \verb|A|$, $Y^{(3)}=\emptyset$, $Y^{(4)}=\verb|C|$, $Y^{(5)} = \emptyset$, $Y^{(6)} = \verb|C|$, $Y^{(7)}=\emptyset$.

The kernel sums over all possible alignments: the outermost sum in Equation~\ref{eqn:pairwise_alignment_terms} is over the number of matches, $l \in \{0, 1, 2, \ldots\}$, and the inner sums are over all possible choices of match positions in $X$ and in $Y$.
Each alignment is then scored according to $k_s$ and $k_I$.

How flexible is the alignment kernel?
\citet{Haussler1999ConvolutionKO} showed it is strictly positive definite.
\citet{Jorgensen2015-na}, motivated by problems outside biological sequence analysis, showed that the ``binomial RKHS'' does not have discrete masses; this corresponds to the alignment kernel with $\Delta\mu=\mu=0$, $|\B|=1$, and $k_s(b, b)=1$.
We now show that the alignment kernel can have discrete masses if and only if $\mu, \Delta\mu$ and $k_s$ satisfy certain constraints.

\begin{theorem}[Alignment kernels can have discrete masses]\label{thm: delta funcs string}
    Define $K$ as the matrix with $K_{b,b'}=k_s(b, b')$ for $b, b'\in\B$, and $\sigma=\mathbf{1}_{\B}^T K^{-1} \mathbf{1}_{\B}$ where $\mathbf{1}_{\B}\in\mathbb R^{\B}$ is the vector with $1$ in each entry.
	If $\Delta\mu>0$, then $k$ has discrete masses if and only if $2\mu \geq \log\sigma$.
	If $\Delta\mu=0$, then $k$ has discrete masses if and only if $2\mu > \log\sigma$.
\end{theorem}
\noindent The proof is in Appendix~\ref{sec: proof of ali discrete masses}. It uses a similar strategy to that employed in proving that the position-wise comparison kernel has discrete masses (Theorem~\ref{thm: hamming delta}): reparameterize the alphabet, then break up sequence space into orthogonal hyperplanes.
Over the orthogonal hyperplanes, the kernel takes a simpler form, which we prove has discrete masses using the alignment kernel's feature representation; those features will be derived in Section~\ref{sec: string kernel basis} below.

Practically, Theorem~\ref{thm: delta funcs string} shows that if we set the hyperparameters of the alignment kernel naively, it may become unreliable, while if we set them appropriately, it has guaranteed flexibility.
In particular, if we want flexibility, our choice of gap penalties ($\mu$ and $\Delta \mu$) is constrained by our choice of substitution penalties ($k_s$) through $\sigma$. 
If there is no penalty for starting a gap ($\Delta \mu=0$) we require the penalty for extending a gap ($\mu$) to be larger than $\frac{1}{2}\log(\sigma)$ in order for the kernel to have discrete masses.
If there is a penalty for starting a gap ($\Delta \mu>0$), the penalty for extending a gap can be slightly smaller, in that $\mu = \frac{1}{2}\log(\sigma)$ also gives discrete masses.
In Appendix \ref{sec: ali with base spec} we make sense of these conditions as arising from the requirement that the position-wise comparison kernel applied to individual alignments has discrete masses.
In Appendix~\ref{sec: ali kernl univ} we investigate relaxations of these conditions, such that the alignment kernel no longer has discrete masses but is still universal; we find that such relaxations are only valid given additional, problematic modifications of the kernel.

\subsection{Heavy-tailed alignment kernels}\label{sec: connection to k-mer kernels}

Practically, when applied to data sets with highly diverse sequences, regression methods that use the alignment kernel often exhibit poor generalization due to diagonal dominance~\citep{Haussler1999ConvolutionKO,Saigo2004-go,Weston2003-sg}.
Previous efforts to address this problem have encountered substantial difficulties; practitioners have gone so far as to introduce alternative ``kernels'' that are not positive semi-definite~\citep{Saigo2004-go}.
In this section, we diagnose possible sources of diagonal dominance in the alignment kernel, and propose heavy-tailed modifications of the alignment kernel that still possess discrete masses.

The key observation is that the alignment kernel, in effect, applies a position-wise comparison kernel to each pairwise alignment between sequences; diagonal dominance stems from the fact that the tail of this position-wise comparison kernel is thin.
More precisely, consider an alignment kernel (Definition~\ref{def:alignment_kernel}) with a letter kernel $k'_s(b, b')=\exp(-\lambda\mathbbm{1}(b\neq b'))$; recall this choice gave us the exponential Hamming kernel in Section~\ref{sec: hamming kernels}.
Each term in the sum defining the alignment kernel (Equation~\ref{eqn:pairwise_alignment_terms}) corresponds to an individual alignment, and contains a factor corresponding to the exponential Hamming kernel applied to the matched positions, namely $\prod_{i=1}^l k_s(X^{(2i)}, Y^{(2i)})$.
As the Hamming distance between the matched positions increases, the term decays very quickly (exponentially), giving rise to diagonal dominance.
To fix this problem, we follow the same strategy as in Example~\ref{ex: hamming thick tail}, and integrate over the bandwidth parameter $\lambda$ to derive a thick-tailed alternative.
\begin{example}[Heavy tailed alignment kernel on matches]\label{ex: ali thick tail}
The ``thick tailed alignment kernel on matches'' is the kernel $\tilde{k}(X, Y) = \Gamma(\beta)^{-1}\int_0^\infty d\lambda\left( \lambda^{\beta}e^{-C\lambda} \right) k_\lambda(X, Y) $, where $k_\lambda$ is the alignment kernel using $k'_s$ with parameter $\lambda$, and $C,\beta > 0$. From Equation~\ref{eqn:iqm_h}, we find that $\tilde{k}(X, Y)$ is,
\begin{equation*}
\begin{split}
\sum_{l=0}^\infty\, \sum_{X^{(1)}+ \dots+ X^{(2l+1)}=X}\, \sum_{Y^{(1)}+ \dots+ Y^{(2l+1)}=Y} (C& + d_H(X^{(2)} + \dots + X^{(2l)}, Y^{(2)} + \dots + Y^{(2l)}))^{-\beta} \\ & \times k_I(X^{(1)}, Y^{(1)})\prod_{i=1}^l k_I(X^{(2i+1)}, Y^{(2i+1)})
\end{split}
\end{equation*}
In this kernel, the dependence on the Hamming distance between matched positions follows a power law, rather than an exponential decay.
If we define $K_{\lambda, b, b'}=k_\lambda(b, b')$ then $\sigma_\lambda=\mathbf{1}_\B^TK_\lambda^{-1}\mathbf{1}_\B=\left(\frac 1 {|\B|} + \frac{|\B|-1}{|\B|}e^{-\lambda}\right)^{-1}$, which approaches $1$ as $\lambda\to 0$.
Thus, as long as $\mu>0$, $k_\lambda$ will have discrete masses for small $\lambda$, by Theorem~\ref{thm: delta funcs string}.
So, by Proposition~\ref{prop: delta integration}, $\tilde k$ has discrete masses.
Note that this modified kernel can, like the standard alignment kernel, be computed efficiently using a dynamic programming algorithm (Appendix~\ref{sec: thick ali kern}).
\end{example}

Besides the substitution penalty $\lambda$, we can also consider the role of the insertion penalties $\mu$ and $\Delta \mu$.
We find a similar issue: as the length of insertions increases, each term of the alignment kernel decays exponentially, since $k_I(X, Y)=\exp\big(-\mathbbm{1}(|X| > 0)(\Delta \mu + \mu |X|)-\mathbbm{1}(|Y| > 0)(\Delta \mu + \mu |Y|)\big)$. 
Now, the parameter $\mu$ plays the role of the bandwidth. 
To produce thick-tails, we can perform the analogous transform, and take the integral $\Gamma(\beta)^{-1}\int_0^\infty d\mu\left( \mu^{\beta}e^{-C\mu} \right) k_\mu(X, Y) $, where $k_\mu$ is the alignment kernel with parameter $\mu$. This gives us a heavy-tailed alignment kernel on gaps, which has discrete masses by Proposition~\ref{prop: delta integration}. 
We could even apply the same transform to the thick tailed alignment kernel on matches, yielding a thick tailed alignment kernel on both gaps and matches.

\subsection{Local alignments}

Genomes are very long, and so in practice machine learning is typically done only on specific genetic elements, such as genes, promoters, enhancers, etc.
It is sometimes ambiguous where exactly these genetic elements start or end.
This motivates the idea of a local alignment, which ignores insertions and deletions at the start and end of a sequence. Using a local alignment as a measure of similarity means two sequences that contain the same genetic element but different flanking regions will still be considered similar~\citep[Chap. 2]{Durbin1998-kn}.
In this section, we study a local alignment kernel, and establish conditions under which it has discrete masses~\citep{Vert2004-qd,Saigo2004-go}.

The local alignment kernel is a modification of the alignment kernel, which does not penalize the creation of gaps at the start or end pairwise alignments. 
\begin{definition}[Local alignment kernel]
Let $k_I$ and $k_s$ be as in Definition~\ref{def:alignment_kernel}. Also define the modified insertion gap penalty kernel $\tilde k_I(X, Y)=\exp(-\mu(|X|+|Y|))$, which lacks the gap start penalty $\Delta\mu$. The local alignment kernel is,
$$k_{\mathrm{la}}=\tilde k_I +\sum_{l=1}^\infty \tilde k_I\star(k_s\star k_I)^{\star (l-1)}\star k_s\star \tilde k_I.$$
\end{definition}
We can extend the logic of the proof of Theorem \ref{thm: delta funcs string} to establish conditions under which local alignment kernels have discrete masses; these conditions turn out to be identical to those for the regular alignment kernel.
\begin{theorem}\label{thm: gen string}
    Define $K$ as the matrix with $K_{b,b'}=k_s(b, b')$ for $b, b'\in\B$, and $\sigma=\mathbf{1}_{\B}^T K^{-1} \mathbf{1}_{\B}$ where $\mathbf{1}_{\B}\in\mathbb R^{\B}$ is the vector with $1$ in each entry.
	If $\Delta\mu=0$, $k_{\mathrm{la}}$ has discrete masses if and only if $2\mu>\log\sigma$.
	If $\infty>\Delta\mu>0$, $k_{\mathrm{la}}$ has discrete masses if and only if $2\mu\geq \log\sigma$.
	If $\Delta\mu=\infty$, $k_{\mathrm{la}}$ has discrete masses regardless of the values of $\mu, \sigma$.
\end{theorem}
\noindent A proof is given in Appendix~\ref{sec: proof of local ali discrete masses}.
Note that for the standard, non-local alignment kernel, if we set the gap start penalty to infinity ($\Delta \mu = \infty$), we recover a position-wise comparison kernel.
For local alignments, however, the situation is more complex, as there are different options for the number of gaps to put at the start and end of each sequence.
We find that when $\Delta\mu=\infty$, the local alignment kernel has discrete masses regardless of the value of its other hyperparameters.

\section{Kmer spectrum kernels}\label{sec: spectra}

In this section we study kmer spectrum kernels.
Kmer spectrum kernels, like alignment kernels, are motivated by the biological observation that sequences which differ by insertions, deletions, or other complex mutations often share similar phenotypes and are related evolutionarily.
Remarkably, we will find that the relationship between kmer spectrum kernels and alignment kernels runs deeper than their motivation: the two are in certain cases equivalent, up to an appropriate tilting.

Kmer spectrum kernels featurize sequences according to the presence and absence of subsequences (kmers).
Since the position of the kmer within the sequence does not affect the value of the corresponding feature, kmer spectrum kernels can be relatively insensitive to insertions or deletions, as compared with position-wise comparison kernels.
In Example~\ref{ex: spectrum kernels}, we saw a commonly used kmer spectrum kernel which had only a finite number of features, and so was neither universal nor characteristic. 
Here we will study infinite-dimensional kmer spectrum kernels, and show that they have discrete masses.

\subsection{Gapped kmer spectrum kernels}\label{sec: string kernel basis}

We start by connecting alignment kernels to kmer spectrum kernels. We show that, for a particular tilting, the features of the alignment kernel are a weighted sum of kmer counts, allowing for some kmer mismatches.
More precisely, we show that the tilted alignment kernel is equivalent to an infinite-dimensional gapped kmer spectrum kernel. Since tilting preserves discrete masses (Proposition~\ref{cor: tilting delta}), this gapped kmer spectrum kernel has discrete masses.

Intuitively, a ``gapped kmer'' is a kmer that includes gaps; it matches another sequence if each of the non-gap characters match, regardless of the letters at the gap positions. For example, the gapped kmer sequence \verb|AT-G| matches \verb|ATCG|, \verb|ATGG|, \verb|ATAG| and \verb|ATTG|. 
Formally, a gapped kmer of length $L$, with $M$ non-gap letters, is defined by an ordered set of indices $J=(j_0, \ldots, j_{M+1})$, with $-1=j_{-1}<j_0<j_2< \dots<j_M<j_{M}=L$, and a sequence of $M$ non-gap letters, $Z \in \B^M$.
Each $j_m$ for $m \in \{0, \ldots, M-1\}$ indexes the location of the non-gap letter $Z_{(m)}$.
There is a gap after the $m$th letter if $j_m +1\neq j_{m+1}$.
Let $G(L, M, g)$ be the set of all gapped kmer indices $J$, with length $L$, $M$ non-gap letters, and $g$ gaps.
Let $G(L, g)=\cup_{M} G(L, M, g)$ be the union over all possible $M$.
For any sequence $X\in S$, and $J=(j_0, \dots, j_{M+1})\in G(|X|, g)$, define $X_{(J)}=X_{(j_0)}+\dots+X_{(j_{M-1})}$.

The following theorem equates alignment kernels to gapped kmer spectrum kernels. In particular, it shows that we can represent alignment kernels in terms of features $u_V(X)$ that depend on the number of times a sequence $V \in S$ appears in $X$, allowing for gaps. In other words, each feature $u_V(X)$ depends on a sum over the counts of all gapped kmers with non-gap letters $V$.
\begin{theorem}[The features of alignment kernels are gapped kmer counts]\label{thm: string basis}
Consider an alignment kernel $k$ with  $k_s(b, b')=\sigma^{-1}|\B|\delta_{b'}(b)$ for $b, b'\in\B$.
Tilting the kernel by $A(X) = e^{\mu|X|}$ gives $k^A(X, Y) = A(X) A(Y) k(X, Y)$.
For all sequences $X \in S$, define the features $\{u_V(X)\}_{V \in S}$, given by
\begin{equation} \label{eqn:alignment_feature}
u_V(X)=e^{\frac{1}{2} \zeta |V|}\sum_{g=0}^\infty e^{-\Delta\mu g}\sum_{J\in G(|X|, g)}\mathbbm{1}(X_{(J)}=V),
\end{equation}
where $\zeta = 2\mu -\log \sigma+\log|\B|$.
Now, $\{u_V(X)\}_{V \in S}$ is an orthonormal basis for $\Hc_{k^A}$. We call the kernel $k^A(X, Y)$ the ``gapped kmer spectrum kernel''. 
\end{theorem}
\noindent The proof is in Appendix~\ref{sec: features of the ali kern}. This theorem shows that the features of the tilted alignment kernel depend on a weighted sum over gapped kmer counts. The weights $e^{-\Delta \mu g}$ decrease as the number of gaps increases, thus penalizing gapped kmers with large numbers of gaps.
Since tilting preserves discrete masses, we can conclude that the gapped kmer spectrum kernel has discrete masses if and only if the conditions of Theorem~\ref{thm: delta funcs string} are satisfied.

The tilting in Theorem~\ref{thm: string basis} down-weights longer sequences, but does not change how the alignment kernel judges sequence similarity. 
Indeed, the theorem is straightforward to extend to other tiltings, such as the popular normalizing tilting $A(X)=\sqrt{k(X, X)}^{-1}$. In this case we find $k^A(X, Y)=\frac{u(X)^Tu(Y)}{\Vert u(X)\Vert\Vert u(Y)\Vert}$, where $u(X)$ is the infinite dimensional vector indexed by $S$ with $u_V(X)$ at position $V\in S$.
The theorem can also be extended to non-diagonal $k_s$, in which case the features are the gapped kmer counts with a reparameterized alphabet (Appendix~\ref{sec: features of the ali kern}).

Theorem~\ref{thm: string basis} provides a practical example of a kmer spectrum kernel with discrete masses. It is also useful as a theoretical tool.
Theorem~\ref{thm: delta funcs string B=1} (Appendix~\ref{sec: proof of ali discrete masses |B|=1}) uses Theorem~\ref{thm: string basis} to establish that a simple version of the alignment kernel has discrete masses, which in turn allows us to prove that the full alignment kernel has discrete masses (Theorem~\ref{thm: delta funcs string}). 
The proof technique used in Theorem~\ref{thm: delta funcs string B=1} relies on a careful analysis of the combinatorics of biological sequence alignments, using the technique of generating functions and the theory of the Riordan group~\citep{Shapiro1991-ak}.

\subsection{Heavy-tailed gapped kmer spectrum kernels}

We now re-examine the diagonal dominance problem in alignment kernels, in light of Theorem~\ref{thm: string basis}.
Equation~\ref{eqn:alignment_feature} says that the feature weights, namely $\exp(\frac{1}{2} \zeta |V|)$, increase exponentially with kmer length $|V|$. 
This exponential scaling means the the kernel $k(X, Y)$ takes a very large value when applied to very similar sequences $X$ and $Y$, as compared to its value when $X$ and $Y$ are very different and only have short kmers in common. This can result in diagonal dominance.
Trying to avoid the problem by making $\zeta$ small leads to loss of flexibility, as the kernel no longer has discrete masses when $\zeta < \log |\B|$.
In other words, $\zeta$ acts as a tuning parameter that trades off kernel flexibility and diagonal dominance.
To address the problem, we propose a gapped kmer kernel with discrete masses in which the weighting on features scales as a power law with respect to kmer length, rather than exponentially.
\begin{example}[Heavy tailed gapped kmer spectrum kernel]\label{ex: ali thick tail feature version}
Call the unscaled features $\tilde u_V(X)=e^{-|V|\zeta/2}u_V(X)$ and notice they do not depend on $\zeta$.
Consider a gapped kmer spectrum kernel tilted by $A(X) = e^{-\frac{1}{2}\zeta|X|}$,
$$k_{\zeta}(X, Y) = e^{-\frac{1}{2}\zeta|X|} e^{-\frac{1}{2}\zeta|Y|}\sum_{V\in S} u_V(X)u_V(Y)=\sum_{V\in S} e^{-\frac{1}{2}(|X| + |Y|)+|V|}\tilde u_V(X)\tilde u_V(Y).$$
Then for $C, \beta>0$, the ``heavy tailed gapped kmer spectrum kernel'' is defined as,
    \begin{equation*}
    \begin{aligned}
    \tilde k(X, Y)=&\Gamma(\beta)^{-1}\int_0^{\infty}k_{\zeta}(X, Y)  \left(\zeta^{\beta}e^{-C\zeta}\right)d\zeta\\
    =&\sum_{V\in S} \left(C+\frac{1}{2}(|X| + |Y|)-|V|\right)^{-\beta}\tilde u_V(X)\tilde u_V(Y).
    \end{aligned}
    \end{equation*}
    In this kernel, shorter features $V$ are down-weighted following a power law, rather than exponential decay.
    Proposition~\ref{prop: delta integration}, Theorem~\ref{thm: delta funcs string} and Theorem~\ref{thm: string basis} together ensure the kernel has discrete masses. The kernel can be computed efficiently using the dynamic programming algorithm given in Appendix~\ref{sec: thick ali kern}.
\end{example}
\noindent Note that since $\zeta$ depends on $\mu$, this kernel is closely related to the heavy tailed alignment kernel on gaps discussed in Section~\ref{sec: connection to k-mer kernels}, which integrates over $\mu$ with the same base measure. 
The exponential decay in pairwise alignment scores with increasing gap length, and the exponential decay in feature weight for shorter kmers, are two ways of looking at the same diagonal dominance problem in alignment kernels.

\subsection{Ungapped kmer spectrum kernels}

We now consider ungapped kmer spectrum kernels, whose features correspond to the number of times a kmer appears exactly in a sequence.
In particular, we show that a local alignment kernel with an infinite gap start penalty, $\Delta \mu = \infty$, is an ungapped kmer spectrum kernel, and that this kmer spectrum kernel therefore has discrete masses.
\begin{proposition}[Infinite kmer spectrum kernel] \label{thm:infinte_kmer_spectrum}
Consider a tilted local alignment kernel $k=k_{\mathrm{la}}^A$, with $\Delta\mu=\infty$, $\zeta=0$, and $A(X)=\exp(\mu|X|)$. This kernel has discrete masses and can be written as
$$k(X, Y) = \sum_{V\in S}u_{\mathrm{la},V}(X)u_{\mathrm{la},V}(Y),$$
where the feature $u_{\mathrm{la},V}(X) = \sum_{l=0}^{|X|-|V|} \mathbbm{1}(X_{(l:(l+|V|))} = V)$ is the number of times the kmer $V$ occurs in $X$.
\end{proposition}
\begin{proof}
Define $G_{\mathrm{la}}(L, M, g)$ as those elements in $G(L) = \cup_g G(L, g)$ that are of length $M$ with $g$ gaps, \textit{not counting gaps at the beginning or end}.
Using the same logic as Theorem \ref{thm: string basis}, the features of the  kernel, after tilting, are, for $V, X\in S$
$$u_{\mathrm{la}, V}(X)=\sum_{g=0}^\infty e^{-\Delta\mu g}\sum_{J\in G_{\mathrm{la}}(|X|, g)}\mathbbm{1}(X_J=V).$$
Now if we set $\Delta\mu=\infty$, these features are simply the number of occurrences, without gaps, of $V$ in $X$. By Theorem~\ref{thm: gen string} and Proposition~\ref{cor: tilting delta}, the kernel has discrete masses.
\end{proof}
The standard kmer spectrum kernel, by comparison, is $k(X, Y) = \sum_{|V| \le L}u_{\mathrm{la},V}(X)u_{\mathrm{la},V}(Y)$ (Example~\ref{ex: spectrum kernels}). That is, it uses a finite rather than an infinite sum over kmers, and so does not possess any flexibility guarantees. The infinite kmer spectrum kernel has strong flexibility guarantees, without sacrificing much computational efficiency: we can compute it using the same dynamic programming methods used for local alignment kernels, since Proposition~\ref{thm:infinte_kmer_spectrum} shows it \textit{is} a local alignment kernel~\citep{Saigo2004-go}.

\section{Embedding kernels}\label{sec: embedding}

In this section we study embedding kernels, which use sequence representations in Euclidean space to judge sequence similarity.
A key motivation for such kernels is to avoid the need for hand-crafted similarity measures, and to instead learn a similarity measure from data.
The data used to learn a representation need not be the same data the kernel is applied to; for example, a common transfer learning technique is to use a large, unlabeled sequence data set to learn a sequence representation, and then, on a small labeled data set, learn a regression from representations to labels~\citep{Yang2018-ao,Alley2019-yy,Biswas2021-pp,Rao2019-qg,Detlefsen2022-jl}.
Popular modern unsupervised representation learning methods include variational autoencoders, recurrent neural networks, and transformers.
Such techniques have often been found to produce low-dimensional representations that reflect biological properties well, in the sense that sequences with similar representations have similar phenotypes.

An embedding kernel consists of a continuous kernel applied to sequence representations.
\begin{definition}[Embedding kernel]
Let $F:S\to \mathbb R^D$ be an embedding function, which maps sequences to representations. The embedding kernel is defined as $k(X, Y) = k_E(F(X), F(Y))$, where $k_E$ is a kernel on $\mathbb{R}^D \times \mathbb{R}^D$.
\end{definition}
\noindent For example, if we use a variational autoencoder to learn a representation, and a Gaussian process to predict labels, $F$ would be given by the encoder and $k_E$ would be the kernel in the Gaussian process.
Deep kernel learning also fits the form of an embedding kernel, with $F$ a neural network~\citep{Wilson2016-rt}.
If we regress directly from sequence to labels using a neural network, we can think of the map from sequence space to the final hidden layer of the network as the embedding function $F$, while the map from the hidden layer to the output is approximately a kernel regression, by the argument that single layer neural networks converge to Gaussian processes in the infinite width limit~\citep{Neal1996-xn,Williams1996-it}. 

\subsection{Universality and characteristicness}

We first describe conditions under which embedding kernels are universal and characteristic.
Intuitively, if we choose a kernel $k_E$ in Euclidean space that is universal and characteristic, then the embedding kernel over sequence space will be universal and characteristic as well, so long as the embedding function $F$ maps each sequence in $S$ to a unique point in $\mathbb{R}^D$.
\begin{proposition}[Universal and characteristic embedding kernels] \label{prop:embedding_injective}
Assume $k_E(z, z') = \Psi(z - z')$ is a translation invariant kernel and that $\Psi$ is a positive continuous function on $\mathbb{R}^D$ that has a strictly positive Fourier transform.\footnote{
    $\Psi$ has strictly positive Fourier transform if $\hat\Psi(\xi)>0$ for all $\xi$ where $\hat\Psi$ is the Fourier transform of $\Psi$. For example, $\Psi(x) = \exp(-x^2)$ has a strictly positive Fourier transform.}
    Then the embedding kernel $k$ is characteristic and $C_0$-universal if and only if $F$ is injective.
\end{proposition}
\noindent The proof is in Appendix~\ref{sec: proofs of flex embedding}. Note that the conditions on $k_E$ ensure that it is universal and characteristic over $\mathbb{R}^D$~\citep{Sriperumbudur2010-ii,Sriperumbudur2011-ay}.

The condition that $F$ is injective conflicts with some representation learning approaches, particularly those aimed at learning sparse, interpretable features.
For the sake of interpretability, it is often desirable to learn a representation that ignores certain features of the data points entirely, such as those features which vary little in the training data.
However, when this representation is applied to make predictions about held out data, those sequences that do vary in the ignored features can be mapped to the same point in the embedding space. Since the embedding $F$ is therefore not injective, downstream supervised predictors will be limited in their flexibility. An example is in Appendix~\ref{sec: proofs of sparse not inj}.

By contrast, a completely ``uninterpretable'' representation, which makes no attempt to organize embedding space by a biological notion of sequence similarity, can easily yield universal and characteristic embedding kernels.
In particular, if we use an embedding function that gives sequences random representations, distributed according to a continuous distribution on $\mathbb{R}^D$, then the chance of any two sequences having the same embedding is zero.
\begin{proposition}[Random embeddings give universal and characteristic kernels] \label{prop:random_universal}
Assume $\{F(X)\}_{X \in S}$ are i.i.d. random variables distributed according to a measure which is absolutely continuous with the Lebesgue measure. Then $F$ is almost surely injective, and so an embedding kernel based on $F$ is almost surely universal and characteristic.
\end{proposition}
\begin{proof}
    Number the elements of $S$, such that $\{X_1, X_2, \ldots \} = S$. For any $n \in \mathbb{N}$, we have that $F(\{X_1, \ldots, X_n\})$ is a set of measure zero, so $p(F(X_{n+1})=F(X_i)\text{ for some }i\leq n)=0$.
    Thus, $p(F(X_i)=F(X_j)\text{ for some }i, j)=0$. 
\end{proof}

One implication of this proposition is that 
the ``meaningless" randomness introduced through stochastic training of large neural networks may help them produce representations that are useful for downstream tasks. 

\subsection{Discrete masses}

We have so far described conditions under which embedding kernels are characteristic and universal; we found that random embeddings suffice.
However, we saw in Example 3 an embedding kernel that is universal and characteristic but cannot reliably be used for optimization, as it does not metrize the space of distributions.
In this section, we address the question of when embedding kernels have discrete masses, and thus metrize $\mathcal{P}(S)$.
We find that many standard representation learning techniques are likely to produce embedding kernels that lack discrete masses even when they are universal and characteristic, and propose a fix.

For an embedding kernel to have discrete masses, the embedding function $F$ must not only map different sequences to different points in $\mathbb{R}^D$---those points must also be sufficiently spread out.
\begin{proposition}[Embedding kernels with discrete masses] \label{prop:embed_discrete_mass}
Consider an embedding kernel with an injective $F$ that meets the conditions of Proposition~\ref{prop:embedding_injective}.
$k_E$ metrizes $\Pc(S)$ if and only if $k_E$ has discrete masses, which occurs if and only if $F(S)$ has no accumulation points, that is, there is no $X \in S$ such that $F(X)$ is in the closure of $F(S \setminus \{X\})$.
\end{proposition}
\noindent The proof is given in Appendix~\ref{sec: proofs of flex embedding}. In Example~\ref{ex:embedding_metrize}, we saw an embedding kernel that was universal and characteristic, but which had an accumulation point at the sequence $A$, and thus did not metrize $\Pc(S)$.

Unlike for universality and characteristicness, to construct an embedding kernel with discrete masses, it is not sufficient to use sequence representations that are drawn i.i.d. from a distribution on $\mathbb{R}^D$.
\begin{proposition}[I.i.d. embeddings do not have discrete masses] \label{prop:iid_embed_no_masses}
    Assume $\{F(X)\}_{X \in S}$ are i.i.d. random variables. Then the embedding function has accumulation points almost surely, and so does not metrize $\Pc(S)$.
\end{proposition}
\begin{proof}
Consider any $X_0\in S$, and note that almost surely, $p(\Vert F(X)- F(X_0)\Vert<\epsilon)>0$ for any $X, \epsilon$. Thus,
    $p(\Vert F(X)- F(X_0)\Vert<\epsilon\text{ for some }X)=1$ for all $\epsilon$.
    So, with probability $1$ there are sequences $X_{n_1}, X_{n_2}, \dots$ which are not $X_0$ but where $F(X_{n_k})\to F(X_0)$ as $k\to \infty$.
\end{proof}
Intuitively, to construct an embedding without accumulation points, one must somehow fit sequence space $S$ into $\mathbb{R}^D$ without letting the representations of each sequence get too close together. 
One approach is to make embeddings of longer and longer sequences more and more spread out, rather than sampling all the embeddings from the same distribution.
\begin{proposition}[Scaled random embeddings have discrete masses] \label{prop:scaled_embedding}
    Consider an initial embedding $\tilde F$ where each $\tilde F(X)$ for $X \in S$ is drawn from the uniform distribution on the sphere, $\{x \in \mathbb{R}^D\, |\, \|x\| \le 1\}$.
    Then, a kernel using the scaled embedding $F(X)=|\B|^{(1+\epsilon)|X|/D}\tilde F(X)$, for $\epsilon>0$, metrizes $\Pc(S)$ almost surely.
\end{proposition}
\noindent The proof can be found in Appendix~\ref{sec: proofs of random embedding}.
Intuitively, the idea is that this rescaling gives each sequence representation ``enough space''. If $V(B(1))$ is the volume of the unit ball in $\mathbb R^D$, we can think of $F$ as embedding the set $\B^L$ into a ball of radius $|\B|^{(1+\epsilon)L/D}$.
    Thus, the average amount of volume per sequence is $V(B(1))(|\B|^{(1+\epsilon)L/D})^D/|\B|^{L}=V(B(1))|\B|^{L\epsilon}$, which is growing with $L$.
    This is just enough space to ensure there is no accumulation.

Practically, Proposition~\ref{prop:iid_embed_no_masses} suggests that many representation learning approaches are unlikely to yield kernels that metrize $\Pc(S)$. Typically, as we embed more and more sequences, their representations stay clustered in a localized region of embedding space near the origin, rather than diverging away.
Indeed, in many applications of representation learning, this behavior is desirable, as it makes the representations easier to visualize and interpret. 
One way to understand this behavior is as a consequence of the priors and regularizers typically used in representation learning.
For instance, in latent variable models, we expect the set of embeddings $\{F(X)\}_{X \in S}$, taken as a whole, to look roughly like a set of samples from the prior on the latent variable; since the prior is typically chosen to be a continuous distribution on $\mathbb{R}^D$, such as a normal distribution, Proposition~\ref{prop:iid_embed_no_masses} suggests the embedding kernel will not have discrete masses.
The fact that we only observe a finite number of data points in the training data can also hurt: even if the representations of the training data are well spread out, the representations of sequences far from the training data are likely to look roughly like samples from the prior.
In short, typical representation learning procedures are unlikely to give sequences enough space to produce embedding kernels with discrete masses. 

Proposition~\ref{prop:scaled_embedding} suggests a simple fix: rescale the embedding.
Given an embedding $\tilde F$ that does not become more spread out with increasing sequence length, modify it to $F(X)=|\B|^{(1+\epsilon)|X|/D}\tilde F(X)$, for $\epsilon>0$.
The resulting kernel will, by the reasoning of Proposition~\ref{prop:scaled_embedding}, likely have discrete masses. 
In Section~\ref{sec:empirical_optimize} we illustrate the effectiveness of this trick in practice.
Another approach would be to use a length-dependent prior, or length-dependent regularization, when learning the embedding in the first place.

\section{Empirical Results}\label{sec: experiments}

In this section we examine the performance, on real biological sequence datasets, of some of our proposed kernels with discrete masses. We compare each to an existing kernel that relies on a similar notion of sequence similarity but lacks discrete masses.
We find that our theoretical guarantees on kernel flexibility lead to systematic improvements in performance on finite data.
Code can be found at \url{https://github.com/AlanNawzadAmin/Kernels-with-guarantees/}.

\subsection{Predicting transcription factor binding} \label{sec:empirical_regression}

We first consider a regression problem, where the aim is to predict whether a transcription factor binds a DNA sequence. 
Understanding transcription factor binding is critical for understanding and controlling gene expression.
The data consists of pairs of DNA sequences and transcription factor binding strengths, measured in terms of the intensity of a fluorescent signal in a micro-array assay~\citep{Barrera2016-im}. We examined three separate data sets from three distinct transcription factors, the homeobox proteins Hox-D13, Nkx25, and Esx1. All the DNA sequences are length 8.
We fit the data using kernel regression, and evaluate success using the root mean squared error on the training data.
Our aim is to compare kernel flexibility; more flexible kernels should provide better fits to the data.

\begin{figure}
    \centering
    \begin{subfigure}[b]{0.3\textwidth}
        \includegraphics[width=1.0\textwidth]{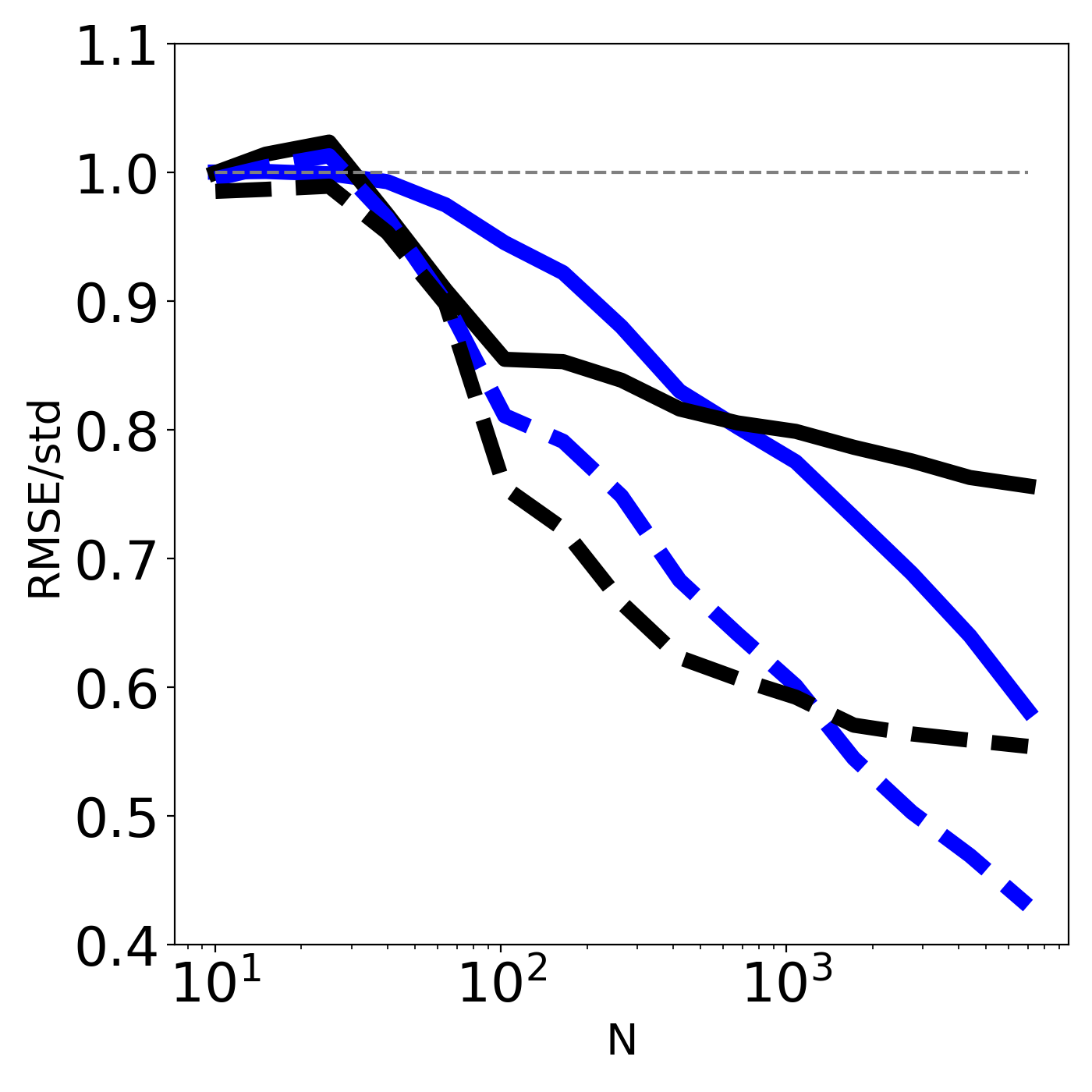}
        \caption{Esx1}
    \end{subfigure}
    \begin{subfigure}[b]{0.3\textwidth}
        \includegraphics[width=1.0\textwidth]{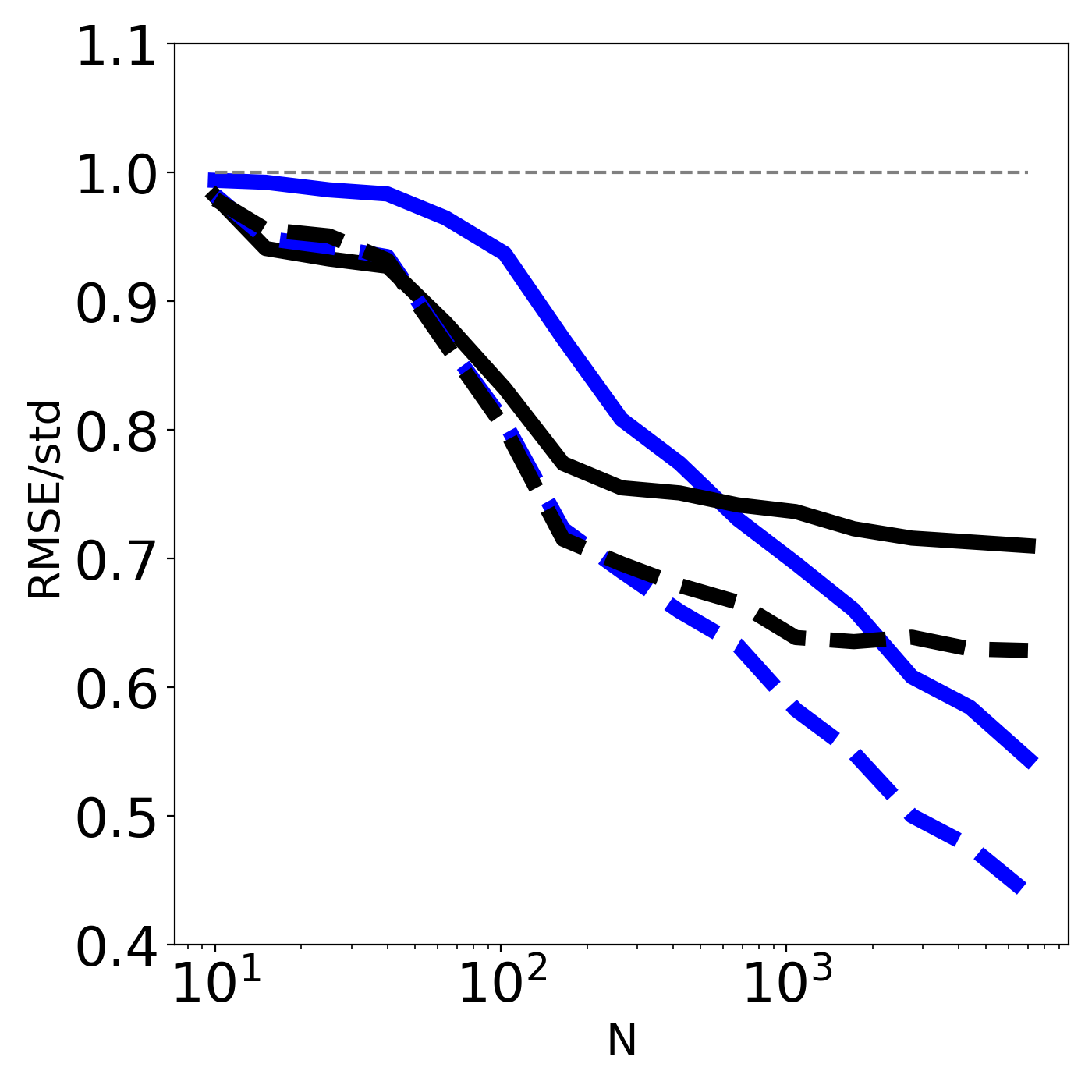}
        \caption{Nkx25} \label{fig:nkx}
    \end{subfigure}
    \begin{subfigure}[b]{0.3\textwidth}
        \includegraphics[width=1.0\textwidth]{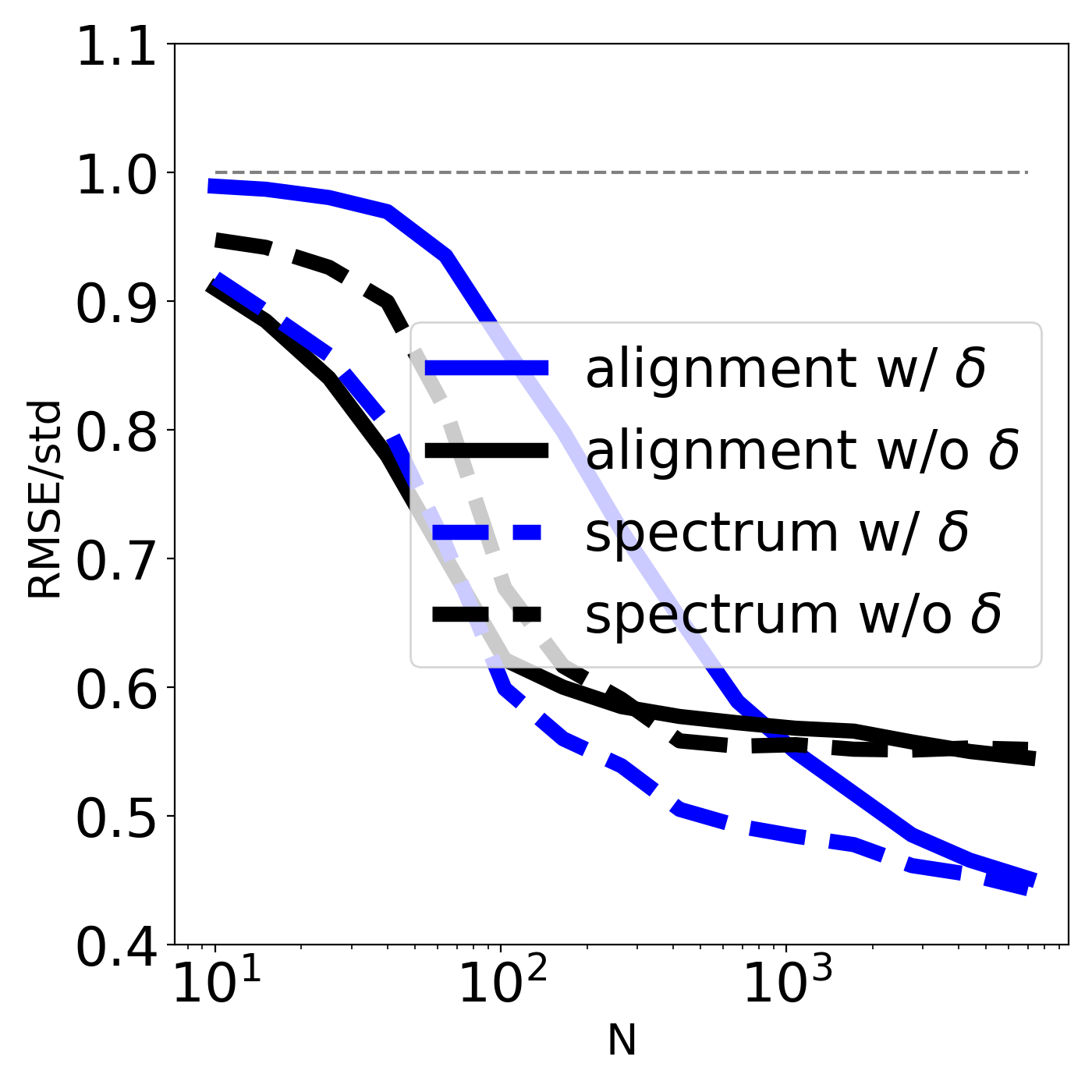}
        \caption{Hox-D13} \label{fig:hox}
    \end{subfigure}
    \caption{\textbf{Kernel regression applied to transcription factor binding} Kernel regression performance, comparing alignment and kmer spectrum kernels with and without discrete masses (``w/$\delta$'' versus ``w/o $\delta$''). $N$ is the size of the sub-sampled data set. Performance is measured by root mean squared error, normalized by the standard deviation of the outcome variable.}
    \label{fig: TF regression}
\end{figure}

We first consider alignment kernels, and, applying Theorem~\ref{thm: delta funcs string}, compare a kernel that has discrete masses ($2\mu < \log \sigma$) to one that does not ($2\mu > \log \sigma$).
We find that the version with discrete masses has better performance in the large data regime across all three example datasets (Fig.~\ref{fig: TF regression}). 
The version without discrete masses eventually starts to plateau with increasing data, while the version with discrete masses reliably approaches the correct answer in Fig.~\ref{fig:nkx} and \ref{fig:hox}.

We next consider kmer spectrum kernels, and again compare a kernel that has discrete masses (the infinite kmer spectrum kernel, Proposition~\ref{thm:infinte_kmer_spectrum}) to one that does not (the finite kmer spectrum kernel of Example~\ref{ex: spectrum kernels}, with $L = 3$).
Here again we find that the version with discrete masses has better performance in the large data regime, while the version without can plateau at a substantial error level (Fig.~\ref{fig: TF regression}).
While in the case of the alignment kernels, the version with discrete masses showed moderately worse performance in the low data regime, here the version with discrete masses performed as well or better than the version without across the full range of data set sizes considered.
In short, guaranteeing discrete masses leads to more reliable regression, especially but not exclusively in the large data regime.

\subsection{Distinguishing immune repertoires}
\begin{figure}
\centering
\includegraphics[width=0.35\textwidth]{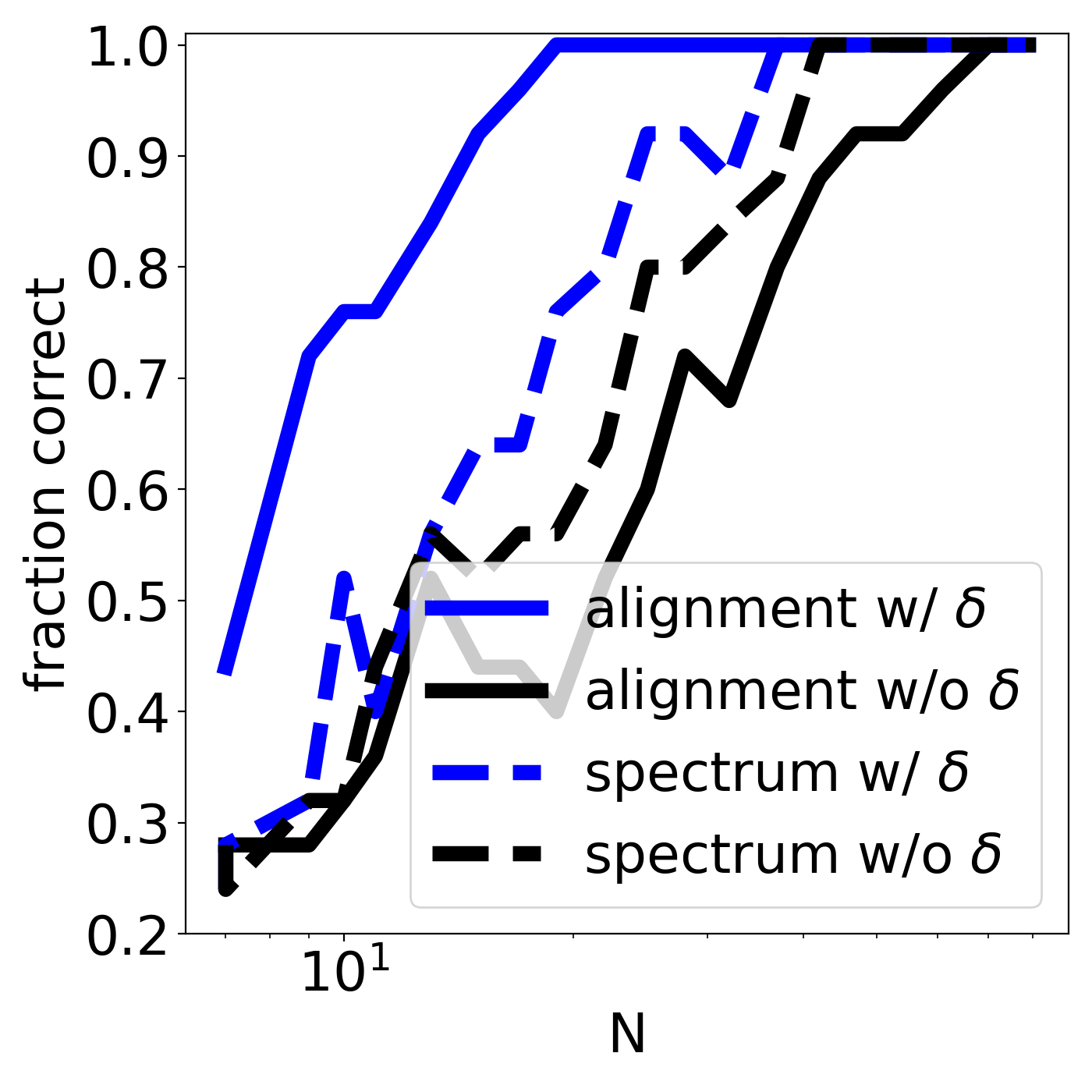}
\caption{\textbf{MMD two sample test applied to patient TCR repertoires.} MMD two-sample test performance, comparing alignment and kmer spectrum kernels with and without discrete masses (``w/$\delta$'' versus ``w/o $\delta$''). $N$ is the size of the sub-sampled data set. Performance is measured by the fraction of independent data sub-samples for which the null hypothesis is, correctly, rejected.}\label{fig: CDR3 MMD}
\end{figure}

In this section we consider a testing problem, where the aim is to distinguish between patients' T cell receptor (TCR) repertoires. 
T cells play a central role in the human adaptive immune system, recognizing foreign antigens through their TCR and then triggering an immune response.
Here, we compare two different patients' immune systems by comparing the distribution of their TCR sequences.
For each patient, we have a data set of TCR CDR3 sequences, which vary in length from 10 to 19 amino acids \citep{Genomics2022-rc}. 
We apply a two-sample test by performing a bootstrap for degenerate U-statistics on the MMD, with the null hypothesis that the patients' TCRs are drawn from the same distribution~\citep{Arcones1992-mg}.
We evaluate performance based on how often the test correctly rejects the null hypothesis at 90\% confidence, averaging over 25 random sub-samples of a large data set.
Our aim is to compare kernel flexibility; more flexible kernels should be able to distinguish between two patients' repertoires more reliably.

We apply the same four kernels from Section~\ref{sec:empirical_regression}. Although all kernels were able to distinguish the two distributions eventually with enough data, the kernels with discrete masses were able to distinguish reliably with much less data (Fig.~\ref{fig: CDR3 MMD}).
In other words, guaranteeing discrete masses improves the power of the two-sample test on real data sets.

\subsection{Optimizing representative receptors} \label{sec:empirical_optimize}

\begin{figure}
\centering
\begin{subfigure}[b]{0.3\textwidth}
\includegraphics[width=1.0\textwidth]{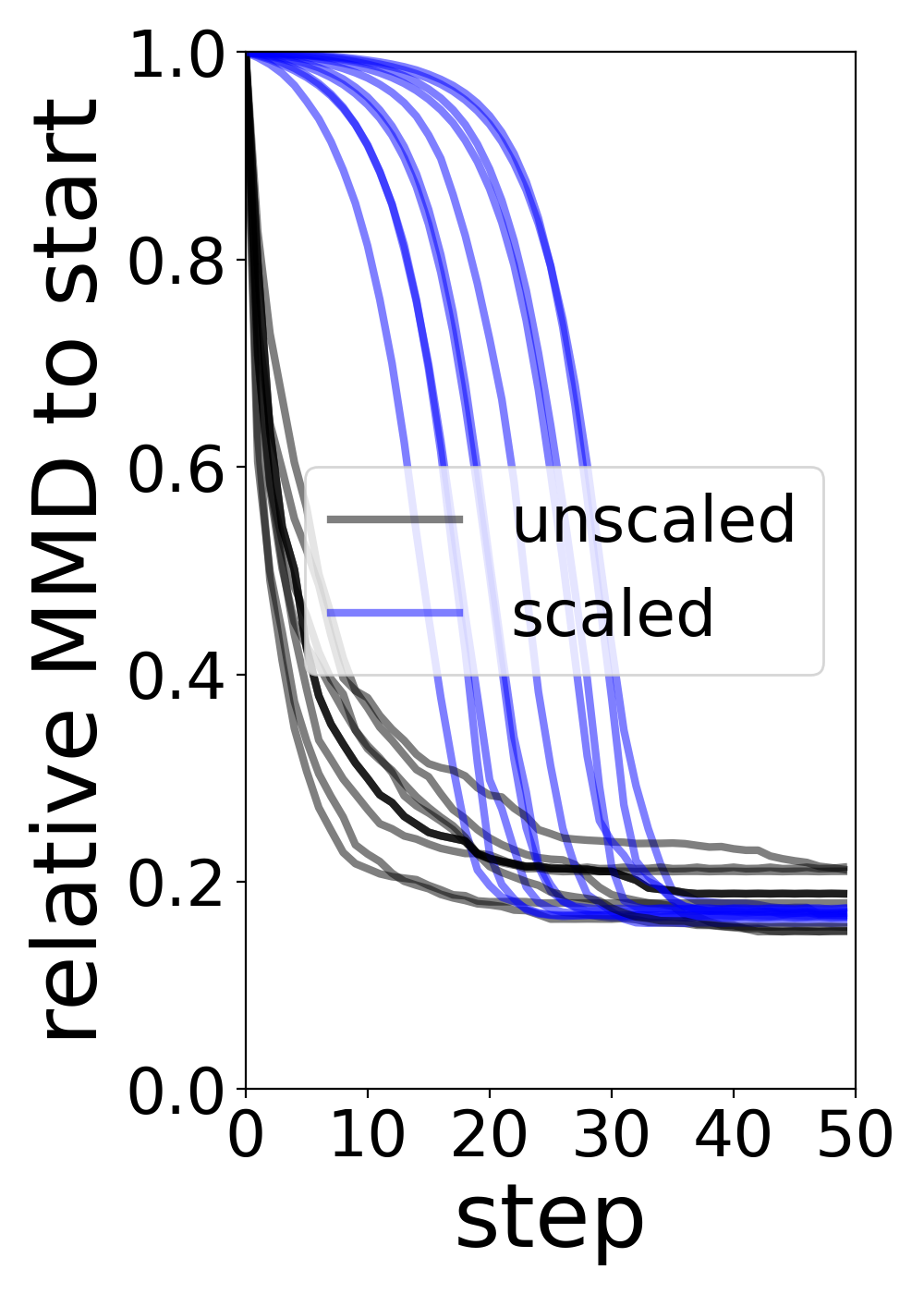}
\caption{}
\end{subfigure}
\begin{subfigure}[b]{0.3\textwidth}
\includegraphics[width=1.0\textwidth]{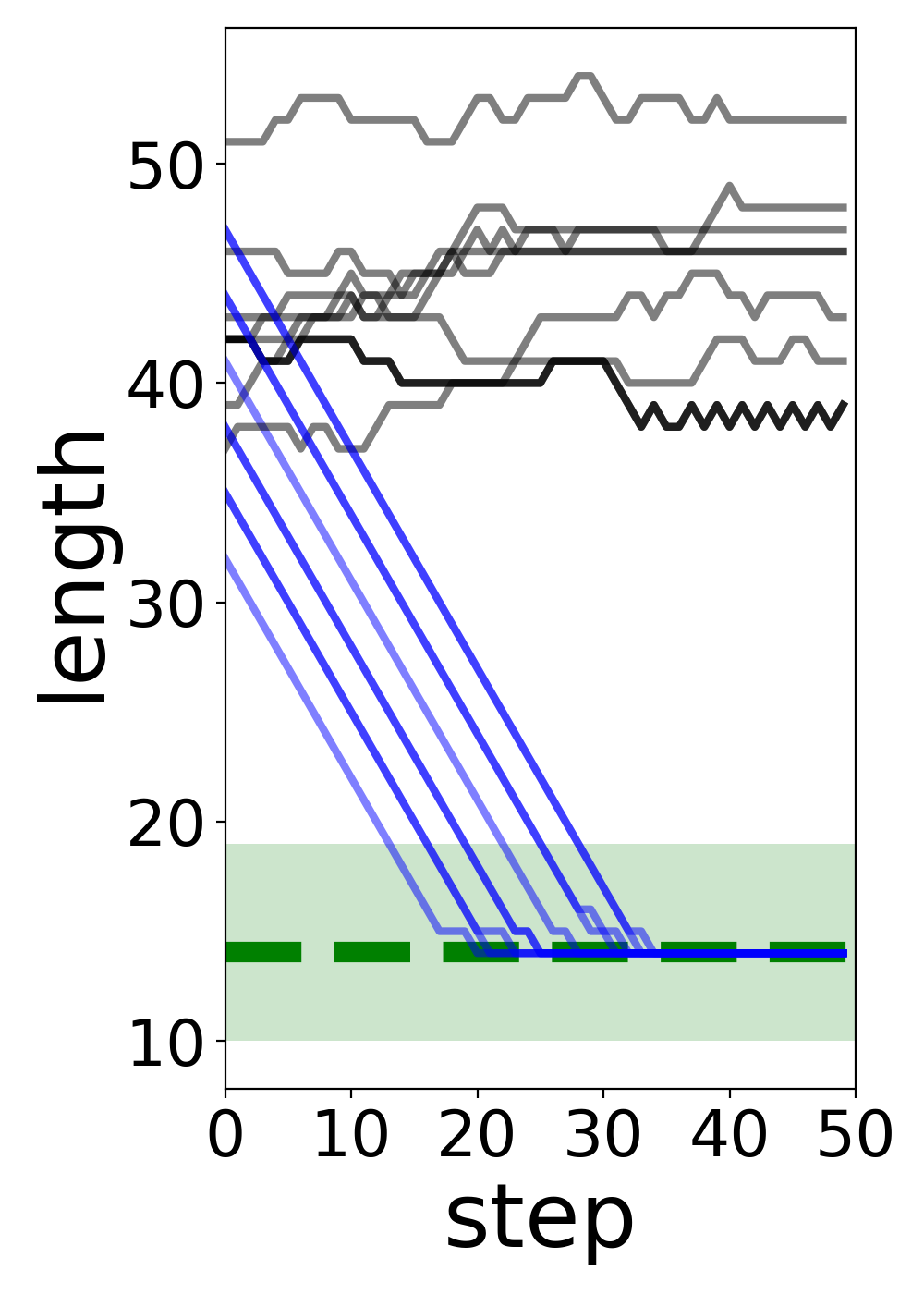}
\caption{}
\end{subfigure}
\begin{subfigure}[b]{0.3\textwidth}
\includegraphics[width=1.0\textwidth]{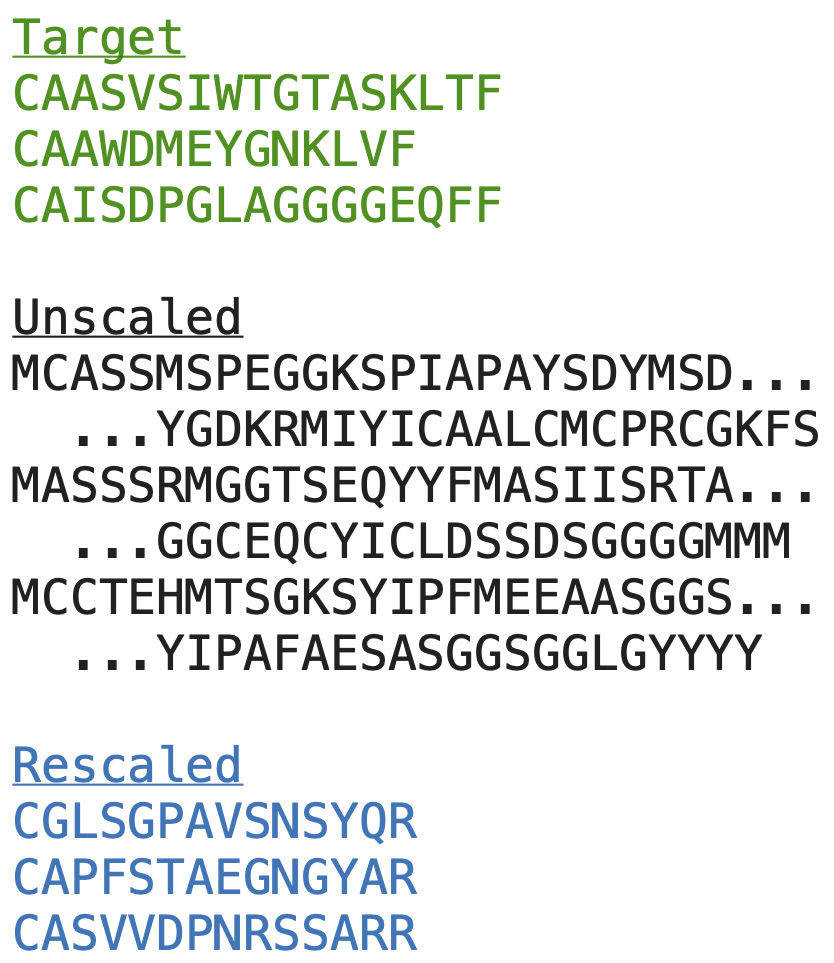}
\vspace{15pt}
\caption{}
\end{subfigure}
\caption{\textbf{Optimizing representative TCR sequences using embedding kernels.} Distribution optimization performance, comparing unscaled embedding kernels (likely to not have discrete masses) to scaled embedding kernels (likely to have discrete masses). (a) MMD per step during optimization. Here, the MMD is normalized to the MMD of the initial sequence $X$. Optimization was repeated with 10 different initial sequences. (b) Length of the optimized sequence during optimization. The maximum, minimum and average value of the target sequence lengths are shown in green. (c) Examples of sequences from the target set and of optimized sequences.}\label{fig: seq optimization}

\end{figure}

In this section we consider a distribution optimization problem, where the aim is to find a TCR sequence that is representative of a larger set.
Often biologists are interested in understanding the properties of many sequences---for instance, the TCRs from all the T cells that infiltrated a patient tumor---but have limited experimental resources, and so can only synthesize and test a small number in the laboratory.
Here, we aim to find a single sequence $X$ that is representative of a large set of TCR sequences, $Y_1, \ldots, Y_N$. We do so by optimizing $\delta_X$ to be close to the empirical distribution $p_Y = \frac{1}{N} \sum_{n=1}^N \delta_{Y_n}$, with the optimization objective $\textsc{MMD}(\delta_X, p_Y)$.
We use $N =100$ human TCR CDR3 sequences, with lengths varying between 10 and 17, as the target set~\citep{Genomics2022-rc}.
We initialize $X$ at a sequence that is longer than those in the target set, but has similar amino acid composition and arrangement (in particular, a random CDR3 from the target set, concatenated to itself). We then optimize $\textsc{MMD}(\delta_X, p_Y)$ by taking the best substitution, insertion or deletion of a single amino acid at each step, for 100 steps.
Our aim is to compare kernels' ability to metrize $\Pc(S)$. Since the main difference between the initial $X$ and the target set is its length, we are particularly interested in seeing if optimizing $\textsc{MMD}(\delta_X, p_Y)$ yields a sequence $X$ that is similar to the target set in length.

We focus on embedding kernels. We use a pre-trained embedding function $\tilde F: S \to \mathbb{R}^{64}$, ``UniRep'', learned by a deep recurrent neural network trained on a large data set of diverse protein sequences drawn from across evolution~\citep{Alley2019-yy}.
We compare embedding kernels that use $\tilde F$ directly (which are likely to be universal and characteristic by Proposition \ref{prop:embedding_injective} but are unlikely to have discrete masses, by Proposition~\ref{prop:iid_embed_no_masses}) to those that use the scaled embedding $F(X) = 20^{(1+\epsilon)|X|/64}\tilde F(X)$ with $\epsilon=0.1$ (which are likely to have discrete masses, by Proposition~\ref{prop:scaled_embedding}).
For both embedding functions, we construct an embedding kernel using the Euclidean inverse multiquadric kernel, $k_E(x, y)= (1+\Vert x-y\Vert_2^2)^{-1}$, which satisfies the conditions of Proposition~\ref{prop:embedding_injective}. For both embedding kernels, our optimization routine converges in MMD within the first 50 steps (Fig.~\ref{fig: seq optimization}A).

The unscaled kernel yields optimized sequences $X$ that are much longer than those in the target set, while the scaled kernel, which is likely to have discrete masses, yields optimized sequences $X$ at the target set's average length (Fig.~\ref{fig: seq optimization}B).
Visual inspection of the optimized $X$ also suggests that the scaled kernel produces substantially more representative designs; for instance, the optimized sequences from the scaled kernel have a cysteine in the first position, like all the target sequences, but the optimized sequences from the unscaled kernel do not (Fig.~\ref{fig: seq optimization}C).
In other words, rescaling the kernel to add discrete masses improves its ability to metrize $\Pc(S)$ in practice.

\section{Discussion}\label{sec: discussion}
In this article, we studied the flexibility of kernels for biological sequences. 
Our aim was to find kernels that are universal, characteristic, and metrize $\Pc(S)$, which guarantee they will be reliable when applied to machine learning problems involving regression, distribution comparison, and distribution optimization.
We found that many existing kernels fail to meet one or all of these criteria, but that all three are met if the kernel has discrete masses. 
We then modified existing kernels to ensure they have discrete masses, imbuing them with strong theoretical guarantees and improved empirical performance.

In applications on Euclidean space, issues of kernel flexibility are now rarely discussed, and the reliability of kernel-based methods like Gaussian processes is often taken for granted.
This is for a good reason: the default kernels implemented in packages and recommended in papers and tutorials are typically universal and characteristic at minimum (such as the Gaussian kernel) and often metrize weak convergence as well (such as the Mat\'{e}rn and inverse multiquadric kernels)~\citep{Sriperumbudur2010-ii,Sriperumbudur2011-ay,Balandat2020-pv}.
We have sought to provide a similarly stable foundation for kernel methods on biological sequence data.

We have considered a variety of different biological notions of sequence similarity, but in some settings, one may want others.
Sequences can be joined at the end, as with cyclic peptides and circular RNA, or joined into large protein complexes, made up of many separate polypeptide chains.
They can possess symmetries, like repeats, and they can be mutated in complex ways, such as in immune receptor development.
One direction for future work is to construct new kernels with discrete masses that are suited to these or other biological phenomena.
The tools we have developed for proving kernels have discrete masses (Section~\ref{sec: delta char and manip}) provide a starting point.



Another direction is to develop methods to automatically learn complex kernels from data, while preserving the discrete mass property.
Our proofs in Sections~\ref{sec: hamming kernels}-\ref{sec: embedding} demonstrate how, starting with a simple kernel with discrete masses, we can apply a series of discrete mass-preserving transformations (Section~\ref{sec: operations}) to develop a more complicated kernel with discrete masses.
A possible approach to learning new kernels is to optimize this series of transformations.

Broadly, our work contributes to the small but growing literature on the theoretical foundations of machine learning for biological sequences.
Machine learning methods for biological sequences are of wide and increasing importance to human health and the environment, and are often applied to problems where trust and reliability are essential. However, they are sometimes mistakenly thought of as mere special cases of ``general purpose'' methods, invented for other types of data.
Our results instead encourage further investigation into the specific theory of biological sequence analysis.

\newpage

\appendix
\section{Kernels with finite features are not characteristic}\label{sec: not char with finite features}

\begin{proposition}
    (proof of Proposition \ref{prop: in text char finite features})
    Consider a kernel $k(X, Y) = (\phi(X) \mid \phi(Y))$  defined using a finite feature vector, $\phi:S\to\mathbb R^d$. If $S$ is infinite, the kernel is not characteristic.
\end{proposition}
\begin{proof}
    First we will modify $\phi$ so that features do not almost always land in an affine subspace of $\mathbb{R}^d$, i.e. to avoid there being a $v\in\mathbb R^d\setminus\{0\}$ and $\beta\in \mathbb R$ such that $(v|\phi(X))=\beta$ for all but finitely many $X\in S$.
    If $\phi$ does have such a property, then there is a unitary transformation $U$ such that the first coordinate $(U\phi(X))_1=\Vert v\Vert \beta$ for all but finitely many $X\in S$.
    Define the embedding into $\mathbb R^{d-1}$, $\tilde \phi(X) = (U\phi(X))_{2:}$ so that
    $k(X, Y)=(U\phi(X)|U\phi(Y))=\Vert v\Vert ^2\beta^2 + (\tilde \phi(X)|\tilde\phi(Y))$
    for all $X, Y\in\tilde S$ where $\tilde S$ is an infinite subset of $S$.
    Calling $\tilde k(X, Y) = (\tilde \phi(X)|\tilde\phi(Y))$, note, for any distributions $\mu, \nu$,
    \begin{equation*}
    \begin{aligned}\mathrm{MMD}_k(\mu, \nu)=&E_{X\sim\mu, Y\sim\nu}\left[k(X, X)+k(Y, Y)-2k(X, Y)\right]\\
    =&E_{X\sim\mu, Y\sim\nu}\left[\tilde k(X, X)+\tilde k(Y, Y)-2\tilde k(X, Y)\right]=\mathrm{MMD}_{\tilde k}(\mu, \nu)
    \end{aligned}
    \end{equation*}
    so that $k$ is characteristic as a kernel on $\tilde S$ if and only if $\tilde k$ is.
    Redefining $k$ as $\tilde k$, $S$ as $\tilde S$, and $\phi$ as $\tilde \phi$ and repeating the above construction as many times as needed, we can assume that for any $v\in\mathbb R^d$, $\beta\in\mathbb R$, there are infinitely many $X\in S$ such that $(v|\phi(X))\neq \beta$,
    that is, $\phi(S)$ does not lie on any affine subspace of $\mathbb R^d$.

    We will now find by induction $d+1$ affinely-independent points in $\phi(S)$ to create a convex set with non-empty interior.
    Let $X_0,X_1\in S$ such that $\phi(X_1)\neq\phi(X_0)$.
    Assume we have picked $X_0, \dots, X_n$ with $\{\phi(X_i)-\phi(X_{i-1})\}_{i=1}^n$ a linearly independent set for $n<d$.
    Call $e_i=\phi(X_i)-\phi(X_{i-1})$.
    Since $\{e_i\}_{i=1}^n$ is a set of $n<d$ vectors,
    we can pick $v\in\mathbb R^d$ such that $(v|e_i)=0$ for all $i\leq n$.
    By our assumption in the paragraph above, we can pick an $X_{n+1}$ such that $(v|\phi(X_{n+1})-\phi(X_n))\neq 0$ in particular implying that $\{\phi(X_i)-\phi(X_{i-1})\}_{i=1}^{n+1}$ is linearly independent.
    Thus we have picked $\{X_i\}_{i=0}^d$ such that $\{e_i\}_{i=1}^d$ is linearly independent.
    
    Let $C$ be the convex hull of $\{\phi(X_i)\}_{i=0}^d$.
    Say $\mu_0, \dots, \mu_d$ is a set of strictly positive numbers adding up to $1$ and $w=\sum_i\mu_i\phi(X_i)\in C$.
    Note that for any $c_1, \dots, c_d\in\mathbb R^d$ with $|c_i|<\min_i \mu_i$, we have 
    \[w+\sum_i c_ie_i=(\mu_0-c_0)\phi(X_0)+(\mu_1+c_0-c_1)\phi(X_1)\dots+(\mu_d+c_d)\phi(X_d)\in C.\]
    In particular, since $\{e_i\}_i$ is a basis of $\mathbb R^d$, $w$ is in the interior of $C$.
    Now let $Z\in S$ have $Z\neq X_i$ for all $i$.
    For small enough $\epsilon>0$, $\epsilon\phi(Z) + (1-\epsilon)w\in C$, in particular there is a set of positive numbers adding up to $1$, $\nu_0, \dots, \nu_d$, such that
    $$\epsilon\phi(Z)+\sum_i(1-\epsilon)\mu_i\phi(X_i)=\sum_i\nu_i\phi(X_i).$$
    Thus, $k$ is not characteristic: the kernel embedding of the distributions $\epsilon \delta_Z + \sum_i (1-\epsilon) \mu_i \delta_{X_i}$ and $\sum_i \nu_i \delta_{X_i}$ are the same.
\end{proof}

\section{The Hamming kernel of lag $L$ fails to distinguish distributions} \label{sec:mmd_test_sims}
Kernels that are not characteristic are limited in the distributions they can distinguish. Here, we illustrate this in simulation, using the Hamming kernel of lag $L$ (Section \ref{sec: toy example}). 

We consider two distributions over DNA sequences of length 10:
the first is a uniform distribution;
the second is uniform over the first 5 letters, while the last 5 letters are equal to the first 5.
We evaluate the ability of an MMD two-sample test to distinguish these two distributions, using different kernels.
The first kernel is the Hamming kernel of lag $L=2$, which is not characteristic.
The other kernels are the exponential Hamming kernel (with lag $L=1$) and the inverse multi-quadric Hamming kernel (with lag $L=1$), which have discrete masses and so are characteristic (Section \ref{sec: hamming kernels}).
For increasing amounts of data sampled from each of the two distributions, we apply an MMD two-sample test with each kernel; the test uses a bootstrap for degenerate U-statistics, with the null hypothesis that the two distributions are the same~\citep{Arcones1992-mg}.
We set a confidence threshold of 95\% for rejecting the null hypothesis. 
A good test should reliably reject the null hypothesis with only a small amount of data.

\begin{figure}
\centering
\includegraphics[width=0.4\textwidth]{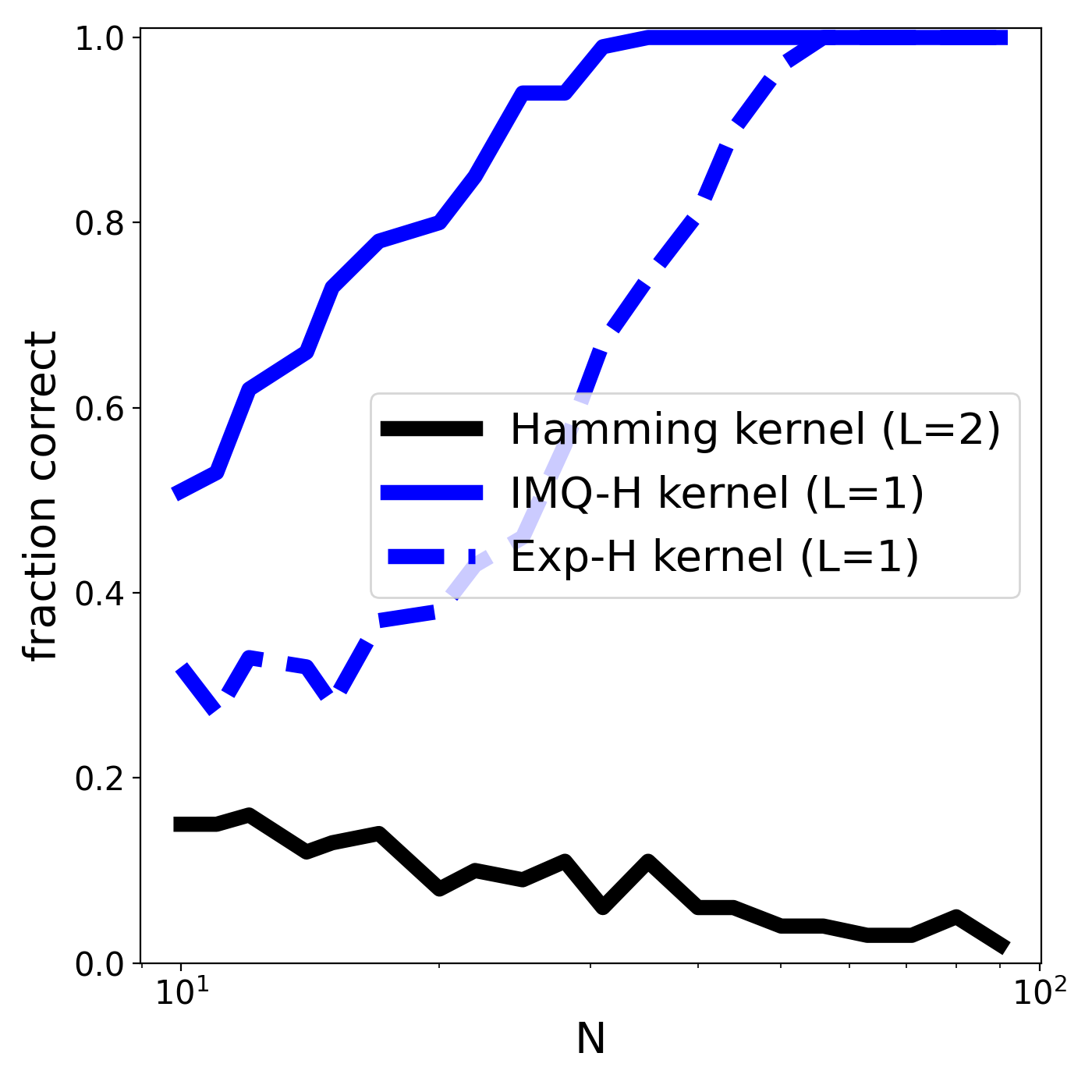}
\caption{MMD two sample test applied to simulated data. We compare the Hamming kernel of lag $L=2$, the exponential Hamming kernel (Exp-H) and the inverse multiquadric Hamming kernel (IMQ-H). Only the last two have discrete masses.}\label{fig: toy MMD}
\end{figure}

Figure \ref{fig: toy MMD} plots the fraction of times the two-sample test correctly rejects the null hypothesis (the test's power), out of 100 independent draws of the data.
We can see that no matter how much data one has, the Hamming kernel with lag $L$ fails to reliably distinguish the two distributions; it has asymptotically zero power.
However, given enough data, the exponential and IMQ-H kernels can reliably distinguish the distributions.
In short, a small modification to the standard Hamming kernel, imbuing it with discrete masses, provides a guarantee against failure.

\section{A universal kernel that metrizes the space of distributions but lacks discrete masses}\label{sec: mmd but not delta}

    Here we show that having discrete masses is not strictly necessary in order to have a kernel that is universal or metrizes $\Pc(S)$.
    Recall Example \ref{ex:embedding_metrize}, where the kernel failed to metrize $\Pc(S)$ because $k_A$ could be well approximated by $k_{n\times A}$ in $\Hc_k$.
	To obtain a universal kernel that metrizes $\Pc(S)$ but lacks discrete masses we will create a kernel on the integers $\{0, 1, -1, 2, -1, \dots\}$ such that $k_0$ can be approximated by $\int d\nu_n(X)k_X$ for $\nu_n\in\Pc(S)$ but only if the total variation of $\nu_n$ blows up with $n$.

	Let $\Hc_k$ be a Hilbert space with an orthonormal basis indexed by the integers $\{e_0, e_1, e_{-1}, \dots\}$.
	Define $k_0=e_0$ and for any positive $n$, $k_n=e_n$ and $k_{-n}=\sqrt{1-\epsilon_n^2-\epsilon_n^4}e_n + \epsilon_n^2 e_{-n} + \epsilon_n e_0$ for a decreasing sequence of positive $\epsilon_n$ that has $\epsilon_n\to 0$.
	Clearly $k$ is a strictly positive definite kernel on the integers and $k(z, z)=1$ for all integers $z$.
	
	$\Hc_k$ cannot contain $\delta_0$.
	To see this, say that $\delta_0\in\Hc_k$.
	Then 
	\begin{equation*}
	\begin{aligned}
	0=&\frac 1 {\epsilon_n}\left(\delta_0(-n)-\delta_0(n)\right)\\
	=&\left(\delta_0\bigg|\frac 1 {\epsilon_n}\left(k_{-n} - k_{n}\right)\right)_k\\
	=& \left(\delta_0\bigg|\frac{\sqrt{1-\epsilon_n^2-\epsilon_n^4}-1}{\epsilon_n}e_n +\epsilon_n e_{-n} + e_0\right)_k\\
	\to&\left(\delta_0| e_0\right)_k=\delta_0(0)=1,
	\end{aligned}
	\end{equation*}
	a contradiction.

    However, $k$ is $C_0$-universal.
    To see this, say for some finite (possibly signed) measures on $S$, $\mu, \nu$, we have $\sum_z \nu(z) k_z = \sum_z \mu(z) k_z$.
    Call $\pi=\nu-\mu$.
    Note $0=(\sum_z \pi(z) k_z|e_{-n})_k=\epsilon_{n}^2\pi(-n)$ and $0=(\sum_z \pi(z) k_z|e_{n})_k=\pi(n) + \pi(-n)\sqrt{1-\epsilon_n^2-\epsilon_n^4}$.
    Thus $\pi(n)=0$ for all $n\neq 0$.
    Finally, $0=(\sum_z \pi(z) k_z|e_{0})_k=\pi(0)$ so $\pi = 0$, i.e. $\mu=\nu$.
    By Proposition 2 of \citet{Sriperumbudur2011-ay} $k$ is universal.

	$k$ also metrizes $\Pc(S)$. To see this,
	say for some $\mu, \nu_1, \nu_2, \dots\in\Pc(S)$, we have $\sum_z \nu_{m}(z) k_z \to \sum_z \mu(z) k_z$.
	Call $\pi_m=\nu_m-\mu$.
    As above, $(\sum_z \pi_{m}(z) k_z|e_{-n})_k=\epsilon_{n}^2\pi_{m}(-n)$ and $(\sum_z \pi_{m}(z) k_z|e_{n})_k=\pi_{m}(n) + \pi_m(-n)\sqrt{1-\epsilon_n^2-\epsilon_n^4}$.
    For each $n$, the left hand sides of both of these equations go to $0$ as $m\to\infty$ so, $\pi_m(n)\to 0$ for all $n\neq 0$.
    Finally note
    $$\left(\sum_z \pi_{m}(z) k_z\bigg|e_{0}\right)_k = \pi_m(0) + \sum_{n=1}^\infty \pi_m(-n)\epsilon_n.$$
    The left hand side goes to $0$ and since $\sum_z |\pi_m(z)|\leq 2$ and $\epsilon_n\to 0$, the sum in the right hand side also goes to $0$.
    Thus $\pi_m(0)\to 0$ and so $\nu_n\to \mu$.

	Interestingly, although $\delta_0 \notin \Hc_k$, $k$ is still able to approximate $\delta_0$ arbitrarily well in infinity norm.
	First pick, for each $n$, an $f_n\in \mathrm{span}\{k_0, k_n, k_{-n}\}$ such that $(f_n|k_0)=(f_n|k_n)=0$ and $(f_n|k_{-n})=1$.
	Then
	$g_n=k_0-\sum_{n=1}^N \epsilon_nf_n$
	is $k_0$ on $\{0, N+1, -(N+1), N+2, -(N+2), \dots\}$ and $0$ on $\{1, -1, \dots, N, -N\}$.
	As $k_0$ vanishes at infinity, $g_n$ converges to $\delta_0$ in infinity norm as $N\to \infty$.
	However, despite approximating $\delta_0$ in infinity norm, $(g_n)_n$ does not approximate $\delta_0$ in $\Hc_k$: $\Vert g_n\Vert_k$ grows unboundedly.\footnote{
    To see why $\Vert g_n\Vert_k$ must blow up, note that if this is not the case, then by the Banach-Alaoglu theorem a subsequence $(g_{n_k})_k$ must ``weakly converge'' to a $g\in \Hc_k$, that is, $g_{n_k}(z)=(g_{n_k} | k_z)_k\to (g|k_z)_k=g(z)$ for all $z$.
    Thus, $\delta_0=g\in\Hc_k$, a contradiction.
    }

\section{RKHSs of manipulated kernels}\label{sec: trans proofs}
Here we provide proof of the propositions made in Section~\ref{sec: delta char and manip}
\begin{proposition} \label{prop: tilt delta}
	$f\mapsto Af$ is a unitary isomorphism of $\Hc_k$ to $\Hc_{k^A}$.
\end{proposition}
\begin{proof}
	Define a linear transformation on $\mathrm{span}\{k_Y\}_{Y\in S}$, $T$, that takes $k_x$ to $A(X)^{-1}k_x^A$.
	$T$ is well defined as if $\sum_{n=1}^N\alpha_nk_{X_n}=0$ for some $(X_n)_n$, $(\alpha_n)_n$, then
	$(\sum_{n=1}^NA(X_n)^{-1}\alpha_nk^A_{X_n}|k_Z^A)_{k^A}=A(Z) \sum_{n=1}^N\alpha_nk(X_n, Z)=0$ for all $Z$.
	$T$ is also unitary as $(Tk_X|Tk_Y)_{k^A}=k(X, Y)=(k_X|k_Y)_{k}$.
	Thus, $T$ can be extended to all of $\Hc_{k^A}$.
	
	Note $(Tk_X|k^A_Y)_{k^A}=k(X, Y)A(Y)=(k_X|k_Y)_{k}A(Y)$ for all $X, Y\in S$.
	By linearity and continuity for all $f\in\Hc_k$, $Tf(Y)=(Tf|k^A_Y)_{k^A}=(f|k_Y)_{k}A(Y)=f(Y)A(Y)$.
	
	We can see that $T$ is surjective by the fact that it has inverse $f\mapsto A^{-1}f$.
\end{proof}

\begin{proposition} \label{prop: tensor kern}
	Let $k_1, k_2$ be kernels on $S$. Define
	$$k_1\otimes k_2((X_1, X_2), (Y_1, Y_2))=k_1(X_1, Y_1)k_2(X_2, Y_2)$$
	for $(X_1, X_2), (Y_1, Y_2)\in S^2$.
	Then $\Hc_{k_1\otimes k_2}=\Hc_{k_1}\otimes\Hc_{k_2}$.
	In particular, if $f_1\in \Hc_{k_1}, f_2\in \Hc_{k_2}$, there is a $f_1\otimes f_2\in \Hc_{k_1\otimes k_2}$ such that
	for $(X_1, X_2)\in S^2$,
	\begin{equation}\label{eq: tensor}
	f_1\otimes f_2(X_1, X_2)=f_1(X_1)f_2(X_2).
	\end{equation}
\end{proposition}
\begin{proof}
	For $f_1=\sum_{n=1}^{N_1}\alpha_{1, n}k_{1, X_{1, n}}\in \Hc_1, f_2=\sum_{n=1}^{N_2}\alpha_{2, n}k_{2, X_{2, n}}\in \Hc_2$, define
	\begin{equation}\label{eq: tensor dot}
	f_1\otimes f_2=\sum_{n=1}^{N_1}\sum_{m=1}^{N_2}\alpha_{1, n}\alpha_{2, m}\left(k_{1}\otimes k_{2}\right)_{(X_{1, n}, X_{2, m})}\in \Hc_{k_1\otimes k_2}.
	\end{equation}
	Clearly if $f_1'\in\text{span}(k_{1, X})_{X\in S}, f_2'\in\text{span}(k_{2, X})_{X\in S}$, then 
	$(f_1\otimes f_2|f'_1\otimes f'_2)_{k_{1}\otimes k_{2}} = (f_1| f'_1)_{k_1}(f_2|f_2')_{k_2}$.
	We can thus continuously extend the bilinear function $\otimes$ to all $f_1\in\Hc_{k_1}, f_2\in\Hc_{k_2}$.
	$\Hc_{k_1}\otimes\Hc_{k_2}$ thus embeds in $\Hc_{k_1\otimes k_2}$ and actually embeds surjectively as $\{k_{1, X_1}\otimes k_{2, X_2}\}_{(X_1, X_2)}=\{\left(k_{1}\otimes k_{2}\right)_{(X_{1}, X_{2})}\}_{(X_1, X_2)}$ has a dense span.
	Finally, Equation \ref{eq: tensor} follows from $k_{1, X_1}\otimes k_{2, X_2}=\left(k_{1}\otimes k_{2}\right)_{(X_{1}, X_{2})}$,
	Equation \ref{eq: tensor dot} and continuity.
\end{proof}

\begin{proposition}~\label{prop: restrict kern}
	Say $k$ is a kernel on $S$ and $S'\subseteq S$.
	Define $k|_{S'}$ as the restriction of $k$ to $S'\times S'$.
	Then $\Hc_{k|_{S'}}$ is the closure of $\text{span}\{k_Y\}_{Y\in S'}$ in $\Hc_k$.
\end{proposition}
\begin{proof}
    Let $T:\mathrm{span}((k|_{S'})_X)_{X\in S'}\to \Hc_{k}$ be a linear mapping that takes $(k|_{S'})_X$ to $k_X$ for $X\in S'$.
    $T$ is well defined as if $f=\sum_{n=1}^N\alpha_n(k|_{S'})_{X_n}=0$ for some $(X_n)_n$, $(\alpha_n)_n$, then
    $$\Vert Tf\Vert_k=\sum_{n=1}^N\sum_{m=1}^N\alpha_n\alpha_m k(X_n, X_m)=\sum_{n=1}^N\sum_{m=1}^N\alpha_n\alpha_m k|_{S'}(X_n, X_m)=\Vert f\Vert_{k|_{S'}}=0.$$
    $T$ is also obviously unitary and thus can be extended to a unitary mapping on all of $\Hc_{k|_{S'}}$.
    Clearly the image of $T$ is the closure of $\text{span}\{k_Y\}_{Y\in S'}$ in $\Hc_k$.
\end{proof}

\section{Discrete masses in the alignment kernel}\label{sec: proof of ali discrete masses}
In this section we prove the alignment kernel has discrete masses under certain parameter settings.

We first establish some properties of sequences with encodings that are not one-hot.
In particular, we show that convolution of kernels works in the same way when we reparameterize the alphabet.
Note, for $V, W\in \cup_{L=0}^\infty \mathbb R^{L\times \B}$, we write $V_{(l)}, V_{(l:l')}$ to index positions along the first dimension and $V+W$ for concatenation along the first dimension, in analogy to the definitions for regular sequences in $S=\cup_{L=1}^\infty \B^L$.
\begin{proposition}\label{prop: reparam and conv}
	Say $k_1, k_2$ are kernels on $S$ and $V\in \mathbb R^{L\times \B}, V'\in \mathbb R^{L'\times \B}$.
	\[k_1\star k_2(V, V')=\sum_{l_1+l_2=L, l'_1+l_2'=L'} k_1\left(V_{(:l_1)}, V'_{(:l_1')}\right)k_2\left(V_{(l_1:)}, V'_{(l_1':)}\right).\]
\end{proposition}

\begin{proof}
	\begin{equation}
	\begin{aligned}
	k_1&\star k_2(V, V')\\
	=&\sum_{|X|=L}\sum_{|Y|=L'}\left( \prod_{l=0}^{L-1}V_{l, X_{(l)}}\right)\left( \prod_{l=0}^{L'-1}V_{l, Y_{(l)}}\right)\sum_{l_1, l'_1} k_1(X_{(:l_1)}, Y_{(:l_1')})k_2(X_{(l_1:)}, Y_{(l_1':)})\\
	=&\sum_{l_1, l'_1}\sum_{|X|=L}\sum_{|Y|=L'}\left(\left( \prod_{l=0}^{l_1-1}V_{l, X_{(l)}}\right)\left( \prod_{l=0}^{l_1'-1}V_{l, Y_{(l)}}\right) k_1(X_{(:l_1)}, Y_{(:l_1')})\right)\\
	&\ \ \ \ \ \ \ \ \ \ \ \ \ \ \ \ \ \ \ \ \times \left(\left( \prod_{l=l_1}^{L-1}V_{l, X_{(l)}}\right)\left( \prod_{l=l_1'}^{L'-1}V_{l, Y_{(l)}}\right) k_2(X_{(l_1:)}, Y_{(l_1':)})\right)\\
	=&\sum_{l_1, l'_1}\left(\sum_{|X|=l_1}\sum_{|Y|=l_1'}\left( \prod_{l=0}^{l_1-1}V_{l, X_{(l)}}\right)\left( \prod_{l=0}^{l_1'-1}V_{l, Y_{(l)}}\right) k_1(X, Y)\right)\\
	&\ \ \ \ \ \ \times \left(\sum_{|X|=L-l_1}\sum_{|Y|=L'-l_1'}\left( \prod_{l=0}^{L-l_1-1}V_{l+l_1, X_{(l)}}\right)\left( \prod_{l=0}^{L'-l_1'-1}V_{l+l_1', Y_{(l)}}\right) k_2(X, Y)\right)\\
	=&\sum_{l_1, l'_1} k_1\left(V_{(:l_1)}, V'_{(:l_1')}\right)k_2\left(V_{(l_1:)}, V'_{(l_1':)}\right).
	\end{aligned}
	\end{equation}
\end{proof}

We now prove Theorem \ref{thm: delta funcs string}.
\begin{theorem} (proof of Theorem \ref{thm: delta funcs string})
    Define $K$ as the matrix with $K_{b', b}=k_s(b, b')$ for $b, b'\in\B$, and $\sigma=\mathbf{1}_{\B}^T K^{-1} \mathbf{1}_{\B}$.
	If $\Delta\mu>0$, then $k$ has discrete masses if and only if $2\mu \geq \log\sigma$.
	If $\Delta\mu=0$, then $k$ has discrete masses if and only if $2\mu > \log\sigma$.
\end{theorem}
\begin{proof}
	The proof is organized as follows.
	First, we reparameterize $\B$ such that $\Hc_k$ decomposes into a set of orthogonal hyperplanes $W_V$.
	Second, we show that each of these hyperplanes is isomorphic to a tensor product of RKHSs of the alignment kernel with $|\B|=1$.
	Finally, the result follows from Theorem~\ref{thm: delta funcs string B=1}, which establishes conditions under which alignment kernels with $|\B|=1$ have discrete masses.
	
	\textit{Part 1:}
	We start by constructing the reparameterized alphabet. Call $\tilde U=K^{-1}\mathbf{1}_{\B}$, where $\mathbf{1}_{\B}\in\mathbb R^{\B}$ is the vector with $1$ in each position,
    and let $U=\tilde U/k_s(\tilde U, \tilde U)$, so, $U^T\mathbf{1}_{\B}=\tilde U^TK\tilde U/\tilde U^TK\tilde U=1$ and $k_s(U, U)=(\tilde U^T K\tilde U)^{-1}=k_s(\tilde U, \tilde U)^{-1}=(\mathbf{1}_{\B}^T K^{-1} \mathbf{1}_{\B})^{-1}=\sigma^{-1}$.
    Let $B_1, \dots, B_{|\B|-1}\in\mathbb R^{\B}$ be chosen so that $\{U, B_1, \dots, B_{|\B|-1}\}$ is an orthogonal basis of the vector space $\mathbb R^{\B}$ with the dot product $(v|x)=v^TKw$, with $(B_i|B_i)=1$ for all $i$.
    Thus, $k_s(B_i, B_j)=\delta_i(j)$ for all $i, j$ and $k_s(U, B_i)=0$ for all $i$.
	Now call $\tilde \B=\{U, B_1, \dots, B_{|\B|-1}\}$.
	We will use $\tilde \B$ as a reparameterized alphabet and note that by Proposition \ref{prop: reparam delta} the alignment kernel has discrete masses if and only if it has discrete masses when $\B$ is replaced with $\tilde\B$.
 
	Now, for any number $L$ and $V\in\tilde\B^L$, the insertion kernel $k_I$ applied to $V$ is,
	$$k_{I, V}=\sum_{|X|=L}\left( \prod_{l=0}^{L-1}V_{l, X_{(l)}}\right)e^{-\Delta\mu-L\mu}g =e^{-\Delta\mu-L\mu} \prod_{l=0}^{L-1}\left(\sum_{b\in\B}V_{l, b}\right)g.$$
    Note for any $B \in \{B_1, \dots, B_{|\B|-1}\} \subset \tilde \B$, we have $\sum_{b\in \B} B_{b}=B^T\mathbf 1_{\B}=B^TK\tilde U=k_s(B, U)/k_s(U, U)=\delta_U(B)$.
	Thus, $k_{I, V}=0$ if any letters in $V$ are in $\tilde \B\setminus\{U\}$.
	Now say $V\in\tilde \B^L$, $V'\in\tilde \B^{L'}$.
    By Proposition \ref{prop: reparam and conv}, $k(V, V')$ is given by
	$$\sum_{l\in \mathbb{N}, V^{(1)}+ \dots+ V^{(2l+1)}=V, V^{'(1)}+ \dots+ V^{'(2l+1)}=V'} \!\!\!\!\!\!\!\!k_I(V^{(1)}, V^{'(1)})\prod_{i=1}^l k_s(V^{(2i)}, V^{'(2i)})k_I(V^{(2i+1)}, V^{'(2i+1)}).$$ 
	Each term in the sum will be non-zero only if $V^{(m)} $ and $ V^{'(m)}$ are composed only of the letter $U$ for all odd $m$, and $V^{(m)}=V^{'(m)}$ and $|V^{(m)}| = 1$ for all even $m$.
 In particular, if $V \in (\tilde \B\setminus\{U\})^L$, sequences $V'$ that satisfy $k(V, V') \neq 0$ must take the form
 $$(t_0\times U)+V_{(0)}+(t_1\times U)+V_{(1)}+\dots+V_{(L-1)}+(t_L\times U)$$
 for some integers $t_0, \dots, t_L$. Call this sequence $V[t]$.
 Let $W_V \subset S$ denote the set of all such sequences, i.e. $W_V = \{V[t] \ |\ t_0, \dots, t_N\in\mathbb{N}\} $.
We see that $W_V$ is orthogonal to any $W_{V'}$ for which $V' \notin W_V$. Each $W_V$ thus defines an orthogonal hyperplane. Moreover, the union of all $\{W_V\}_V$ is the set of all sequences with letters in $\tilde \B$.
Proposition~\ref{cor: ortho decomp delta} therefore says that if the alignment kernel has discrete masses when restricted to each $W_V$, it has discrete masses. 

	\textit{Part 2:} In this section we find that, for the alignment kernel to have discrete masses over $W_V$, it is sufficient that a simpler alignment kernel, with an alphabet size of just one, have discrete masses.
	Say $V[t], V[t']\in W_V$ for some $V \in (\tilde \B\setminus\{U\})^L$. 
 Let $V[t_{1:}]$ denote $V[t]$ with the first $t_0+1$ letters dropped.
	Since the only alignments that have positive weight are those that align the $i$-th set of $U$s in $V[t]$ to the $i$-th set of $U$s in $V[t']$ for all $i$, we can decompose the alignment kernel into a product of alignment kernels comparing each of the strings of $U$s:
	\begin{equation}\label{eqn: tesor decom string}
	\begin{aligned}
	k(V[t], V[t'])=&\sum_{{\substack{l\in\mathbb N\\ Y^{(1)}+ \dots+ Y^{(2l+1)}=t_0\times U \\ Z^{(1)}+ \dots+ Z^{( 2l+1)}=t'_0\times U}}}\sum_{{\substack{l'\in\mathbb N \\ \tilde{Y}^{(1)}+ \dots+ \tilde{Y}^{(2l'+1)}=V[t_{1:}] \\ \tilde{Z}^{(1)}+ \dots+ \tilde{Z}^{(2l'+1)}=V[t'_{1:}]}}}\Bigg(\\
	&\times\left(k_I(Y^{(1)}, Z^{(1)})\prod_{i=1}^l k_s(Y^{(2i)}, Z^{(2i)})k_I(Y{(2i+1)}, Z^{(2i+1)})\right)\\
    &\times k_s(V_{(0)}, V_{(0)})\\
    &\times\left(k_I(\tilde Y^{(1)}, \tilde Z^{(1)})\prod_{i=1}^{l'} k_s(\tilde Y^{(2i)}, \tilde Z^{(2i)})k_I(\tilde Y^{(2i+1)}, \tilde Z^{(2i+1)})\right)\Bigg)\\
	=& k(t_0\times U, t'_0\times U) k(V[t_{1:}], V[t'_{1:}])\\
	=&\dots\\
	=& \prod_{l=0}^L k(t_l\times U, t'_l\times U).
	\end{aligned}
	\end{equation}
 
	Now call $k_1$ the alignment kernel when $\B=\{A\}$, with parameter $k_s(A, A)=\sigma^{-1}$.
	Since $k_s(U, U)=\sigma^{-1}$, we have $k_1(l\times A,l'\times A)=k(l\times U,l'\times U)$.
	Thus $k$ on $W_V$ acts as the $L+1$-times tensor product of $k_1$.
	Thus, by Proposition \ref{prop: tensor delta}, if $k_1$ has discrete masses, so does $k$ on $W_V$ for every $V$, so that $k$ has discrete masses on $S$; on the other hand, if $k_1$ does not have discrete masses, neither does $k$ on $W_\emptyset$, so that $k$ does not have discrete masses on $S$.

    Finally, in Theorem \ref{thm: delta funcs string B=1} we will see that if $|\B|=1$ then if $\Delta\mu>0$, $k_1$ has discrete masses if and only if $2\mu + \log k(A, A) \geq 0$.
    Thus if $\Delta\mu>0$, we find that $k_1$, and therefore $k$, has discrete masses if and only if $2\mu \geq \log\sigma$.
    The result similarly follows for $\Delta\mu=0$.
\end{proof}

Note that this proof also extends to alignment kernels that have different penalties for inserting different letters.
More precisely, we can modify $k_I$ such that it differs for different letters by using
	$\tilde g(X)=\left(\prod_{l=0}^{|X|-1}t(X_{(l)})\right) g(X)$ for a function $t\in\mathbb R^\B$.
    Above we chose $t=\mathbf{1}_{\B}$.
	For general $t$ define $U=K^{-1}t/ t^TK^{-1}t$ and define $\tilde\B=\{U, B_1,\dots, B_{|\B|-1}\}$ the same way.
    Again, for any $B\in\tilde \B$, ${\sum_b t(b)B_b}=k_s(B, U)/k_s(U, U)=\delta_{U}(B)$,
	so, $k_{I, V}=0$ for any sequence $V$ that has any letters orthogonal to $U$. 
	Calling $\sigma = t^TK^{-1}t$, we can repeat the proof of theorem \ref{thm: delta funcs string} to reach the same conclusion.
\section{Interpreting the conditions for discrete masses in alignment kernels}\label{sec: ali with base spec}

In Theorem \ref{thm: delta funcs string} we saw that the alignment kernel has discrete masses only under certain conditions.
An interesting aspect of these conditions is that they involve an interplay between the the kernel's insertion parameters (through $\mu$) and the substitution parameters (through $\sigma$). Often these parameters are thought of as entirely distinct; after all, biologically, substitution and insertion mutations can stem from quite distinct molecular processes. 
Here we build some intuition for their interplay in determining discrete masses.
	We focus on the special case where $\Delta\mu = 0$.

    The key idea is to rewrite the alignment kernel as a sum over a particular position-wise comparison kernel applied to each pairwise alignment; then, the conditions for the alignment kernel to have discrete masses turn out to be the same as the conditions for the corresponding position-wise kernel to have discrete masses.
    In detail, recall that Equation~\ref{eqn:pairwise_alignment_terms} decomposes the alignment kernel into a sum over pairwise alignments.
    For example, consider the alignment between $X = \verb|ATGC|$ and $Y=\verb|ACC|$ we saw in Section~\ref{sec: the string kernel}.
    We can rewrite the corresponding term in Equation~\ref{eqn:pairwise_alignment_terms} in the form
    $$k'_s(\verb|A|, \verb|A|) k'_s(\verb|T|, \verb|C|), k'_s(\verb|G|,\verb|-|) k'_s(\verb|C|, \verb|C|) k'_s(\verb|-|,\verb|-|) k'_s(\verb|-|,\verb|-|) \ldots$$
    where we define the extended letter comparison kernel $k'_s$ as $k'_s(b,b') = k_s(b,b')$ for $b,b' \in \B$, $k'_s(b, \verb|-|) = e^{-\mu}$ for $b \in \B$, and $k'_s(\verb|-|,\verb|-|) = 1$.
    This matches the form of a position-wise comparison kernel (Def.~\ref{def:position-wise}) between the sequences \verb|ATGC--...| and \verb|AC-C--...|, with letter kernel $k'_s$, and where the gap symbol is also the stop symbol.
    In general, every term in the sum making up the alignment kernel (Equation~\ref{eqn:pairwise_alignment_terms}) can be written in this way, with the same letter kernel.

    Applying Theorem~\ref{thm: hamming delta}, the position-wise kernel has discrete masses if (and only if) $k'_s$ is strictly positive definite.\footnote{The ``only if'' direction is straightforward: if $k'_s$ is not strictly positive definite, Theorem~\ref{prop: matrix det delta} is violated.}
    This, in turn, depends on the relationship between the substitution and insertion parameters, since both go into $k'_s$.
    Let $V\in \mathbb R^{\B\cup\{\verb|-|\}}$ such that $V_{\verb|-|} = 1$. Call $\tilde V=(V_b)_{b\in\B}$ the components of $V$ in $\B$. Define $\tilde U=K^{-1}\mathbf{1}_{\B}$ so $\tilde UK\tilde U=\sigma$. Now,
    $$k'_s(V, V)=\tilde V^TK\tilde V+2e^{-\mu}\tilde V^T \mathbf{1}_{\B}+1=\tilde V^TK\tilde V+2e^{-\mu}\tilde V^TK\tilde U+1.$$
    Now decompose $\tilde V$ as $\tilde V=\alpha \tilde U+\tilde V^\perp$ where the columns of $\tilde V^\perp$ are orthogonal to those of $\tilde U$ under the inner product $(v|w)=v^TKw$.
    Then, 
    $$k'_s(V, V)=\tilde V^{\perp T}K\tilde V^\perp + \alpha^2\tilde UK\tilde U +2\alpha e^{-\mu}\tilde UK\tilde U+1 =\tilde V^{\perp T}K\tilde V^\perp+\sigma\left( \alpha^2+2e^{-\mu}\alpha\right)+1.$$
    This expression is minimized at $\alpha=-e^{-\mu}, \tilde V^\perp=0$, in which case
    $$k'_s(V, V)=-\sigma e^{-2\mu} + 1.$$
    Thus $k'_s$ is strictly positive definite if and only if $e^{-2\mu} \sigma< 1$, i.e. $2\mu > \log\sigma$.
    Therefore, the alignment kernel has discrete masses if and only if the position-wise comparison kernel has discrete masses.

At bottom, then, the flexibility of the alignment kernel depends on the flexibility of the kernel we use for comparing aligned sequences, and this in turn depends on the flexibility of the kernel we use for comparing letters -- both real letters (nucleotides/amino acids) and gaps.
The interplay between substitution and insertion parameters stems from the fact that we have in essence added the gap symbol to the alphabet, and need the letter kernel over this extended alphabet to still be strictly positive-definite.

\section{Features of the alignment kernel}\label{sec: features of the ali kern}

In this section, we derive a representation of the alignment kernel in terms of features. These features are gapped kmer counts.

\begin{theorem}(proof of Theorem \ref{thm: string basis})
Consider an alignment kernel $k$ with  $k_s(b, b')=\sigma^{-1}|\B|\delta_{b'}(b)$ for $b, b'\in\B$.
Tilting the kernel by $A(X) = e^{\mu|X|}$ gives $k^A(X, Y) = A(X) A(Y) k(X, Y)$.
For all sequences $X \in S$, define the features $\{u_V(X)\}_{V \in S}$, given by
\begin{equation}
u_V(X)=e^{\frac{1}{2} \zeta |V|}\sum_{g=0}^\infty e^{-\Delta\mu g}\sum_{J\in G(|X|, g)}\mathbbm{1}(X_{(J)}=V),
\end{equation}
where $\zeta = 2\mu -\log \sigma+\log|\B|$.
Now we show that $\{u_V(X)\}_{V \in S}$ is an orthonormal basis for $\Hc_{k^A}$.
\end{theorem}
\begin{proof}
	\sloppy
    We can represent alignments between two sequences $X \in S$ and $Y \in S$ by pairs of indices $J\in G(|X|, M),J'\in G(|Y|, M)$ that describe which letters in $X$ and $Y$ are aligned. In particular,
	$k(X, Y)$ is equal to
	\begin{equation*}
	\begin{aligned}
	\sum_{M=0}^\infty \sum_{(j_l)_{l=-1}^{M}\in G(|X|, M)}\sum_{(j'_l)_{l=-1}^{M}\in G(|Y|, M)}
	&k_i(X_{(:j_0)}, Y_{(:j'_0)})\\
    &\times\prod_{m=0}^{M-1}k_s(X_{(j_m)}, Y_{(j'_m)})k_i(X_{(j_{m}+1:j_{m+1})}, Y_{(j'_{m}+1:j'_{m+1})})
	\end{aligned}
	\end{equation*}
	By our assumption on $k_s$, for each alignment $J\in G(|X|, M, g), J'\in G(|Y|, M, g')$, the contribution from the aligned letters, $\prod_{m=0}^{M-1}k_s(X_{(j_m)}, Y_{(j'_m)})$, is $0$ if $X_{(J)}\neq Y_{(J')}$ and is $\sigma^{-M}|\B|^M$ otherwise.
	On the other hand, the contribution from the insertions and deletions is $\prod_{m=0}^{M-1}k_I(X_{(j_{m}:j_{m+1})}, Y_{(j'_{m}+1:j'_{m+1})})= e^{-\mu(|X|-M+|Y|-M)}e^{-\Delta\mu (g+g')}$.
	Thus, $k^A(X, Y)$ is equal to
	\begin{equation}\label{eq: string basis dot}
	\begin{aligned}
	\sum_{M=0}^\infty&\left( e^{2\mu}\sigma^{-1}|\B|\right)^M\sum_{g, g'}e^{-\Delta\mu (g+g')}\sum_{J\in G(|X|, M, g)}\sum_{J'\in  G(|Y|, M, g')}\mathbbm{1}(X_{(J)}=Y_{(J')})\\
	=&\sum_{M=0}^\infty e^{\zeta M} \sum_{V\in \B^M} \sum_{g, g'}e^{-\Delta\mu (g+g')}\sum_{J\in G(|X|, M, g)}\sum_{J'\in G(|Y|, M, g')}\mathbbm{1}(X_{(J)}=V)\mathbbm{1}(Y_{(J')}=V)\\
	=&\sum_{M=0}^\infty \sum_{V\in \B^M} u_V(X) u_V(Y)
	\end{aligned}
	\end{equation}
    Thus the features of $k^A$ are $X\mapsto (u_V(X))_{V\in S}$.
	
    Now we will show that $u_V\in \Hc_k$ for all $V\in S$ and that $(u_V)_V$ make up an orthonormal basis.
	Define the infinite matrix $Q$ as $Q_{V,Z} = e^{-\frac{1}{2} \zeta |V|}u_V(Z)$ for all $V, Z \in S$.
	Pick some total ordering, $\leq$, of $S$ such that $V\leq Z$ if $|V|\leq |Z|$.
	In this case, $Q$ is an infinite upper triangular matrix with $1$'s along the diagonal.
    Thus $Q$ has an infinite upper triangular matrix $Q^{-1}$ as its inverse with $1$s along its diagonal.
    In the remainder of the proof, we will represent any $\sum_Z \alpha_Z k_Z^A\in\text{span}\{k^A_Z\}_{Z\in S}$ as an infinite vector, indexed by elements of $S$, with $\alpha_Z$ in position $Z$. 
    So, for any function $f \in \text{span}\{k^A_Z\}_{Z\in S}$, we can think of the matrix-vector product $Q f$ as the featurization of $f$.
    Now, for any $f, g\in\text{span}\{k^A_Z\}_{Z\in S}$, we have from Equation~\ref{eq: string basis dot},
	$$(f|g)_{k^A}=f^T Q^T \text{diag}(e^{\zeta |V|})_{V\in S} Q g,$$
 where $\text{diag}(e^{\zeta |V|})_{V\in S}$ denotes the infinite diagonal matrix with entries $e^{\zeta |V|}$.
 
	We will next show that $u_X \in \Hc_{k^A}$, with the representation $f_X = e^{- \frac{1}{2} \zeta |X|}\sum_{Z}Q^{-1}_{Z, X}k_Z \in \text{span}\{k^A_Z\}_{Z\in S}$ (note the sum has only finitely many non-zero terms as $Q^{-1}_{Z, X}\neq 0$ for only finitely many $Z$ with $Z\leq X$). We have, 
 $$\text{diag}(e^{\zeta |V|})_{V\in S}Q f_X=e^{\frac{1}{2} \zeta |X|} d_X,$$
    where $d_X$ is the infinite-dimensional vector indexed by $S$ with $0$'s in every position except $1$ in position $X$.
	For any $Y\in S$,
	$(f_X|k^A_Y)_{k^A}=e^{\frac{1}{2} \zeta |X|} Q_{X, Y}=u_X(Y)$.
	Thus, $u_X\in\Hc_{k^A}$.
    We also have,
    $$(u_X|u_Y)_{k^A}=f_X^T Q^T \text{diag}(e^{\zeta |V|})_{V\in S} Q f_Y=d_X^T d_Y=d_X(Y).$$
	Thus $\{u_X\}_{X\in S}$ is a set of orthogonal non-zero vectors, normalized such that $\|u_X\|_{k^A} = 1$ for all $X \in S$.
	Finally, by Equation \ref{eq: string basis dot},
	$$k^A(X, Y)=\sum_{V\in S}^\infty (u_V|k^A_X)_{k^A}(u_V|k^A_Y)_{k^A}\,$$
	for all $X, Y\in S$ so that $\{u_X\}_{X\in S}$ is also a basis.
\end{proof}

We now consider the general case, allowing $k_s$ to be any strictly positive definite kernel on letters, instead of requiring that it is diagonal.
The first step is to show that any alignment kernel can be written as a reparameterization of an alignment kernel with diagonal $k_s$.
\begin{proposition}\label{prop:reparam into diag k_s}
Consider an alignment kernel $k$, with parameters $\mu$, $\Delta \mu$ and $k_s$.
    Let $K_{b, b'}=k_s(b, b')$ for $b, b'\in\B$ and $\sigma = \mathbf{1}_\B^TK^{-1}\mathbf{1}_\B$.
    There is an alphabet $\tilde \B$, inducing the set of sequences $\tilde S$, such that
    $k$ applied to $\tilde S$ is an alignment kernel with parameters $\mu, \Delta\mu$ and $k_s(\tilde b, \tilde b')=\sigma^{-1}|\B|\delta_{\tilde b}(\tilde b')$ for all $\tilde b, \tilde b' \in \tilde \B$.
    As well, for any $b\in \B$, $\sum_{\tilde b\in\tilde \B} k_s(\tilde b, b)=\sigma^{-1}|\B|$.
    We call $\tilde \B$ a rectified alphabet for $k$ and $\tilde S$ a rectified set of sequences for $k$.
\end{proposition}
\begin{proof}
    Let $B_1 = K^{-1}\mathbf{1}_\B/\sqrt{\sigma}$ so $k_s(B_1, B_1)=1$ and let $\{B_{2}, \dots, B_{|\B|}\}$ be an orthonormal set orthogonal to $B_1$ in $\mathbb R^\B$ with inner product $k_s$.
    Let $\Phi\in\mathbb R^{|\B|\times|\B|}$ be a unitary matrix with $|\B|^{-1/2}$ in every entry of its first column and define the letters $C_i=\sqrt{\sigma^{-1}|\B|}\sum_{j=1}^{|\B|}\Phi_{i, j}B_j$.
    Thus we have that $k_s(C_i, C_j)=\sigma^{-1}|\B|(\Phi_{i, \cdot}|\Phi_{j, \cdot})=\sigma^{-1}|\B|\delta_{i}(j)$ and $\sum_{b\in\B}C_{i, b}=C_i^T\mathbf 1_\B=\sqrt{\sigma} C_i^T KB_1=\sqrt{|\B|} \sum_{j=1}^{|\B|}\Phi_{i, j}k_s(B_i, B_1)=\sqrt{|\B|}\Phi_{i, 1}=1$.

    Call $\tilde \B=\{C_i\}_i$ and $\tilde S$ the set of sequences with letters in $\tilde \B$.
    By Proposition \ref{prop: reparam and conv}, for $V, V'\in \tilde S$, $k(V, V')$ is given by
	$$\sum_{l\in \mathbb{N}, V^{(1)}+ \dots+ V^{(2l+1)}=V, V^{'(1)}+ \dots+ V^{'(2l+1)}=V'} \!\!\!\!\!\!\!\!\!\!k_I(V^{(1)}, V^{'(1)})\prod_{i=1}^l k_s(V^{(2i)}, V^{'(2i)})k_I(V^{(2i+1)}, V^{'(2i+1)}).$$
    For $\tilde b, \tilde b'\in\tilde \B$, $k_s(\tilde b, \tilde b')=\sigma^{-1}|\B|\delta_{\tilde b}(\tilde b')$.
    As well, for any number $L$ and $V\in\tilde\B^L$, the insertion kernel $k_I$ applied to $V$ is,
	$$k_{I, V}=\sum_{|X|=L}\left( \prod_{l=0}^{L-1}V_{l, X_{(l)}}\right)e^{-\Delta\mu-L\mu}g =e^{-\Delta\mu-L\mu} \prod_{l=0}^{L-1}\left(\sum_{b\in\B}V_{l, b}\right)g=e^{-\Delta\mu-L\mu} g.$$
    Thus, for $V, V'\in\tilde S$ with $|V|,|V'|>0$,
    $$k_I(V, V')=e^{-\Delta\mu-|V|\mu}e^{-\Delta\mu-|V'|\mu}.$$
    Thus $k$ on $\tilde S$ is the alignment kernel with the same parameters $\mu, \Delta\mu$ and with $k_s(\tilde b, \tilde b')=\sigma^{-1}|\B|\delta_{\tilde b}(\tilde b')$ for $\tilde b,\tilde b' \in \tilde \B$.

    Finally, note, for any $b\in\B$,
    \begin{equation*}
	\begin{aligned}
    \sum_ik_s(C_i, b)=&\sqrt{\sigma^{-1}|\B|}\sum_{i,j}\Phi_{i, j}k_s(B_j, b)\\
    =&\sqrt{\sigma^{-1}|\B|}\sum_{j}\left(\mathbf{1}_{|\B|}^T\Phi_{\cdot, j}\right)k_s(B_j, b)\\
    =&\sqrt{\sigma^{-1}|\B|}\sum_{j}\sqrt{|\B|}\delta_1(j)k_s(B_j, b)\\
    =&\sqrt{\sigma^{-1}}|\B|k_s(B_1, b)\\
    =&\sigma^{-1}|\B|.
	\end{aligned}
	\end{equation*}
\end{proof}
We can now describe the features of an alignment kernel with arbitrary $k_s$.
When $k_s$ was diagonal, each feature of a sequence $X$ depended on the number of positions at which a gapped kmer matched $X$ exactly.
Now, instead of demanding an exact match, we score each match according to $k_s$.
\begin{corollary}\label{cor:general basis}
Consider an alignment kernel $k$ and let $\tilde S$ be a rectified set of sequences.
For $Y\in S$, $V\in\tilde S$ with $L=|Y|=|V|$, define the pairwise comparison using $k_s$ as,
$$c_{pw}(Y, V)=\prod_{l=0}^{L-1}\sigma |\B|^{-1}k_s(Y_{(l)}, V_{(l)}).$$
Tilting the kernel by $A(X) = e^{\mu|X|}$ gives $k^A(X, Y) = A(X) A(Y) k(X, Y)$.
For all sequences $X \in S$, define the features $\{u_V(X)\}_{V \in \tilde S}$, given by
\begin{equation}
u_V(X)=e^{\frac{1}{2} \zeta |V|}\sum_{g=0}^\infty e^{-\Delta\mu g}\sum_{J\in G(|X|, g)}c_{pw}(X_{(J)},V),
\end{equation}
where $\zeta = 2\mu -\log \sigma+\log|\B|$.
Now, $\{u_V(X)\}_{V \in \tilde S}$ is an orthonormal basis for $\Hc_{k^A}$.
\end{corollary}
\begin{proof}
    Let $\tilde\B$, $\tilde S$ a rectified alphabet and a rectified set of sequences for $k$.
    By Theorem~\ref{thm: string basis}, there is an orthonormal basis of $\Hc_{k^A}$, $\{u_V\}_{V\in\tilde S}$ such that for any $V, X\in\tilde S$
    $$(u_V|k_{X}^A)_{k^A}=e^{\frac 1 2 \zeta |V|}\sum_{g=0}^\infty e^{-\Delta\mu g}\sum_{J\in G(|X|, g)}\mathbbm{1}(X_{(J)}=V).$$
    Note that reparameterization does not affect the RKHS $\Hc_{k^A}$, so that $\{u_V\}_{V\in\tilde S}$ is an orthonormal basis for the kernel $k^A$ applied to the space $S$ as well.
    
    Now, consider a $Y \in S$, with $L = |Y|$. Since $\tilde \B$ is a basis for $\mathbb{R}^{\B}$, we can write each letter $Y_{(l)}$ as $Y_{(l)}=\sum_{\tilde b\in\tilde\B}Y_{(l), \tilde b}\, \tilde b$ for some scalar coefficients $\{Y_{(l), \tilde b}\}_{\tilde b \in \tilde \B}$ (to be clear, we are treating $Y_{(l)}$ here as a one-hot encoding of the $l$th letter of $Y$). 
    To derive the coefficients $Y_{(l), \tilde b}$, recall that $\{\sqrt{\sigma |\B|^{-1}}\, \tilde b\}_{\tilde b \in \tilde \B}$ is an orthonormal basis of $\mathbb{R}^\B$, with inner product defined by $k_s$.
    Thus, we have $Y_{(l), \tilde b} = k_s(Y_{(l)}, \sqrt{\sigma|\B|^{-1}}\,\tilde b) \sqrt{\sigma|\B|^{-1}} =\sigma |\B|^{-1}k_s(Y_{(l)}, \tilde b)$ for all $\tilde b \in\tilde \B$.
    We also have, applying the properties of the rectified alphabet in Proposition~\ref{prop:reparam into diag k_s}, $\sum_{\tilde b \in\tilde \B} Y_{(l),\tilde b} = \sigma |\B|^{-1} \sum_{\tilde b \in\tilde \B} k_s(Y_{(l)},\tilde b) = 1$.
    Thus,
	\begin{equation*}
	\begin{aligned}
    (u_V|k_{Y}^A)_k^A=&\sum_{X\in\tilde S\ :\ |X|=L}\left(\prod_{l=0}^{L-1}Y_{(l), X_{(l)}}\right)e^{\frac 1 2 \zeta |V|}\sum_{g=0}^\infty e^{-\Delta\mu g}\sum_{J\in G(|X|, g)}\mathbbm{1}(X_{(J)}=V)\\
    =&e^{\frac 1 2 \zeta |V|}\sum_{g=0}^\infty e^{-\Delta\mu g}\sum_{J\in G(|X|, g)}\sum_{X\in\tilde S\ :\ |X|=L}\left(\prod_{l=0}^{L-1}Y_{(l), X_{(l)}}\right)\prod_{m=0}^{M-1}\mathbbm{1}(X_{(j_m)}=V_{(m)})\\
    =&e^{\frac 1 2 \zeta |V|}\sum_{g=0}^\infty e^{-\Delta\mu g}\sum_{J\in G(|X|, g)}\left(\prod_{l\not\in J}\sum_{b\in\tilde \B}Y_{(l), b}\right)\prod_{m=0}^{M-1}Y_{(j_m),V_{(m)}}\\
    =&e^{\frac 1 2 \zeta |V|}\sum_{g=0}^\infty e^{-\Delta\mu g}\sum_{J\in G(|X|, g)}1\times\prod_{m=0}^{M-1}\sigma |\B|^{-1}k_s(Y_{(j_m)}, V_{(m)})\\
    =&e^{\frac 1 2 \zeta |V|}\sum_{g=0}^\infty e^{-\Delta\mu g}\sum_{J\in G(|X|, g)}c_{pw}(Y_{(J)}, V)\\
	\end{aligned}
	\end{equation*}
\end{proof}
\section{Alignment kernels are universal and characteristic for certain tiltings}\label{sec: ali kernl univ}

In this section we show that all alignment kernels are, under an appropriate tilting, universal and characteristic---even if they do not have discrete masses.
Note that while a kernel with discrete masses maintains discrete masses after any tilting, the same is not true for universal or characteristic kernels; here, the tilting matters.

Functions in $C_0$ vanish at infinity, so to be sure the alignment kernel can approximate any $C_0$ function, even when the kernel does not have discrete masses, the tilting will have to be sufficiently small.
The normalizing tilting $A(X) = \sqrt{k(X, X)}^{-1}$ is not necessarily small enough.
Instead, one would in general need to tilt so much that $k^A(X, X)\to 0$ as $|X|\to\infty$.
From a modeling perspective, this is problematic: if $f$ is drawn from a Gaussian process with kernel $k^A$, then its variance will go to zero as the length of sequences increases: $\mathrm{Var}(f(X))=k^A(X, X)\to 0$ as $|X|\to\infty$. Thus, on longer sequences, the value of $f$ will be determined essentially just by the prior. 
So, although we can force alignment kernels that lack discrete masses to be universal if we tilt them enough, the tilted kernels will generalize poorly.

\begin{proposition}\label{prop: ali universal}
    Say $k$ is an alignment kernel and let $\tilde k$ be an alignment kernel with the same values of $\mu, \Delta\mu$ as $k$ but with $\tilde k_s(b, b')=\sigma e^{-2\mu+\epsilon}k_s(b, b')$ for some $\epsilon>0$.
    Choose an $A:S\to (0, \infty)$ such that $\tilde k^A$ is a $C_0$-kernel, meaning $Y\mapsto A(Y)A(X)\tilde k(X, Y)$ is in $C_0(S)$ for every $X$ and $\sup_{X\in S}A(X)A(X)\tilde k(X, X)<\infty$.
    Then the tilted alignment kernel $k^A$ is $C_0$-universal and characteristic.
\end{proposition}
\begin{proof}
    Assume throughout that $2\mu\leq \log\sigma$, otherwise the result follows from Theorem \ref{thm: delta funcs string} and Corollary \ref{cor: tilting delta}.
    Note $\tilde k(X, Y)\geq k(X, Y)$ for all $X, Y\in S$.
    Thus, $k^A$ is a $C_0$-kernel.
    Also note that, defining $\tilde K_{b, b'}=\tilde k_s(b, b')$, we have $\log(\textbf 1_{\B}\tilde K^{-1}\textbf 1_{\B})=2\mu-\epsilon < 2 \mu$, so $\tilde k$ has discrete masses by Theorem \ref{thm: delta funcs string}. 
    
    We now derive an orthogonal basis for $\Hc_{\tilde k^A}$. Let $\bar A(X)=\exp(-\mu |X|)$ and
    let $\tilde S$ be a rectified set of sequences for $k^A$. Now, by Corollary \ref{cor:general basis} and Proposition~\ref{prop: tilt delta}, $\{A \bar A u_V\}_{V\in \tilde S}$ is an orthonormal basis of $\Hc_{ k^A}$.
    Note $\tilde S$ is also a rectified set of sequences for $\tilde k^A$, so let
    $\{A \bar A \tilde u_V\}_{V\in \tilde S}$ be the orthonormal basis of $\Hc_{\tilde k^A}$ from Corollary \ref{cor:general basis}.
    Note $c_{pw}$ from Corollary \ref{cor:general basis} is identical for $k^A$ and $\tilde k^A$ so that each $\tilde u_V$ is a scaled version of $u_V$;
    in particular, $\{A \bar A u_V\}_{V\in \tilde S}$ is an orthogonal basis for $\Hc_{\tilde k^A}$.

    Since $\tilde k$ has discrete masses, it is universal, so  for any $f\in C_0(S)$ and $\epsilon>0$ we can pick a $g\in  \Hc_{\tilde k}$ such that $\Vert f-g\Vert_\infty<\epsilon$.
    Define $M=\sup_{X\in S}\sqrt{A(X)A(X)\tilde k(X, X)}<\infty$ and choose a $g'\in\mathrm{span}\{\tilde k_X^A\}_X$ such that $\Vert g'-g\Vert_{\tilde k^A}\leq \epsilon/M$. Now, $g'\in C_0(S)$ and
    $$\vert g'(X)-g(X)\vert=\vert(\tilde k_X^A|g'-g)_{\tilde k^A} \vert\leq \Vert \tilde k^A_X\Vert_{\tilde k^A}\Vert g'-g\Vert_{\tilde k^A}\leq M\epsilon/M=\epsilon$$
    for all $X\in S$.
    Thus, $\Vert g' - f\Vert_\infty\leq 2\epsilon$.
    Thus, every function in $\Hc_{\tilde k^A}$ can be approximated in $C_0$ by a function in $\mathrm{span}\{\tilde k_X^A\}_X$.
    Thus, $\mathrm{span}\{\tilde k_X^A\}_X$ is dense in $C_0(S)$.

    Notice the finite dimensional vector spaces $\mathrm{span}\{\tilde k_X^A\}_{X \in S \ :\ |X|\leq L}$ and $\mathrm{span}\{ A \bar A u_X\}_{X\in\tilde S\ :\ |X|\leq L}$ are identical for all $L$ (recall from the proof of Proposition~\ref{prop: reparam delta} that reparameterizing the alphabet preserves the span of the kernel embedding vectors). As a result, $\mathrm{span}\{\tilde k_X^A\}_{X\in S}=\mathrm{span}\{ A\bar A u_X\}_{X\in \tilde S}$.
    But this latter vector space is similarly equal to $\mathrm{span}\{k^A_X\}_X$.
    Thus, $\mathrm{span}\{k^A_X\}_X=\mathrm{span}\{\tilde k^A_X\}_X$, so $\mathrm{span}\{k^A_X\}_X$ is dense in $C_0(S)$.
    $k^A$ is thus universal. Further, by proposition 2 of \citet{Sriperumbudur2011-ay}, $k^A$ is also characteristic.
\end{proof}

To interpret this result, and the tiltings it requires, note that we must choose a tilting $A$ such that $\tilde k^A(X, X)$ is bounded. 
Roughly speaking, we expect $\frac{\tilde k(X, X)}{k(X, X)}$ to grow exponentially with the length $|X|$, since each match in each pairwise alignment has been up-weighted by a factor of $\sigma e^{-2\mu+\epsilon} > 1$. 
The normalizing tilting $A(X) = \sqrt{k(X, X)}^{-1}$ would help us bound $k^A(X, X)$, but to bound $\tilde k^A(X, X)$ we need to go further and exponentially down-weight longer sequences.
This means choosing a tilting $A$ such that $k^A(X, X) \to 0$ as $|X| \to 0$.

\section{Discrete masses in the one-letter alignment kernel}\label{sec: proof of ali discrete masses |B|=1}

    In this section we show that alignment kernels have discrete masses under certain parameter settings \textit{when the alphabet size is one}. 
    This result is crucial for proving the more general result, covering arbitrary alphabet sizes, in Theorem \ref{thm: delta funcs string} and Appendix \ref{sec: proof of ali discrete masses}. It builds on the feature representation described in Theorem \ref{thm: string basis} and Appendix \ref{sec: features of the ali kern}.

    The result will depend on the theory of the Riordan group, for which we will need to introduce some notation.
    If $f(x)=\sum_{i=0}^\infty C_i x^i$ is a formal power series, define $[x^n]f=C_n$.
    For two formal power series $f(y), g(y)$ we define the Riordan array of $f$ and $g$ as the infinite upper triangular matrix with entries
    $$Q_{L', L}=[x^{L'}y^L]\frac{g(y)}{1-xyf(y)}=[x^{L'}y^L]\sum_{l=0}^\infty g(y)x^l y^l f(y)^l=[y^L]y^{L'}g(y)f(y)^{L'}$$
    for numbers $L, L'\geq 0$.
    We say that $Q$ is in the Bell subgroup if $f=g$.
    The result we will need is that one can invert a Riordan array by inverting the power series $f$ and $g$ \citep{Shapiro1991-ak}.
    \begin{proposition}\label{prop: riordan inverse}
        \citep{Shapiro1991-ak} Let 
        $$Q_{L', L}=[x^{L'}y^L]\frac{g(y)}{1-xyf(y)}$$
        and let $\bar f$ be the unique power series such that the equation
        $y\bar f(y)f(y\bar f(y))=y$ holds.\footnote{
        Another way to write this condition is that $tf(t)=y$ when $t=y\bar f(y)$ so that $yf$ is the compositional inverse of $y\bar f$.}
        Then, letting
        $\bar g(y)=\left(g(y\bar f(y))\right)^{-1},$
        $$Q^{-1}_{L', L}=[x^{L'}y^L]\frac{\bar g(y)}{1-xy\bar f(y)}.$$
        
        Note if $Q$ is in the Bell subgroup, $g(y)=f(y)$ and $\bar f(y)f(y\bar f(y))=1$ so $\bar g(y)=\bar f(y)$. Thus $Q^{-1}$ is also in the Bell subgroup.
    \end{proposition}
    \begin{proof}
        \begin{equation*}
        \begin{aligned}
            \sum_{l=0}^\infty Q_{L, l}Q^{-1}_{l, L'}=&\sum_{l=0}^\infty [y^l]g(y)\left(y f(y)\right)^L[z^{L'}]\bar g(z)\left(z \bar f(z)\right)^l\\
            =&[z^{L'}]\bar g(z)\sum_{l=0}^\infty\left(z \bar f(z)\right)^l [y^l]g(y)\left(y f(y)\right)^L\\
            =&[z^{L'}]\bar g(z) g(z \bar f(z))\left(z\bar f(z) f(z\bar f(z))\right)^L\\
            =&[z^{L'}]1\times z^L = \delta_{L'}(L).
        \end{aligned}
        \end{equation*}
    \end{proof}

\begin{theorem}\label{thm: delta funcs string B=1}
	Say $|\B|=1$ and call $A$ the only letter in $\B$.
	If $\Delta\mu>0$, $k$ has discrete masses if and only if $2\mu + \log k_s(A, A) \geq 0$.
	If $\Delta\mu=0$, $k$ has discrete masses if and only if $2\mu + \log k_s(A, A) > 0$.
\end{theorem}

\begin{proof}
	\sloppy 
    Note in this case $\zeta=2\mu + \log k_s(A, A)$.
	We will replace $k$ with $k^A(X, Y)=e^{\mu |X|}e^{\mu |Y|}k(X, Y)$ throughout.
	By Proposition \ref{prop: tilt delta} this does not affect whether or not the kernel has discrete masses.
	Index $S=\{0, 1, \dots\}$.
	Call $Q$ the infinite upper triangular matrix with $e^{-\frac 1 2\zeta L'}u_{L'}(L)=\sum_{g=0}^\infty e^{-g\Delta\mu}|G(L, L', g)|$ in position $(L', L)$.
	Note that the Gram matrix of $k$ on $\{0, 1, \dots L\}$ is $Q_{:L+1, :L+1}^T\mathrm{diag}(e^{\zeta L'})_{L'=0}^L Q_{:L+1, :L+1}$.
	Since $Q$ is infinite upper triangular with no zeros along its diagonal it has an infinite upper triangular inverse $Q^{-1}$ with the property that $(Q_{:L, :L})^{-1}=(Q^{-1})_{:L, :L}$.
	Thus the inverse of the Gram  matrix of $k$ on $\{0, 1, \dots L\}$ is $Q_{:L+1, :L+1}^{-1 T}\mathrm{diag}(e^{-\zeta L'})_{L'=0}^L Q^{-1}_{:L+1, :L+1}$ which has, in position $(L', L')$,
	$\sum_{l=L'}^L e^{-\zeta l}(Q_{L', l}^{-1})^2$.
	Thus, by Proposition \ref{prop: matrix det delta}, $\delta_{L'}\in\Hc_k$ if and only if 
	\begin{equation}\label{eq: string delta cond for M}
	\sum_{l=L'}^\infty e^{-\zeta l}(Q^{-1}_{L', l})^2<\infty.
	\end{equation}
	The proof will now proceed in two parts using the technique of generating functions and the theory of the Riordan group.
	In part 1 we will equate $u_L(L')$ with the coefficients of a formal power series.
	We will then explicitly invert the matrix $Q$ by inverting this power series.
	In part 2 we will equate Equation \ref{eq: string delta cond for M} with the convergence of a power series, which we will use to prove the result.
	
	Part 1:
	Say $g$ is an integer greater than or equal to $1$, and $L\geq L'$.
	We want to find the size of $G(L, L', g)$.
	Each $J=(j_{-1}, \dots, j_{L'})\in G(L, L', g)$ is uniquely characterized by the numbers $j_{m_1}, \dots, j_{m_g}$, which indicate the positions at which each of the $g$ gaps start, and $j_{m_1+1}, \dots, j_{m_g+1}$, which indicate the position at which each of the $g$ gaps end. In particular, 
	$$J=(-1, 0, 1, \dots, j_{m_1}-1, j_{m_1}, j_{m_1+1}, j_{m_1+1}+1, \dots, j_{m_2}-1, j_{m_2},j_{m_2+1}, \dots, L).$$
	Thus each $J$ is also uniquely characterized by the sizes of its $g$ gaps $l_1=j_{m_1+1}-j_{m_1}-1, l_2=j_{m_2+1} -j_{m_2}-1, \dots$ as well as the sizes of its $g+1$ contiguous regions
	$c_0=j_{m_1}+1, c_1=j_{m_2}-j_{m_1+1}+1, c_{2}=j_{m_3}-j_{m_2+1}+1,\dots, c_g=L-j_{m_g+1}$.
	On the other hand, given numbers $l_1, \dots, l_g, c_0, \dots, c_{g}$, 
	one gets a unique valid $J\in G(L, L', g)$ if and only if $\sum_{i=0}^gc_i=L', \sum_{i=1}^gl_i=L-L'$, $l_i\geq 1$ for all $i$, and $c_i\geq 1$ for all $i=1, 2, \dots, g-1$ ($c_0=0$ implies a gap at the beginning of the sequence while $c_g=0$ implies a gap at the end).
	The number of $g+1$ ordered numbers $(c_0, \dots, c_g)$ with $\sum_{i=0}^gc_i=L'$ and $c_i\geq 1$ for all $i=1, 2, \dots, g-1$ is the $L'$-th coefficient of the power series
	$$(1+x+x^2+\dots)\left(x+x^2+\dots\right)^{g-1}(1+x+x^2+\dots)=\frac{x^{g-1}}{(1-x)^{g+1}}.$$
	Similarly, the number of sets of number of $g$ ordered numbers $(l_1, \dots, l_g)$ with $\sum_{i=1}^gl_i=L-L'$ and $l_i\geq 1$ for all $i$ is the $L-L'$-th coefficient of the power series $\frac {y^g}{(1-y)^g}$.
	On the other hand if $g=0$, $G(L, L', g)$ is empty if $L\neq L'$ otherwise it has one element.
	In particular, $G(L, L', 0)$ is (trivially) the coefficient in front of $x^{L'}y^{L-L'}$ of $1/(1-x)$.
	Thus, the size of $G(L, L', g)$ is the coefficient in front of $x^{L'}y^{L-L'}z^g$ of the power series 
	\begin{equation*}
	\begin{aligned}
	\sum_{g=1}^\infty z^g \frac{x^{g-1}}{(1-x)^{g+1}}\frac{y^{g}}{(1-y)^{g}} + \frac 1 {1-x} =&\frac{zy}{(1-x)^2(1-y)}\sum_{g=0}^\infty \left(z \frac{x}{1-x}\frac{y}{1-y}\right)^g + \frac 1 {1-x}\\
 =&\frac{zy}{(1-x)^2(1-y)}\frac{(1-x)(1-y)}{(1-x)(1-y)-zxy}+\frac 1 {1-x}\\
	=&\frac 1{1-x}\left(\frac{zy}{(1-x)(1-y)-zxy}+1\right)\\
	=&\frac 1{1-x}\left(\frac{zy+1-x-y+xy-zxy}{(1-x)(1-y)-zxy}\right)\\
	=&\frac 1{1-x}\left(\frac{(zy+1-y)(1-x)}{(1-x)(1-y)-zxy}\right)\\
	=&\frac{1-(1-z)y}{(1-x)(1-y)-zxy}\\
	=&\frac{\frac{1-(1-z)y}{1-y}}{1-x\frac{1-(1-z)y}{1-y}}.
	\end{aligned}
	\end{equation*}
	Call $f_z(y)=\frac{1-(1-z)y}{1-y}$.
	Next we replace $x$ with $xy$ so that we get that $\vert G(L, L', g)\vert$ is $[(xy)^{L'}y^{L-L'}z^g]\frac{f_z(y)}{1-(xy)f_z(y)}=[x^{L'}y^{L}z^g]\frac{f_z(y)}{1-xyf_z(y)}$.
	Thus, substituting $z=e^{-\Delta\mu}$ and calling $f=f_{e^{-\Delta\mu}}$, 
    $$Q_{L', L}=\sum_{g=0}^\infty e^{-g\Delta\mu}|G(L, L', g)|=[x^{L'}y^{L}]\sum_{g=0}^\infty e^{-g\Delta\mu}[z^g]\frac{f_z(y)}{1-xyf_z(y)}=[x^{L'}y^{L}]\frac{f(y)}{1-xyf(y)}.$$
	
	$Q$ is in the Bell subgroup of Riordan arrays (\citet{Shapiro1991-ak} and chapter 3 of \citet{Shapiro2022-lc}).
	This means that by Proposition \ref{prop: riordan inverse}, defining $\bar f$ as the unique power series with
    $y\bar f(y)f(y\bar f(y))=y$,
	$$Q^{-1}_{L', L}=[x^{L'}y^{L}]\frac{\bar f(y)}{1-xy\bar f(y)}.$$
	
	Part 2:
	Since
	$$Q^{-1}_{L', L}= [x^{L'}y^{L}]\frac{\bar f(y)}{1-xy\bar f(y)} = [y^{L}]\bar f(y)(y\bar f(y))^{L'}=[y^{L-L'}]\bar f(y)^{L'+1},$$
 we have,
	$$\sum_{l=L}^\infty e^{-\zeta l}(Q^{-1}_{L, l})^2=e^{-\zeta L}\sum_{l=0}^\infty (e^{-\frac 1 2\zeta l}[y^{l}]\bar f(y)^{L+1})^2.$$
	This sum is finite if $e^{-\frac 1 2\zeta}$ is smaller than the radius of convergence of $\bar f(y)^{L+1}$ (as the terms in the series decay exponentially) and infinite if it is larger (as the terms do not approach $0$).
    We will now show that the radius of convergence of $\bar f$ is exactly $1$, so that if $\zeta < 0$ then $\delta_0\not\in\Hc_k$ and if $\zeta > 0$ then, since the radius of convergence of $\bar f(y)^L$ is also at least $1$ for all $L$, $k$ has discrete masses.
	We will afterwards carefully consider the case of $\zeta = 0$.

	Let us first consider the case $\Delta\mu=0$.
    In this case we have, from the definition of $f$, $yf(y)=\frac y {1-y}$, so $y\bar f(y)=\frac y {1+y}$.
	In this case $\bar f$ clearly has radius of convergence $1$. We can also see explicitly that
	$[y^{l}]\bar f(y)^{L+1}=(-1)^{l}\binom{L+l}{l}$.
	Thus Equation \ref{eq: string delta cond for M} also does not hold for any $L$ when $\zeta=0$.

 	We now assume $\Delta\mu>0$.
	We will derive the form of $\bar f$.
	Call $\xi= 1-e^{-\Delta\mu}\in(0, 1)$, and $t=y\bar f(y)$. 
	\begin{equation*}
	tf(t)=y\\
	\iff t(1-\xi t) =(1-t)y\\
	\iff 0=\xi t^2 - (y+1) t +y.
	\end{equation*}
	Thus, picking the root that is $0$ when $y=0$, we get
	$$y\bar f(y)=\frac 1 {2\xi}\left(1+y-\sqrt{(1+y)^2-4\xi y^2}\right).$$
	The roots of the polynomial $(1+y)^2-4\xi y^2$ are 
	$$(2\xi-1)\pm\sqrt{(2\xi-1)^2-1}= (2\xi-1)\pm i\sqrt{1-(2\xi-1)^2}.$$
	Note both of these roots have norm $1$ and can thus be written $e^{i\theta}, e^{-i\theta}$ for some $\pi>\theta>0$.
	Thus, we may write
	$$y\bar f(y)=\frac 1 {2\xi}\left(1+y-\sqrt{(1-e^{i\theta}y)(1-e^{-i\theta}y)}\right).$$
	$\sqrt{1-e^{i\theta}y}, \sqrt{1-e^{-i\theta}y}$ each have radius of convergence $1$, so, $y\bar f$ has radius of convergence at least $1$.
	On the other hand, the derivative of $y\bar f$ diverges when $y$ approaches $e^{i\theta}$ or $e^{-i\theta}$, so that $y\bar f$, and thus $\bar f$, has radius of convergence exactly $1$.
	
	The case when $\zeta=0$ is more delicate.
	First call $g(y) = \sqrt{(1+y)^2-4\xi y^2}=\sqrt{(1-e^{i\theta}y)(1-e^{-i\theta}y)}$.
	For real $0<r<1$, define
	\begin{equation*}
	\begin{aligned}
	q_L(r)=&\int_0^{1}dt g(\sqrt{r}e^{2\pi it})^Lg(\sqrt{r}e^{-2\pi it})^L\\
    =&\int_0^{1}dt \sum_{n}r^{n/2}e^{2\pi nit}[y^n]g(y)^L\sum_m r^{m/2}e^{-2\pi mit}[z^m]g(z)^L\\
	=&\sum_{n, m} \left([y^n]g(y)^L[z^m]g(z)^L\right)r^{(n+m)/2}\int_0^{1}dte^{2\pi it(n-m)}\\
	=&\sum_n ([y^n]g(y)^L)^2r^n.
	\end{aligned}
	\end{equation*}
    $g$ is analytic with real coefficients at $0$ so $([y^n]g(y)^L)^2\geq 0$ for all $L, n$.
	Thus, by the monotone convergence theorem, $\sum_n ([y^n]g(y)^L)^2=\lim_{r\to 1^-}q_L(r)$.
	However,
	$$g(\sqrt{r}e^{2\pi iz})^Lg(\sqrt{r}e^{-2\pi iz})^L=\left|1-e^{i(\theta+2\pi z)}\sqrt r\right|^L\left|1-e^{i(\theta-2\pi z)}\sqrt r\right|^L\leq4^L,$$
	so $\sum_n ([y^n]g(y)^L)^2<\infty$ for all $L$.
	Now note
	\begin{equation*}
	\begin{aligned}
	\sum_n([y^n](y\bar f(y))^L)^2 =&\sum_n\frac 1 {(2\xi)^{2L}}\left(\sum_{l=0}^L  [y^n](1+y)^l (-1)^{L-l}g(y)^{L-l}\right)^2\\
	=&\sum_n\frac 1 {(2\xi)^{2L}}\left(\sum_{l=0}^L(-1)^{L-l}  \sum_{l'=0}^l\binom{l}{l'}[y^{n-l'}]g(y)^{L-l}\right)^2\\
	\leq&\sum_n\frac 1 {(2\xi)^{2L}}\left(L^2\max_{l, l'}\left\vert\binom{l}{l'}[y^{n-l'}]g(y)^{L-l}\right\vert\right)^2\\
	\leq&\frac 1 {(2\xi)^{2L}}(L^2)^2\sum_n\max_{l, l'}\left\vert\binom{l}{l'}^2\left([y^{n-l'}]g(y)^{L-l}\right)^2\right\vert\\
	\leq&\frac 1 {(2\xi)^{2L}}(L^2)^2\sum_{l=0}^L \sum_{l'=0}^l\binom{l}{l'}^2\sum_n\left([y^{n-l'}]g(y)^{L-l}\right)^2<\infty.
	\end{aligned}
	\end{equation*}
    Thus $k$ has discrete masses when $\zeta=0$ as well.
	
\end{proof}
    
\section{Discrete masses in the local alignment kernel}\label{sec: proof of local ali discrete masses}
In this section we prove that the local alignment kernel has discrete masses under certain parameter settings.

\begin{theorem}
    (proof of Theorem \ref{thm: gen string})
	If $\Delta\mu=0$, $k_{\mathrm{la}}$ has discrete masses if and only if $2\mu>\log\sigma$.
	If $\infty>\Delta\mu>0$, $k_{\mathrm{la}}$ has discrete masses if and only if $2\mu\geq\log\sigma$.
	If $\Delta\mu=\infty$, $k_{\mathrm{la}}$ has discrete masses.
\end{theorem}

\begin{proof}
    We will only provide a sketch of the proof highlighting the differences to the case of the regular sequence kernel.
    The details are very similar to those of the proofs of Theorems \ref{thm: delta funcs string} and \ref{thm: delta funcs string B=1}.
    
	First we repeat the logic of Theorem \ref{thm: delta funcs string} to reduce to the case of $|\B|=1$.
    Let $U, W_V, V[t]$, $\tilde B$ as in Theorem \ref{thm: delta funcs string}.
    First note if $V=\emptyset$, $W_V$ is simply $k_{\mathrm{la}}$ with $k_s(U, U)=\sigma^{-1}$ on a space with $|\B|=1$.
    Thus, if this local alignment kernel does not have discrete masses, neither does $k_{\mathrm{la}}$ on the entire space.
    Next consider $|V|\geq 1$ and define the left-local alignment kernel
    $$k_{\mathrm{l}}(X, Y)=\sum_{l=0}^\infty\tilde k_I\star (k_s\star k_I)^l$$
    and the right-local alignment kernel $k_{\mathrm{r}}$ similarly.
    Now we repeat the logic of Eqn~\ref{eqn: tesor decom string}:
	\begin{equation}
	\begin{aligned}
	k(V[t], V[t'])=&\sum_{{\substack{l\in\mathbb N\\ Y^{(1)}+ \dots+ Y^{(2l+1)}=t_0\times U \\ Z^{(1)}+ \dots+ Z^{( 2l+1)}=t'_0\times U}}}\sum_{{\substack{l'\in\mathbb N \\ \tilde{Y}^{(1)}+ \dots+ \tilde{Y}^{(2l'+1)}=V[t_{1:}] \\ \tilde{Z}^{(1)}+ \dots+ \tilde{Z}^{(2l'+1)}=V[t'_{1:}]}}}\Bigg(\\
	&\times\left(\tilde k_I(Y^{(1)}, Z^{(1)})\prod_{i=1}^l k_s(Y^{(2i)}, Z^{(2i)})k_I(Y{(2i+1)}, Z^{(2i+1)})\right)\\
    &\times k_s(V_{(0)}, V_{(0)})\\
    &\times\left(\prod_{i=1}^{l'} k_I(\tilde Y^{(2i-1)}, \tilde Z^{(2i-1)}k_s(\tilde Y^{(2i)}, \tilde Z^{(2i)}))\right)\times \tilde k_I(\tilde Y^{(1)}, \tilde Z^{(1)})\Bigg)\\
	=& k_{\mathrm{l}}(t_0\times U, t'_0\times U) k_{\mathrm{r}}(V[t_{1:}], V[t'_{1:}])\\
	=&\dots\\
	=& k_{\mathrm{l}}(t_0\times U, t'_0\times U)\left(\prod_{l=1}^{L-1} k_{\mathrm{ali}}(t_l\times U, t'_l\times U)\right)k_{\mathrm{r}}(t_L\times U, t'_L\times U).
	\end{aligned}
	\end{equation}
    Thus, $k_{\mathrm{la}}$ has discrete masses on $W_V$ if $k_{\mathrm{ali}}, k_{\mathrm{l}},$ and $k_{\mathrm{r}}$ with $k_s(U, U)=\sigma^{-1}$ and $|\B|=1$ have discrete masses.
    Thus, the theorem holds if we prove the case where $|\B|=1$ for $k_{\mathrm{ali}}, k_{\mathrm{la}}, k_{\mathrm{l}},$ and $k_{\mathrm{r}}$.

	We therefore study the case where $|\B|=1$.
	We've already shown the theorem for the regular alignment kernel.
    We now consider $k_{\mathrm{l}}$.
    Let $G(L, L', g)$ be the number of length $L'$ sub-strings of a sequence of length $L$ with $g$ gaps \textbf{not including a gap on the left end}.
    In this case, each element of $G(L, L', g)$ is characterized by the length of its $g+1$ contiguous regions, each of which must be at least length $1$ except for the region after the last gap; and $g+1$ gap regions, each of which must be at least length $1$ except the first gap.
    Thus by the same argument as in Theorem~\ref{thm: delta funcs string B=1}, we get that $|G(L, L', g)|$ is the coefficient in front of $x^{L'}y^{L-L'}z^g$ of the power series
	\begin{equation*}
	    \begin{aligned}
	        \sum_{g=0}^\infty z^g \frac{x^g}{(1-x)^{g+1}}\frac{y^g}{(1-y)^{g+1}}=&\frac{1}{(1-x)(1-y)}\frac{(1-x)(1-y)}{(1-x)(1-y)-xyz}\\
	        =&\frac{\frac 1 {1-y}}{1-xf_z(y)}.
	    \end{aligned}
	\end{equation*}
	Thus in this case, defining the matrix $Q$ as in Theorem~\ref{thm: delta funcs string B=1},
	$$Q_{L', L}=[x^{L'}y^L]\frac{\frac 1 {1-y}}{1-xyf(y)}.$$
 We can invert this matrix using Proposition~\ref{prop: riordan inverse}.
	In particular, the numerator $\bar g (y)$ of the power series corresponding to the matrix inverse is
    $$\left(\left(\frac 1 {1-y}\right)\circ(y\bar f(y))\right)^{-1}=1-y\bar f(y).$$
	Thus,
	$$Q_{L', L}^{-1}=[x^{L'}y^L]\frac{{1-y\bar f(y)}}{1-xy\bar f(y)}.$$
    In this case, $k_{\mathrm{l}}$ has discrete masses if and only if for all $L$,
    $$\sum_{l=0}^\infty e^{-\zeta l} (Q_{L, l}^{-1})^2=\sum_{l=0}^\infty\left(e^{-\frac{1}{2}\zeta l}[y^l](1-y\bar f(y))(y\bar f(y))^L\right)^2<\infty.$$
	The proof of the theorem is almost identical in the case that $\Delta\mu<\infty$.
	If $\Delta\mu = \infty$ then $f(y)=1$ so $\bar f(y) = 1$ and $Q_{L', L}^{-1}=\delta_{L'}(L)-\delta_{L'+1}(L)$. Thus, only finitely many terms of the sum are non-zero, and Equation \ref{eq: string delta cond for M} is satisfied.
    The case is identical for $k_{\mathrm{r}}$.

	Finally we look at $k_{\mathrm{la}}$.
    Let $G(L, L', g)$ be the number of length $L'$ sub-strings of a sequence of length $L$ with $g$ gaps \textbf{not including gaps on either end}.
    In this case, when $g>0$, each element of $G(L, L', g)$ is characterized by the length of its $g+1$ contiguous regions, each of which must be at least length $1$; and $g+2$ gap regions, each of which must be at least length $1$ except the first and last gap.
    When $g=0$ we must be sure to include the case where $L'=0$, even though $L'=0$ is impossible when $g > 0$.
    Thus, $|G(L, L', g)|$ is the coefficient in front of $x^{L'}y^{L-L'}z^g$ of the power series
	\begin{equation*}
	    \begin{aligned}
	        \sum_{g=0}^\infty z^g \frac{x^{g+1}}{(1-x)^{g+1}}\frac{y^g}{(1-y)^{g+2}}+\frac{1}{1-y}=&\frac{x}{(1-x)(1-y)^2}\frac{(1-x)(1-y)}{(1-x)(1-y)-xyz}+\frac{1}{1-y}\\
	        =&x\frac{\left(\frac{1}{1-y}\right)^2}{1-xf_z(y)}+\frac{1}{1-y}.
	    \end{aligned}
	\end{equation*}
	So in this case,
	$$Q_{L', L}=[x^{L'}y^L]xy\frac{\left(\frac{1}{1-y}\right)^2}{1-xyf(y)}+\frac{1}{1-y}.$$
	The first row of $Q$ is just $1$'s and
	$$(Q_{1:, 1:})_{L', L}=[x^{L'}y^L]\frac{\left(\frac{1}{1-y}\right)^2}{1-xyf(y)}.$$
	$Q_{1:, 1:}$ is a Riordan array with inverse
	$$(Q_{1:, 1:})_{L', L}^{-1}=[x^{L'}y^L]\frac{\left({1-y\bar f(y)}\right)^2}{1-xy\bar f(y)}.$$
	Let $R$ be the upper triangular metrix with $R_{1:, 1:}=(Q_{1:, 1:})^{-1}$, $R_{0, 0}=1$ and $R_{0, 1:}=-\mathbf{1}^T(Q_{1:, 1:})^{-1}$ where $\mathbf{1}$ is the infinite vector of ones. We will see that $R=Q^{-1}$.
    First, for $L, L'\geq 1$,
    $$Q_{L, \cdot}R_{\cdot, L'}=\delta_L(L').$$
    Next for $L\geq 0$,
    $$Q_{L, \cdot}R_{\cdot, 0}=\delta_L(0).$$
    Finally, for $L'\geq 1$,
    $$Q_{0, \cdot}R_{\cdot, L'}=-(\mathbf{1}^T(Q_{1:, 1:})^{-1})_{L'-1} +\mathbf{1}^T(Q_{1:, 1:})^{-1}_{\cdot, L'-1}=0.$$
    Thus $R=Q^{-1}$.
 
	Since $\mathbf{1}_L=1=[x^L](1-x)^{-1}$, 
	\begin{equation*}
	    \begin{aligned}
	        \left(\mathbf{1}^T(Q_{1:, 1:})^{-1}\right)_{L} =&\sum_{L'}(Q_{1:, 1:})^{-1}_{L', L}[x^{L'}](1-x)^{-1}\\
	        = &\sum_{L'}[y^L]\left((1-y\bar f(y))^2\left(y\bar f(y)\right)^{L'}\right)[x^{L'}](1-x)^{-1}\\
	        =&[y^L](1-y\bar f(y))^2\sum_{L'}\left(y\bar f(y)\right)^{L'}[x^{L'}](1-x)^{-1}\\
	        =&[y^L](1-y\bar f(y))^2(1-y\bar f(y))^{-1}\\
	        =&[y^L](1-y\bar f(y))\\
	    \end{aligned}
	\end{equation*}
	Thus, 
	$$Q^{-1}_{L', L}=[x^{L'}y^L]xy\frac{\left({1-y\bar f(y)}\right)^2}{1-xy\bar f(y)} + 1 - y(1-y\bar f(y)).$$
	The rest of the proof is again almost identical to the previous case when $\Delta\mu<\infty$.
	When $\Delta\mu=\infty$, $Q_{0, L}^{-1}=\delta_0(L)-\delta_1(L)+\delta_2(L)$
	and 
	$$(Q_{1:, 1:})^{-1}_{L', L}=[x^{L'}y^L]\frac{\left({1-y}\right)^2}{1-xy}=\delta_{L'}(L)-2\delta_{L'+1}(L)+\delta_{L'+2}(L).$$
	Thus $k_{\mathrm{la}}$ again has discrete masses by Equation \ref{eq: string delta cond for M}.
	
\end{proof}


\section{Computation of thick tailed alignment kernels}\label{sec: thick ali kern}
    We now describe dynamic programming procedures that can be used to calculate the thick tailed alignment kernels described in Examples \ref{ex: ali thick tail} and \ref{ex: ali thick tail feature version}.
    We will deal with both examples as special cases of a single general dynamic programming algorithm.
    Let $\ell(b, b')$, taking values in $1, 0$, be a function for counting occurrences of a certain type of letter pairs.
    Call pairs $(b, b')$ ``$\ell$-type'' if $\ell(b, b')=1$.

    Define
    $$M(i, j, l)=\sum^{(l)}\prod_{i=0}^{c-1} k_I(X_{(:i)}^{(2i)}, Y_{(:j)}^{(2i)})k_s(X_{(:i)}^{(2i+1)}, Y_{(:j)}^{(2i+1)})$$
    Where the sum is over $c$ and over partitions $X_{(:i)}^{(0)}+\dots+X_{(:i)}^{(2c+1)}=X_{(:i)}$, $Y_{(:j)}^{(0)}+\dots+Y_{(:j)}^{(2c+1)}=Y_{(:j)}$ for which $l=\sum_{i=1}^c\ell(X_{(:i)}^{(2i+1)}, Y_{(:J)}^{(2i+1)})$. Thus,
    $M(i, j, l)$ sums over all alignments between $X_{(:i)}$ and $Y_{(:j)}$ with $l$ comparisons that are $\ell$-type and which end in a match between $X_{(i-1)}$ and $Y_{(j-1)}$. 
    Let $I_X(i, j, l)$ be defined similarly as the sum of alignment scores of $X_{(:i)}$ and $Y_{(:j)}$ that have $l$ comparisons that are $\ell$-type and end in an insertion for $X$ but not $Y$.
    Finally let $I_Y(i, j, l)$ be defined similarly as the sum of alignment scores of $X_{(:i)}$ and $Y_{(:j)}$ that have $l$ comparisons that are $\ell$-type and end in an insertion for $Y$.
    Call the sum over all alignments of $X_{(:i)}$ and $Y_{(:j)}$ with $l$ comparisons that are $\ell$-type,
    $$J(i, j, l)=M(i, j, l)+ I_X(i, j, l)+I_Y(i, j, l).$$

    We can now compute $M$, $I_X$ and $I_Y$ with dynamic programming. We initialize at,
    \begin{equation*}
    \begin{split}
    	M(i, 0, l)&=M(0, j, l)=0\text{ for }l, i, j\\
        M(0, 0, 0)&=1\\
        I_X(0, j, l)&=I_Y(i, 0, l)=0\text{ for }l, i, j.
    \end{split}
    \end{equation*}
    The update equations are,
    \begin{equation*}
    \begin{split}
    	M(i, j, l) &= \mathbbm{1}(l>0)\ell(X_{(i-1)}, Y_{(j-1)})k_s(X_{(i-1)}, Y_{(j-1)})J(i-1, j-1, l-1)\\
     &\ \ \ \ \ \ \ \ \ \ +\big(1-\ell(X_{(i-1)}, Y_{(j-1)})\big)k_s(X_{(i-1)}, Y_{(j-1)})J(i-1, j-1, l)\\
        I_X(i, j, l) &= e^{-\Delta\mu-\mu} M(i-1, j, l)+ e^{-\mu}I_X(i-1, j, l)\\
    	I_Y(i, j, l) &= e^{-\Delta\mu-\mu} M(i, j-1, l)+ e^{-\Delta\mu-\mu}I_X(i, j-1, l)+ e^{-\mu}I_Y(i-1, j, l).
     \end{split}
    \end{equation*}
    (We use the arbitrary convention that, between any two adjacent match positions, insertions in $Y$ come before insertions in $X$; thus $I_Y$  does not appear in the update equation for $I_X$.)
    We can then define
    $R(L) = J(|X|, |Y|, L)$
    as the sum over all alignments of $X$ and $Y$ with $L$ comparisons that are $\ell$-type. The dynamic programming algorithm calculates $R(L)$ with $O(|X||Y|(|X|\wedge |Y|))$ operations.

    Now consider the choice $k_s(b, b')=1$ and $\ell(b, b')=\mathbbm{1}(b\neq b')$. In this case, $R(L)$ sums over all alignments where $L$ of the matched positions have mismatched letters, i.e. where the Hamming distance between the matched positions is $L$.  
    Thus, the kernel described in Example \ref{ex: ali thick tail} can be computed as
    \begin{equation*}
    	\tilde k(X, Y) = \sum_{L=0}^{|X|\wedge|Y|}\left(C+L\right)^{-\beta}R(L).
    \end{equation*}
    
    Next set $k_s(b, b')=\delta_b(b')$ and $\ell(b, b')=1$, so $R(L)$ sums over all alignments where there are $L$ matched positions and all have matched letters.
    Recall the definition of $\tilde u_V$ as in Example \ref{ex: ali thick tail feature version}.
    For sequences $V, X, Y$, $\tilde u_V(X)\tilde u_V(Y)$ is the sum over all alignments of $X$ and $Y$ such that the matched positions are $V$ in both $X$ and $Y$, so, $R(L)=\sum_{|V|=L}\tilde u_V(X)\tilde u_V(Y)$.
    Thus, the kernel described in Example \ref{ex: ali thick tail feature version} can be computed as
    \begin{equation*}
    	\tilde k(X, Y) = \sum_{L=0}^{|X|\wedge|Y|}\left(C+\frac 1 2(|X|+|Y|)-L\right)^{-\beta}R(L).
    \end{equation*}

\section{Flexibility of embedding kernels}\label{sec: proofs of flex embedding}
    \begin{proposition}(proof of Propositions~\ref{prop:embedding_injective} and \ref{prop:embed_discrete_mass})
        Assume $k(z, z') = \Psi(z - z')$ is a translation invariant kernel and that $\Psi$ is a positive continuous function on $\mathbb{R}^D$ that has a strictly positive Fourier transform.
        Then the embedding kernel $k$ is characteristic and $C_0$-universal if and only if $F$ is injective.
        Furthermore, $k$ metrizes $\Pc(S)$ if and only if $k$ has discrete masses, which occurs if and only if $F(S)$ has no accumulation points, that is, there is no sequence $X \in S$ such that $F(X)$ is in the closure of $F(S \setminus \{X\})$.
    \end{proposition}
    \begin{proof}
    	First we prove that $k$ metrizes $\Pc(S)$ only if $F$ is injective and $F(S)$ has no accumulation points.
    	Say $F(X_n)\to F(X)$ for a sequence of points with $X_n\neq X$ for all $n$.
    	$k(X_n, X_n)=\Psi(0)=k(X, X)$ and $k(X_n, Y)=\Psi(F(X_n)-F(Y))\to\Psi(F(X)-F(Y))=k(X, Y)$ for all $Y$.
    	Thus 
    	$$\left\Vert \int d\delta_{X_n}(Y)k_{Y}- \int d\delta_X(Y)k_Y\right\Vert_k^2=\Vert k_{X_n}- k_X\Vert_k^2=2k(X, X)-2k(X_n, X)\to 0$$
    	So $k$ does not metrize $\Pc(S)$ as $\delta_{X_n}\not\to\delta_X$.
    	
    	Now we show $k$ has discrete masses if $F$ is injective and $F(S)$ has no accumulation points.
    	Say $k$ does not have discrete masses, so that, by Proposition~\ref{prop: hahn banach delta}, $\sum_{m=1}^{M_n}\alpha_{n, m}k_{X_{n, m}}\to k_{X}$ in $\Hc_k$ for $X_{n, m}\neq X$.
    	Call $\mu_n=\sum_{m=1}^{M_n}\alpha_{n, m}\delta_{F(X_{n, m})}-\delta_{F(X)}$.
        Defining $\hat\Psi$ and $\hat\mu_n$ as the Fourier transforms of $\Psi$ and $\mu_{n}$, we have, as in \citet{Sriperumbudur2011-ay}, that
    	$$\int d\xi |\hat{\mu}_n(\xi)|^2 \hat{\Psi}(\xi)=\int \int d\mu_n(x)d\mu_n(y)\Psi(x-y)=\left\Vert\int d\mu_n(X)k_X\right\Vert_k\to 0.$$
        Thus there is a subsequence $(\hat{\mu}_{n_k})_k$ such that $\hat{\mu}_{n_k}\to 0$ pointwise almost everywhere with respect to the measure $\hat{\Psi}(\xi)d\xi$. Since $\hat{\Psi}$ is strictly positive, $\hat{\mu}_{n_k}\to 0$ pointwise almost everywhere with respect to the Lebesgue measure.
    	This implies that $\mu_{n_k}\to 0$ in distribution.
    	This can happen only if there is a sequence $F(X_{n_k, m_{n_k}})\to F(X)$, i.e. $F(S)$ has an accumulation point.
    	
    	Finally we show $k$ is characteristic and universal if and only if $F$ is injective.
        First, $k$ is not characteristic if $F$ is not injective, since this implies there are two sequences with the same kernel embedding.
    	Now say $F$ is injective and $\nu$ is a finite signed measure with $\int k_Xd\nu(X)=0$.
    	Define $\mu(C)=\nu(F^{-1}(C))$ for all sets $C\subset\mathbb R^D$.
    	Then by the logic above, $\int d\xi |\hat{\mu}(\xi)|^2 \hat{\Psi}(\xi)= \|\int k_X d\nu(X)\|_k = 0.$
    	Thus, $\hat{\mu}=0$ so $\mu=0$.
    	Since $F$ is injective, this implies that $\nu=0$.
    	Thus the map $\nu\mapsto \int d\nu(X) k_X$ is injective for all \textit{signed measures}.
        This is equivalent to $C_0$-universality by Proposition 2 of \citet{Sriperumbudur2011-ay}.
        This also means that the mapping is injective for all probability distributions, so $k$ is characteristic. 
    \end{proof}

\section{Sparse representation learning can yield inflexible embedding kernels}\label{sec: proofs of sparse not inj}
    Consider learning a sequence representation using a matrix factorization model with sparsity-encouraging regularization. For example, consider the following objective,
    \begin{equation}
        \min_{W,B,Z}\,\, \sum_{i=1}^N \| X_i - \sum_d W_d Z_{id} - B \|_2^2 + \lambda \sum_{d=1}^D \sum_{l=0}^\infty \sum_{b \in \B} |W_{dlb}| + \beta \sum_{i=1}^D \|Z_i\|_2^2.
    \end{equation}
    Here, as in the position-wise comparison kernel (Section~\ref{sec: hamming kernels}), we add an infinite tail of stop symbols to each sequence $X \in S$. We treat sequences as one-hot encodings. The vectors $Z_i \in \mathbb{R}^D$ are the representations of the training data, and are mapped linearly, with slope $W$ and offset $B$, to one-hot encoded sequence space. There is $\ell^2$ regularization on the embedding $Z$, with weight $\beta$, and $\ell^1$ regularization on $W$, with weight $\lambda$. 
    This lasso regularization on $W$ encourages each of its entries to be zero, and can make representations more interpretable, in the sense that each dimension of $Z$ only affects a small number of sequence positions.
    Once the model has been trained, we can use it to define an embedding function $F(X) = \mathrm{argmin}_{Z\in \mathbb{R}^D} \|X - \sum_d \hat{W}_d Z_d - \hat{B}\|_2^2 + \beta \|Z\|_2^2$, where $\hat{W}$ and $\hat{B}$ are the parameter values estimated from the training data.
    
    On a finite training data set, there are likely to be positions $l$ in $\hat{W}$ with more than one zero, i.e. for $b \neq b'$ and all $d \in \{1, \ldots, D\}$ we have $W_{dlb} = W_{dlb'} = 0$.
    In this case, a sequence $\tilde X$ with $\tilde X_{lb} = 1$ will map to the same representation as another sequence $\tilde X'$ that is identical to $\tilde X$ except with a different letter at position $l$, namely $\tilde X_{lb'} = 1$. In other words, $F(\tilde X) = F(\tilde X')$ for $\tilde X' \neq \tilde X$, and so $F$ is not injective, and an embedding kernel based on $F$ will be neither universal nor characteristic.

\section{Scaled random embeddings}\label{sec: proofs of random embedding}
    \begin{proposition} (proof of Propositon~\ref{prop:scaled_embedding})
        Consider an initial embedding $\tilde F$ where each $\tilde F(X)$ for $X \in S$ is drawn from the uniform distribution on the sphere, $\{x \in \mathbb{R}^D\, |\, \|x\| \le 1\}$.
        Then, a kernel using the scaled embedding $F(X)=|\B|^{(1+\epsilon)|X|/D}\tilde F(X)$, for $\epsilon>0$, metrizes $\Pc(S)$ almost surely.
    \end{proposition}
    \begin{proof}
    The probability that $F$ is injective is $1$, following the same logic as in the proof of Proposition~\ref{prop:random_universal}.
        Now, for any $N$ we will show that with probability $1$, $\{X\ |\ \Vert F(X)\Vert \leq N\}$ is finite.
        This will imply that for any sequence of distinct sequences $X_1, X_2, \dots$, we must have $\Vert F(X_n)\Vert_\infty \to \infty$ as $n\to\infty$, and in particular the embedding $F(S)$ has no accumulation points.
    	Now, $$p(\Vert F(X)\Vert\leq N)=p(\Vert \tilde F(X)\Vert\leq N/|\B|^{(1+\epsilon)|X|/D}).$$
        If $|\B|^{-(1+\epsilon)|X|/D}N\geq 1$ then this quantity is $1$.
        Otherwise, calling $V$ the Lebesgue measure and $B(1)$ the unit ball $\{x \in \mathbb{R}^D\, |\, \|x\| \le 1\}$, then
        $$p(\Vert F(X)\Vert\leq N)=\frac{V(|\B|^{-(1+\epsilon)|X|/D}NB(1))}{V(B(1))}=|\B|^{-(1+\epsilon)|X|}N^D.$$
        Thus, 
        $$\sum_{X\in S} p(\Vert F(X)\Vert\leq N)\leq N^D \sum_{X\in S}|\B|^{-(1+\epsilon)|X|}=N^D \sum_{L=0}^\infty|\B|^{-L\epsilon}<\infty.$$
        So by the Borel-Cantelli lemma, with probability $1$, $F(S)$ has no accumulation points.
    \end{proof}

\bibliography{sample}

\end{document}